\RequirePackage{fix-cm}
\documentclass[smallextended]{svjour3}       
\smartqed  
\usepackage{fixltx2e}

\usepackage[pass]{geometry}                       	

\usepackage{amsmath,graphicx, hyperref, footnote}
\usepackage{longtable}
\usepackage[table]{xcolor}
\usepackage{array}
\usepackage{enumerate}
\usepackage{color}
\usepackage{graphics}
\usepackage{epsfig}
\usepackage{amssymb}
\usepackage{fancyhdr}
\usepackage{tabularx}
\usepackage{pbox}

\usepackage{latexsym}
\usepackage{graphicx}
\usepackage{subfigure}
\usepackage{amsmath}
\usepackage{times}
\usepackage{algorithm}
\usepackage{algorithmic}

\usepackage{url}
\usepackage{amsmath}
\usepackage{tabularx} 
\usepackage{amssymb}
\usepackage{verbatim}
\usepackage{amsfonts}
\usepackage{longtable}
\usepackage{graphicx}
\usepackage{booktabs}
\usepackage{epstopdf}
\usepackage{longtable}
\usepackage{multirow}
\usepackage{varwidth}
\usepackage{footnote}
\usepackage{newclude}

\usepackage{multirow}
\usepackage{url}
\usepackage{appendix}
\usepackage{subfloat}

\DeclareMathOperator{\val}{=}  

\def\happensAt{\textsf{\scriptsize happensAt}}

\def\holdsAt{\textsf{\scriptsize holdsAt}}
\def\holdsFor{\textsf{\scriptsize holdsFor}}
\def\initiatedAt{\textsf{\scriptsize initiatedAt}}
\def\terminatedAt{\textsf{\scriptsize terminatedAt}}

\def\startE{\textsf{\scriptsize start}}
\def\endE{\textsf{\scriptsize end}}

\def\intersectall{\textsf{\scriptsize intersect\_all}}

\def\nbf{\textsf{\scriptsize not}}
\def\true{\textsf{\scriptsize true}}

\newenvironment{mysplit}%
  {\arraycolsep 0pt \begin{array}{l}}%
  {\end{array}}

\newcommand{\footnoteremember}[2]{\footnote{#2}\newcounter{#1}\setcounter{#1}{\value{footnote}}}

\begin{document}

\title{Online Event Recognition from Moving Vessel Trajectories}


\author{Kostas Patroumpas \and Elias Alevizos \and Alexander Artikis \and Marios Vodas \and Nikos Pelekis \and Yannis Theodoridis}


\institute{K. Patroumpas \at School of Electrical \& Computer Engineering, National Technical University of Athens, Greece\\
           \email{kpatro@dblab.ece.ntua.gr}           
           \and
           E. Alevizos \at Institute of Informatics \& Telecommunications, NCSR Demokritos, Athens, Greece\\
           \email{alevizos.elias@iit.demokritos.gr}
           \and
           A. Artikis \at Department of Maritime Studies, University of Piraeus, Greece\\
           Institute of Informatics \& Telecommunications, NCSR Demokritos, Athens, Greece\\
           \email{a.artikis@unipi.gr}
           \and
           M. Vodas \at Department of Informatics, University of Piraeus, Greece\\
           \email{mvodas@unipi.gr}
           \and
           N. Pelekis \at Department of Statistics \& Insurance Science, University of Piraeus, Greece\\
           \email{npelekis@unipi.gr}
           \and
           Y. Theodoridis \at Department of Informatics, University of Piraeus, Greece\\
           \email{ytheod@unipi.gr}
}

\date{Received: date / Accepted: date}

\maketitle

\begin{abstract}

We present a system for online monitoring of maritime activity over streaming positions from numerous vessels sailing at sea. It employs an online tracking module for detecting important changes in the evolving trajectory of each vessel across time, and thus can incrementally retain concise, yet reliable summaries of its recent movement. In addition, thanks to its complex event recognition module, this system can also offer instant notification to marine authorities regarding emergency situations, such as risk of collisions, suspicious moves in protected zones, or package picking at open sea. 
Not only did our extensive tests validate the performance, efficiency, and robustness of the system against scalable volumes of real-world and synthetically enlarged datasets, but  its deployment against online feeds from vessels has also confirmed its capabilities for effective, real-time maritime surveillance.

\keywords{AIS \and event recognition \and geostreaming \and moving objects \and trajectory }
\end{abstract}

\section{Introduction} %
\label{sec:intro}

Maritime surveillance systems have been attracting considerable attention for economic as well as environmental reasons \cite{DBLP:conf/sysose/IdiriN12,terroso-saenz_complex_2015,DBLP:journals/inffus/SnidaroVB15}. For instance, accidents at sea may cause ecological disasters (e.g., oil spill) and shipping companies may be fined to pay billions of dollars. In the past decade, monitoring vessel activity has emerged as a precious tool for preventing such risks, thanks to the Automatic Identification System (AIS)\footnote{\scriptsize \url{http://www.imo.org/OurWork/Safety/Navigation/Pages/AIS.aspx}}. By integrating a VHF transceiver with positioning and navigational devices (e.g., GPS, gyrocompass), AIS can be used to track vessels at sea in real-time through data exchange with other ships nearby, coastal stations, or even satellites. The initial purpose of AIS was to prevent collisions; yet, the amount and precision of the collected data and its real-time availability can be used by a broader spectrum of maritime monitoring applications. International regulations require AIS to be aboard cargo ships of at least 300 gross tonnage, as well as all passenger ships, regardless of size. Considering that AIS data is continuously emitted from over 580,000 vessels worldwide\footnote{\url{https://www.vesselfinder.com}}, maritime surveillance systems certainly demand capabilities of highly scalable, continuous, spatiotemporal processing over massive data streams.

To address this requirement, we have been developing a maritime surveillance system that consists of two main components. A \textit{trajectory detection} component accepts a positional stream of AIS messages and tracks major changes along each vessel's movement. Given that vessels normally follow planned routes (except for accidents, storms, etc.), this process can instantly identify  {\em``critical points''} along its trajectory, such as a stop, a sudden turn, or slow motion. Therefore, we may discard redundant locations along a ``normal'' course, and approximately reconstruct each vessel's trajectory from the sequence of its critical points only. This online summarization achieves data compression close to 98\%, with negligible loss in approximation quality. But, apart from archiving or displaying it on maps, this derived stream of critical points is mostly useful in recognizing complex maritime phenomena that involve interaction among vessels or spatiotemporal relationships between vessels and geographical areas of interest. This is handled by our \textit{complex event recognition} component, which can efficiently detect suspicious or potentially dangerous situations, such as fast approaching vessels or package picking at open sea, and accordingly issue alert notifications to marine authorities.




 Of course, several platforms and monitoring applications have been proposed for managing and analyzing data streams. For instance, the system in \cite{dindar11} focuses on recency-probing pattern queries against both live and archived streams. UpStream platform \cite{[MT11]} offers low latency response to continuous queries over massively updated data, but it lacks support for the specific demands of trajectory detection. Besides, automating ingestion of streaming data feeds from various sources into a data warehouse is also important \cite{[GT13]}, but does not pay attention to complex event recognition. Our particular interest is on {\em geostreaming} data \cite{[KDA+10]} from sailing vessels acquired continuously over time, which must be processed on-the-fly in order to recognize important phenomena regarding their movement and their interaction with the maritime environment. In \cite{shahir15} an approach for anomaly detection and classification of vessel interactions is presented. Patterns of interest are expressed as left-to-right Hidden Markov Models and classified using Support Vector Machines, also taking into account contextual information via first-order logic rules. However, this work focuses more on predictive accuracy rather than real-time performance in a streaming scenario. To the best of our knowledge, no streaming framework has been specifically tailored for maritime surveillance over fluctuating, noisy, intermittent, geostreaming AIS messages from large fleets, as the one we present in this work.
 
This paper is an extended and revised version of previous works presented in \cite{[PAK+15],alevizos15}, and developed in the context of the AMINESS project\footnote{\scriptsize \url{http://www.aminess.eu/}}. It now offers heuristics for coping with noisy situations, improved algorithms for better capturing important events related to vessel mobility and interaction, as well as a more thorough empirical validation. In particular, our contributions are:

\begin{itemize}
\item We introduce single-pass heuristics to drastically reduce noisy AIS positions, much to the benefit of the resulting trajectory synopses (in size and quality).

\item We provide a detailed account of online spatiotemporal filters that can detect important changes in each vessel's mobility and also incrementally maintain succinct, reliable representations of their evolving trajectories.

\item We analyze in depth several rules and conditions for efficiently recognizing complex maritime events, which may also involve topological relationships between vessels and geographical zones of interest. 

\item We empirically validate our methodology in terms of performance and quality of results against a large AIS dataset of real vessel traces. Moreover, we extensively investigate the robustness, efficiency, and timeliness of the system against scalable volumes of synthetically enlarged datasets, as well as its capacity to recognize complex events related to sensitive environmental zones.



\item Finally, we outline a deployment of the system in a real-world scenario, against streaming AIS data feeds that are being collected across the Aegean Sea.
\end{itemize}

The remainder of the paper is organized as follows. In Section \ref{sec:architecture}, we present the architecture of the proposed maritime surveillance system. Sections \ref{sec:trajectory} and \ref{sec:cer} respectively present the two main components for online trajectory detection and complex event recognition, respectively. Experimental results are reported in Section~\ref{sec:empirical}. In Section \ref{sec:deployment}, we discuss a deployment of this system against real-time AIS data. Finally, in Section~\ref{sec:summary} we summarize our approach and outline directions for future work.

\section{System Architecture}
\label{sec:architecture}

\begin{figure*}[t]
\centering
{\includegraphics[width=\textwidth]{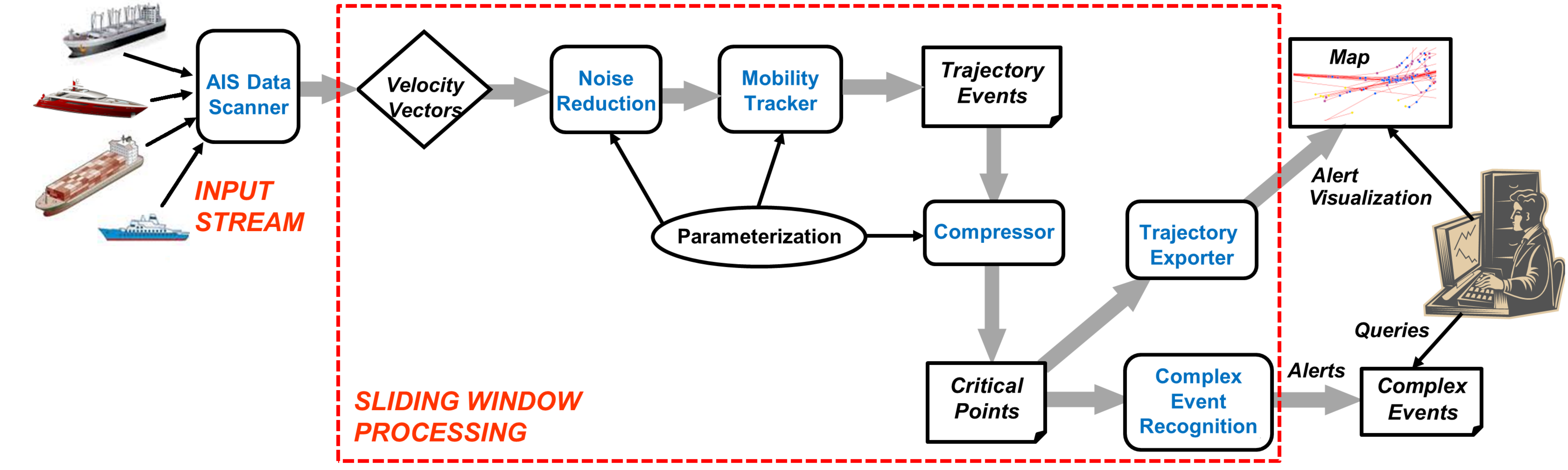}}
\caption{Processing flow of streaming AIS data for online maritime surveillance.} \label{Architecture}
\end{figure*}

Next, we outline the processing flow of the proposed maritime surveillance system. As illustrated in Figure~\ref{Architecture}, this system consumes a geospatial stream of AIS tracking messages from vessels, continuously detects important features that characterize their movement and recognizes complex events such as suspicious vessel activity. 


In order to meet the real-time requirements of the geostreaming paradigm \cite{[KDA+10]}, this online process necessitates the use of a {\em sliding window} \cite{kramer09,[PS11]}. Typically, such a window abstracts the time period of interest by focusing only on phenomena that occurred in a recent {\em range} $\omega$ (e.g., positions received during past 10 minutes). This window slides forward to keep in pace with newly arrived stream tuples, so it gets refreshed at a specific {\em slide step} every $\beta$ units (e.g., each minute). For instance, an aggregate query could report at every minute ($\beta$) the distance traveled by a ship over the past 10 minutes ($\omega$). Typically, $\beta < \omega$, hence successive window instantiations may share positional tuples over their partially overlapping ranges across time.

As input, we consider particular AIS messages (specifically, of types 1, 2, 3, 18, or 19 according to AIS regulations) and extract position updates. Each message specifies the $\mathit{MMSI}$ (Maritime Mobile Service Identity) of the reporting vessel. For a given $\mathit{MMSI}$, each of its successive positional samples $p$ consists of geographical coordinates $(Lon, Lat)$ observed at discrete, totally ordered timestamps $\tau$ (e.g., at the granularity of seconds). Without loss of generality, we abstract vessels as 2-dimensional point entities moving across time, because our primary concern is to capture their motion features. By monitoring the timestamped locations from a large fleet of $N$ vessels, the system must deal with a {\em positional stream} of tuples $\langle \mathit{MMSI}, Lon, Lat, \tau \rangle$. A {\em Data Scanner} decodes each AIS message, identifies those four attributes (the rest are ignored in our analysis), and cleans them from distortions caused during transmission (e.g., discard corrupt messages with bad checksum). This constitutes an {\em append-only} data stream, as no deletions or updates are allowed to already received locations.

But it is the sequential nature of each vessel's trace that mostly matters for capturing movement patterns {\em en route} (e.g., a slow turn), as well as spatiotemporal interactions (e.g., ships traveling together). Such a {\em trajectory} is approximated as an evolving sequence of successive point samples 
that locate this vessel at distinct timestamps (e.g., every few seconds). Our system accounts for stream imperfections, i.e., the noise inherent in vessel positions due to sea drift, delayed arrival of messages, or discrepancies in GPS signals. Indeed, prior to any processing, all incoming AIS positions are filtered through the {\em Noise Reduction} module by applying heuristics against a velocity vector maintained per vessel\footnote{\scriptsize Typically for trajectories \cite{[CWT06]}, linear interpolation is applied between each pair of successive measurements 
($p_i, \tau_i$) and ($p_{i+1}, \tau_{i+1}$). For simplicity, we assume that this also holds in the case of vessels. With the exception of intermittent signals, 
their course between any two consecutive positions practically evolves in a very small area, which can be locally approximated with a Euclidean plane using Haversine distances.}. Afterwards, the {\em Mobility Tracker} module accepts clean data and checks when and how velocity changes with time. Working entirely in main memory and without any index support, it can detect {\em trajectory events}, either instantaneous (e.g., a sudden turn) or of longer duration (e.g., a smooth turn). At each window slide, those events are compiled by a {\em Compressor} and a sequence of {\em ``critical''} points (such as a stop) are emitted, which are much fewer compared to the originally relayed positions. Accordingly, the current vessel motion can be characterized in real time with particular {\em annotations} (e.g., stop, turn). Once new trajectory events are detected per vessel upon each window slide, the annotated critical points can be readily emitted and visualized on maps through a {\em Trajectory Exporter}, e.g., as KML polylines (for trajectories) and placemarks (for vessel locations).

Not surprisingly, detecting trajectory events from positional streams essentially performs a kind of path simplification. In the literature, some strategies like  \cite{[CWT06],[LDR11],[MdB04]} specify an error tolerance for the resulting approximation. The memory footprint occupied by the compressed trajectory may also be a constraint in a single-pass evaluation \cite{[PPS07]}. Mainly focusing on savings in communication cost, dead-reckoning policies like \cite{[WSCY99]} may be employed on board of the moving objects to relay positional updates only upon significant deviation from the course already known to a centralized server. However, this does not hold for AIS data, as maritime control centers wish to locate ships as frequently as possible. Most importantly, a major advantage of our proposed scheme is that it annotates the simplified representations according to particular trajectory events (turn, stop, etc.), thus adding rich semantic information all along each compressed trace.

Moreover, the derived critical points are propagated to the {\em Complex Event Recognition} module, which combines this event stream with static geographical data, such as protected areas. The objective of this process is to detect potentially suspicious or dangerous situations, such as loitering and vessel pursuit. The recognized complex events are pushed in real-time to the marine authorities for decision-making.

\section{Detecting Trajectory Events}
\label{sec:trajectory}

As illustrated in Figure~\ref{Architecture}, the system accepts fresh AIS messages from ships and extracts positional tuples $\langle \mathit{MMSI}, Lon, Lat, \tau \rangle$. With the possible exception of local manoeuvres near ports, marine regulations, or harsh weather conditions, vessels are normally expected to follow almost straight, predictable routes. In terms of vessel mobility, what matters most is to detect when and how the general course has changed, e.g., identify a stop, a turning point, or slow motion. Such {\em trajectory movement events} (ME) suffice to indicate ``critical points'' along the trace of each vessel and thus offer a concise, yet quite reliable representation of its course. 

In order to identify significant changes in movement, we employ an instantaneous velocity vector $\overrightarrow{v}_{now}$ computed from the two most recent positions reported by each vessel $\mathit{MMSI}$. In addition, we maintain the mean velocity $\overrightarrow{{v}_{m}}$ per ship over its previous $m$ positions ($m$ is a small integer) so as to abstract its short-term course. With our heuristics, it turns out that a large portion of the raw positional reports can be suppressed with minimal loss in accuracy, as they hardly contribute any additional knowledge. We distinguish two kinds of trajectory events:

\begin{itemize}
\item {\em Instantaneous trajectory events} involve individual time points per route, by simply checking potentially important changes with respect to the previously reported location (e.g., a sharp change in heading).
\item {\em Long-lasting trajectory events} are deduced after examining a sequence of instantaneous events over a longer (yet bounded) time period in order to identify evolving motion changes. For example, a few consecutive changes in heading may be very small if each is examined in isolation from the rest, but cumulatively they could signify a notable change in the overall direction.
\end{itemize}

In this Section, we first present simple, yet quite effective filters as a means of eliminating noise inherent in the streaming AIS data. Next, we describe how the sequence of vessel positions can be processed {\em online} in order to detect trajectory movement events and thus maintain a lightweight synopsis of each vessel's course.

\subsection{Online Noise Reduction}
\label{sec:noise}

Despite its high value in maritime surveillance, AIS data is not error-free. In fact, there are several sources of error that render a portion of this data noisy and inadequate for monitoring. First, no precise timestamp value is present in AIS messages relayed by the transponders on board; instead, they only report a lag value (in seconds) from the previously transmitted message. Obviously, this value cannot be used for establishing a temporal order, since positional updates from a single vessel may come from a series of base stations (those within range of its antenna along the route). 

Therefore, a {\em transaction timestamp} marking the arrival of each AIS message at a station has to be used instead. Inevitably, transmission delays may frequently occur between the original message and its arrival. Successive positional messages from a single vessel may often arrive intermingled at a distorted order. Figure~\ref{noise}(a) illustrates such {\em out-of-sequence} messages, where numbers signify timestamp values since the beginning of this trajectory. Had those positions been retained according to their order of arrival, the vessel would occasionally appear in a state of sudden {\em agility}, moving back and forth very rapidly at a quite unusual speed. To make matters worse, AIS networks do not have synchronized clocks. Hence, if a vessel is within range of two stations, then a broadcasted message may be received by each one and possibly assigned with a different transaction timestamp. It may also happen that the same timestamp is assigned to different locations (maybe of considerable distance) of the same vessel. Therefore, {\em duplicate} or {\em contradicting positions} of a vessel may be present in the collected data. Noise might not always be caused by technical issues or the inherent errors and discrepancies in the GPS positions. It may be also due to deliberate, {\em suspicious actions}, e.g., switching off the transponder or emitting ``spoofed'' coordinates in order to avoid surveillance in a sensitive area. To the extent possible, such intermittent or falsified positions should be detected and cleared.

Coping with noisy situations over AIS data is particularly challenging and has attracted significant research interest. As argued in \cite{[PVB13]}, noise reduction should not be confused with anomaly detection, because data must be cleaned in advance, before performing any analysis. In that particular work, the well-known DBSCAN clustering algorithm \cite{[EKSX96]} was used to identify outlier positions, which could then be removed from the dataset. With respect to time delays, an adaptive filtering strategy was suggested in \cite{[MBBW15]}, which employed Kalman filters and Monte-Carlo simulations to sequentially detect such delays and probabilistically rearrange a ``correct'' timestamp order. Besides, a stochastic method can be used to cope with position spoofing \cite{[KBC13]} by exploiting auxiliary data from radars or comparing it with previously tracked information. A common characteristic of all such methods is that they work in offline fashion employing expensive, iterative filters over archived AIS datasets.

In contrast, in this work we wish to apply {\em online}, single-pass filters over the incoming stream of AIS positions. Besides, given that trajectory compression is one of our principal objectives, we can afford to lose garbled, out-of-sequence positions and not consider correcting their timestamps. After all, unless in cases of malicious activity, a fresh noise-free location will be soon received from a vessel, effectively compensating for the removal of any erroneous preceding one(s). In order to effectively and efficiently eliminate noise in timestamped AIS positions, we resort to applying a series of simple heuristics that examine the instantaneous velocity vector $\overrightarrow{v}_{now}$  of each vessel as computed by its two most recent observations. A noisy situation is identified if at least one of the following conditions apply:

\begin{figure}[t]
\centering
\subfigure[Out-of-sequence] {\fbox{\includegraphics[width=33mm]{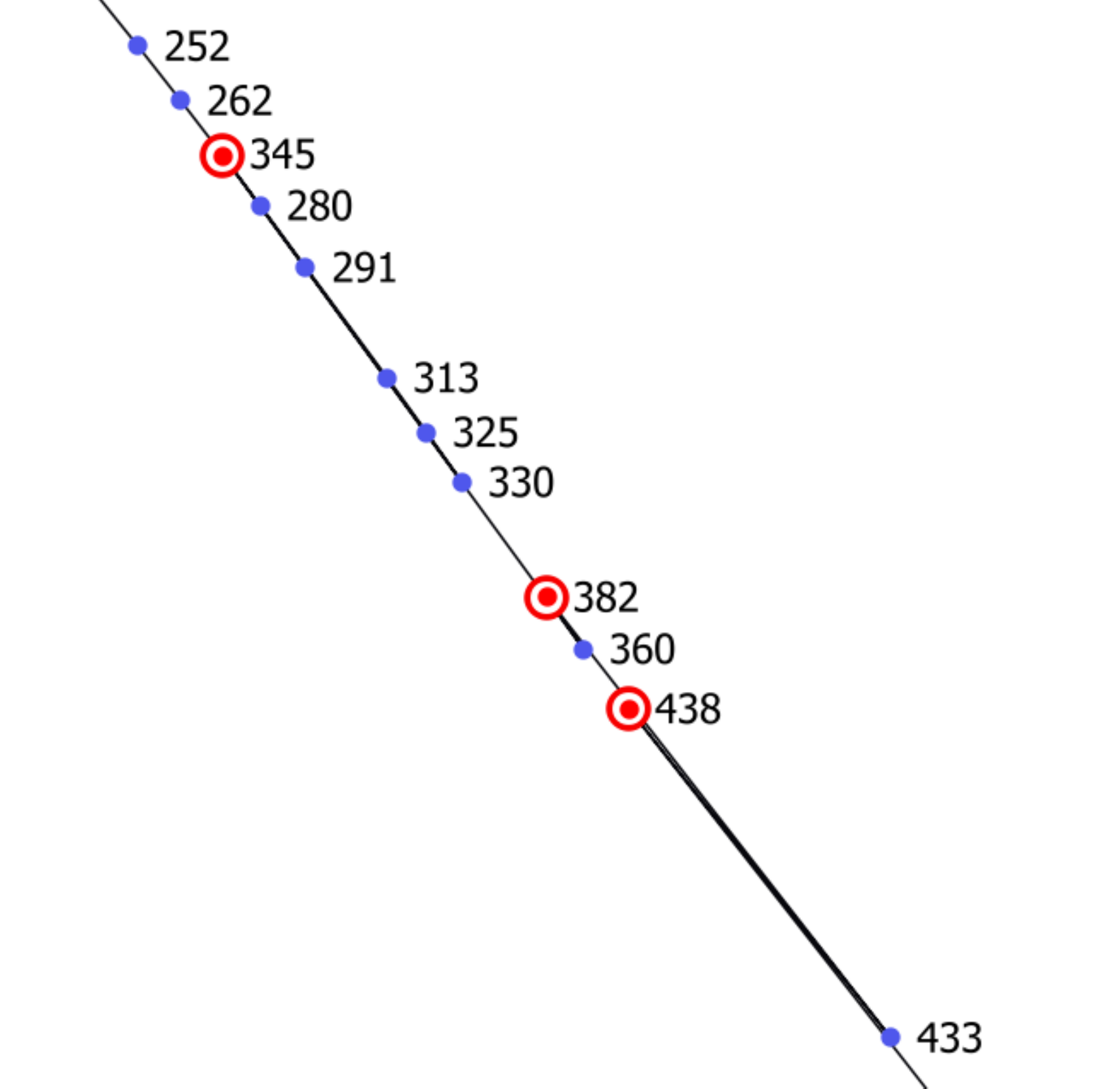}}}
\subfigure[Off-course position] {\fbox{\includegraphics[width=33mm]{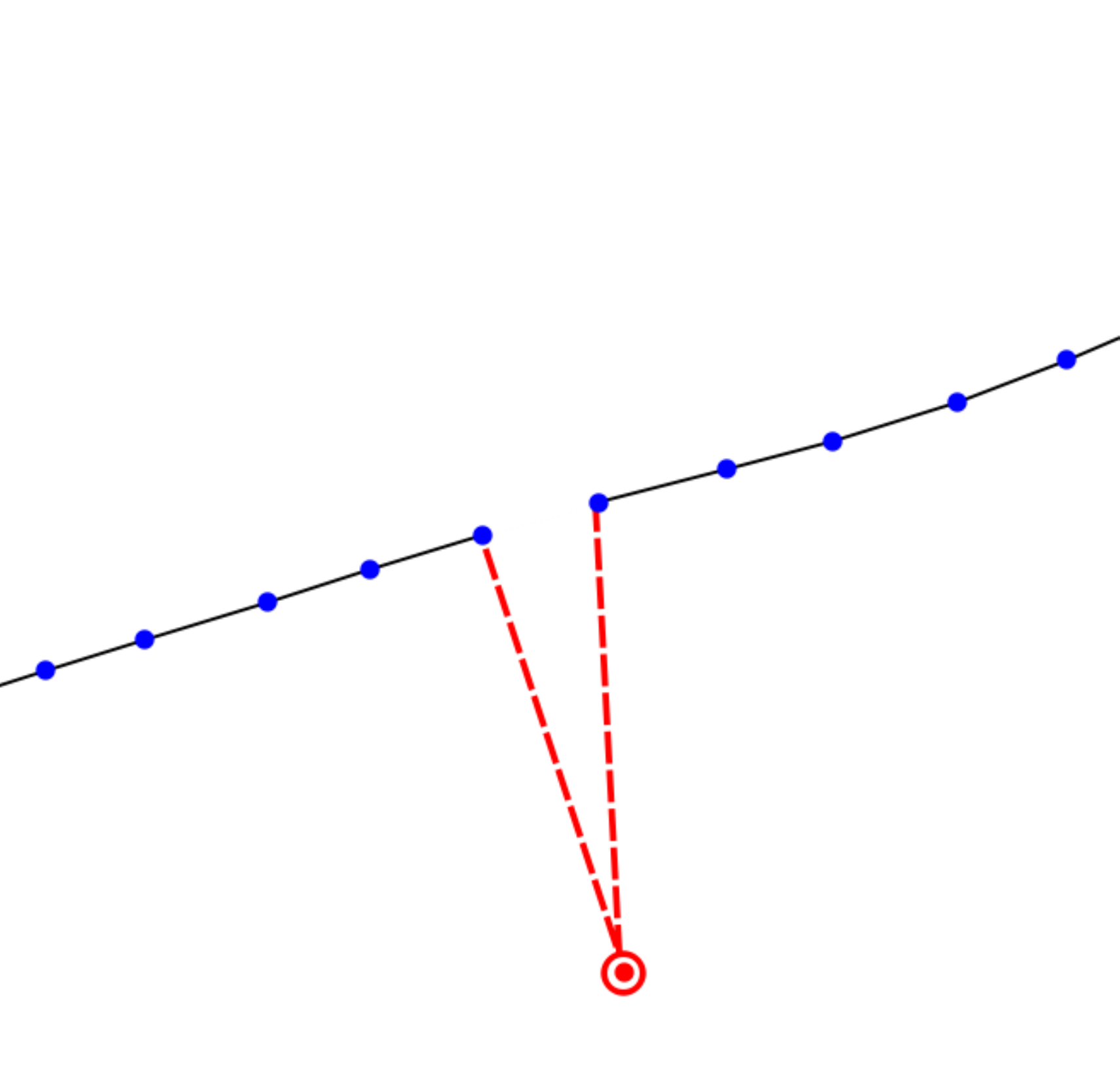}}}\label{outlier}
\subfigure[Zig-zag movement] {\fbox{\includegraphics[width=33mm]{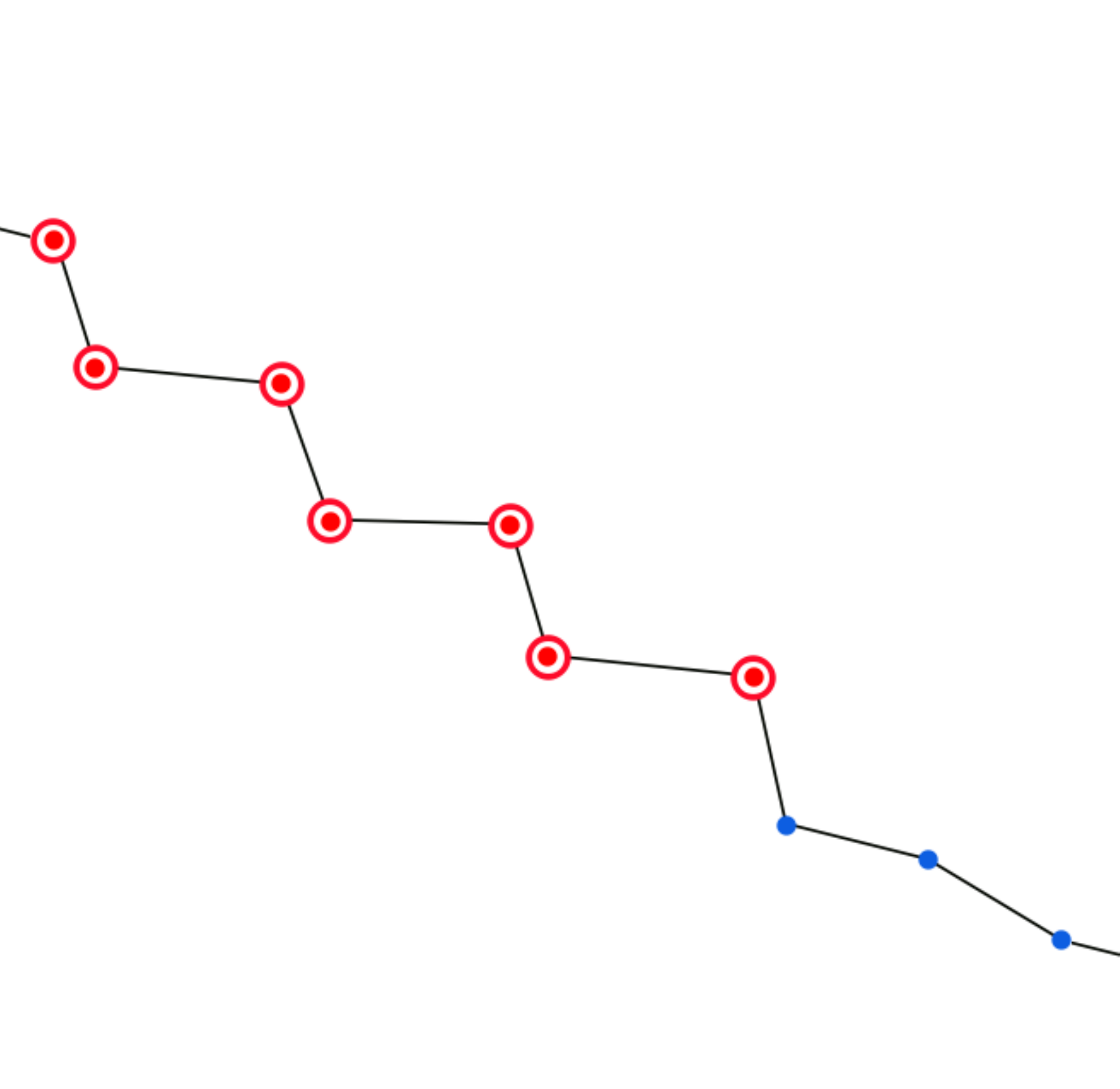}}}\label{zigzag}
\caption{Noise-related situations along a vessel's course.} \label{noise}
\end{figure}

\begin{itemize}
\item {\em Off-course positions} incur an abrupt change both in speed and heading of velocity $\overrightarrow{v}_{now}$. Such an outlier can be easily detected since it signifies an abnormal, yet only temporary, deviation from the known course as abstracted by mean velocity $\overrightarrow{{v}_{m}}$ of the ship over its previous $m$ positions. Figure~\ref{noise}(b) illustrates such a case with a vessel that is unexpectedly located far away from its anticipated route.
\item When vessels are on the move, they normally take their turns very smoothly (especially larger ships), so a series of AIS locations are transmitted, each marking a small change in heading as in Figure~\ref{longterm}(f). However, if the latest position update indicates that a vessel has suddenly made a very abrupt turn (e.g., over $60^o$) with respect to its known course (even though its speed may not be altered significantly), then this message should better be ignored altogether. Note that in case of adverse weather conditions (e.g., a storm) a vessel's route may appear as a {\em `zig-zag'} polyline with a series of such abrupt turns as shown in Figure~\ref{noise}(c). Dropping those consecutive points as noise is not typically correct; yet, in terms of data reduction this is quite desirable, as the vessel does not make any intentional turn and generally follows its planned course. 
\item When a vessel appears to accelerate too much, i.e., at a rate that it is not usual for a ship, this is another indication of noise. This is typical for {\em out-of-sequence} messages with twisted timestamps, as the three red spots in Figure~\ref {noise}(a). Each of these three locations is along the course of the ship, but due to their late arrival to the base station, the vessel is seen as retracting backwards suddenly. The location at timestamp $t=438$ seconds is 270 meters back from the position at $t=433$ seconds, resulting in a speed of 105 knots, quite unrealistic for any vessel.
\item If an identical location from the same vessel has been already recorded before, then this might be a sign of error. Note that even if a vessel remains anchored, its successive GPS measurements usually differ by a few meters. In that case, instantaneous velocity $\overrightarrow{v}_{now}$ is infinitesimal, but not exactly zero; instead, coincidental coordinates in succession should be deemed as almost certain duplicates. 
\item A similar problem occurs with conflicts in timestamping, when the same timestamp is assigned to two distinct messages from a given vessel, even though they may be probably reporting different coordinates. In this case, instantaneous velocity $\overrightarrow{v}_{now}$ cannot be computed, signifying that these messages are contradictory (if not violating previous rules, we arbitrarily retain the latest one).
\end{itemize}

As we experimentally verified (cf. Section~\ref{sec:empirical}), as much as 20\% of the raw AIS positions may be qualifying as noise, falling in one of the aforementioned cases. Most importantly, accepting noisy positions would drastically distort the resulting trajectory synopsis, as the red dashed line in Figure~\ref{noise}(b) illustrates. Even worse, noise may affect proper detection of movement events, as we will discuss next. Hence, although based in empirical heuristics, such noise reduction has proven certainly beneficial in terms of performance without sacrificing accuracy in the resulting approximation.

\subsection{Online Tracking of Moving Vessels}
\label{sec:tracking}

\begin{figure}[t]
\centering
\subfigure[Pause] {\fbox{\includegraphics[width=33mm]{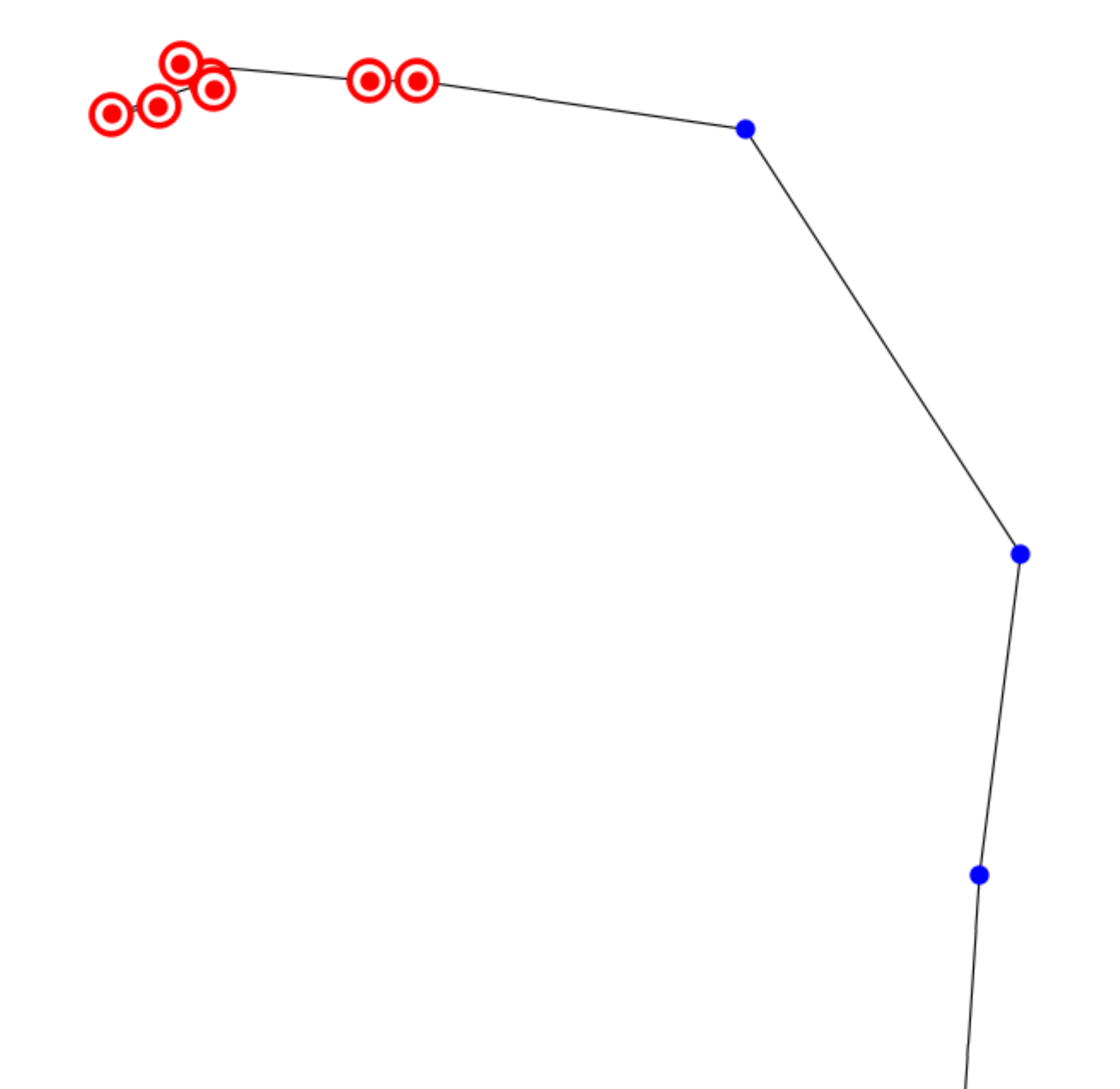}}}
\subfigure[Change in speed] {\fbox{\includegraphics[width=33mm]{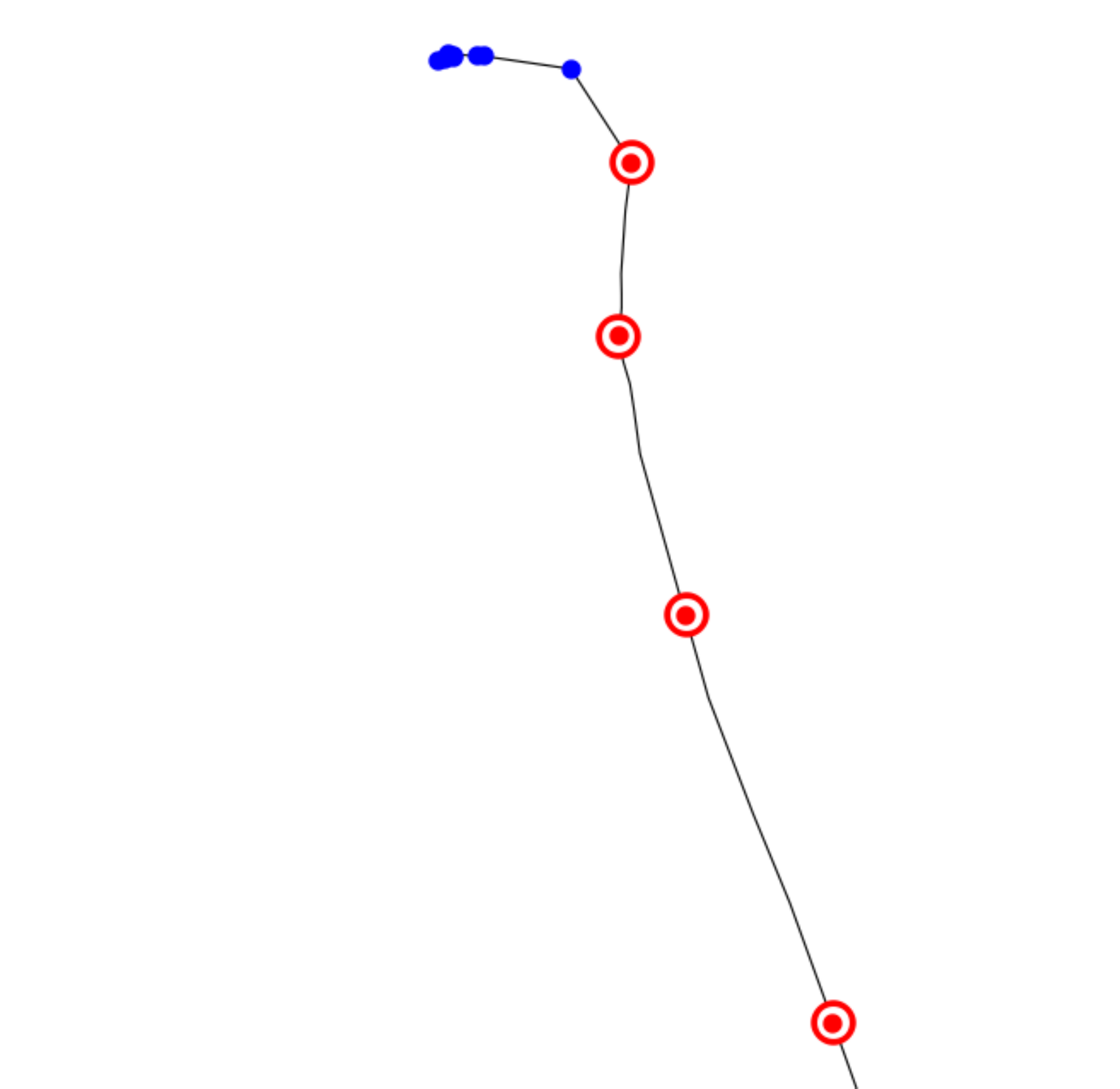}}}
\subfigure[Gap in reporting] {\fbox{\includegraphics[width=33mm]{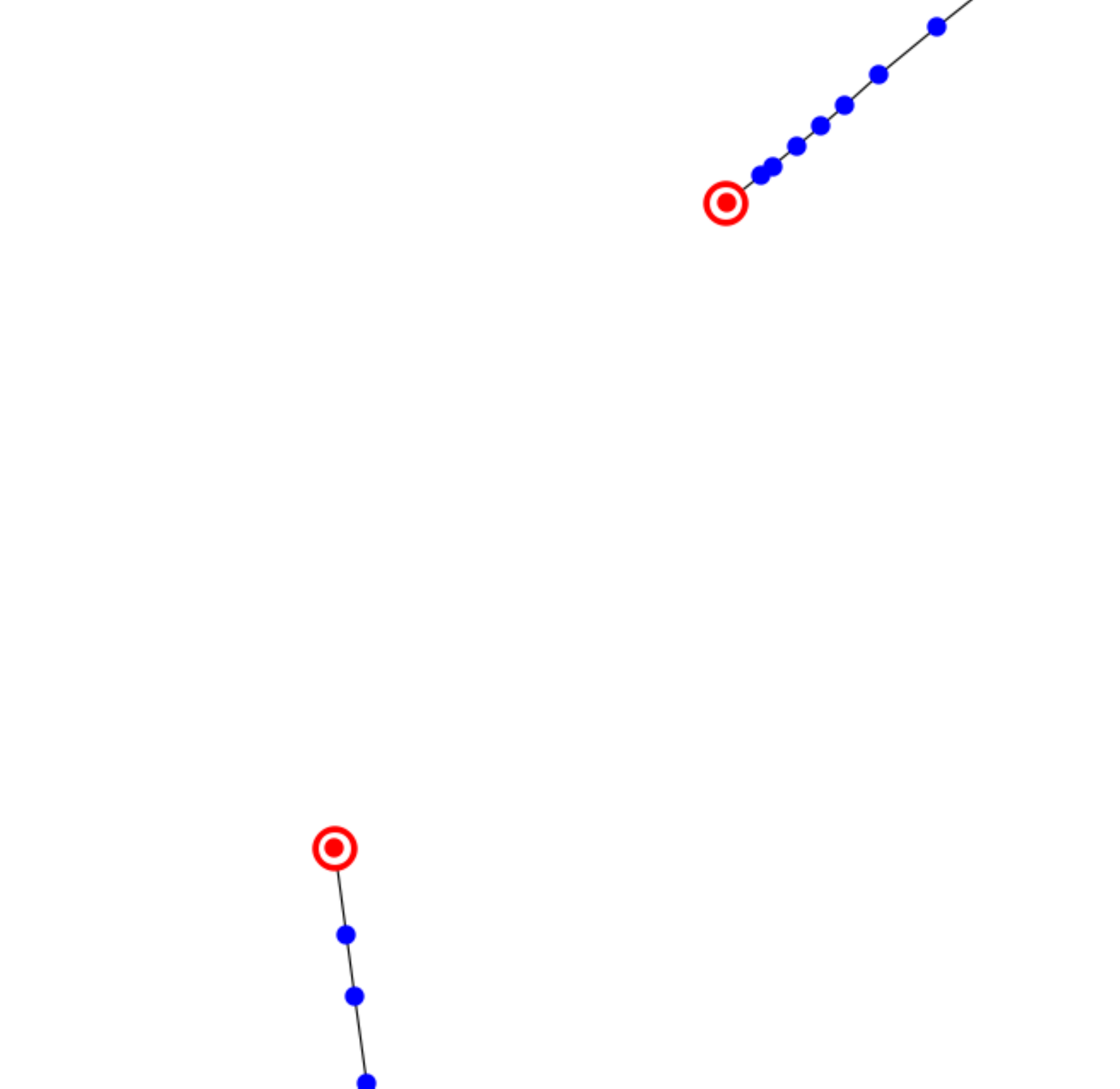}}}
\linebreak
\subfigure[Long-term stop] {\fbox{\includegraphics[width=33mm]{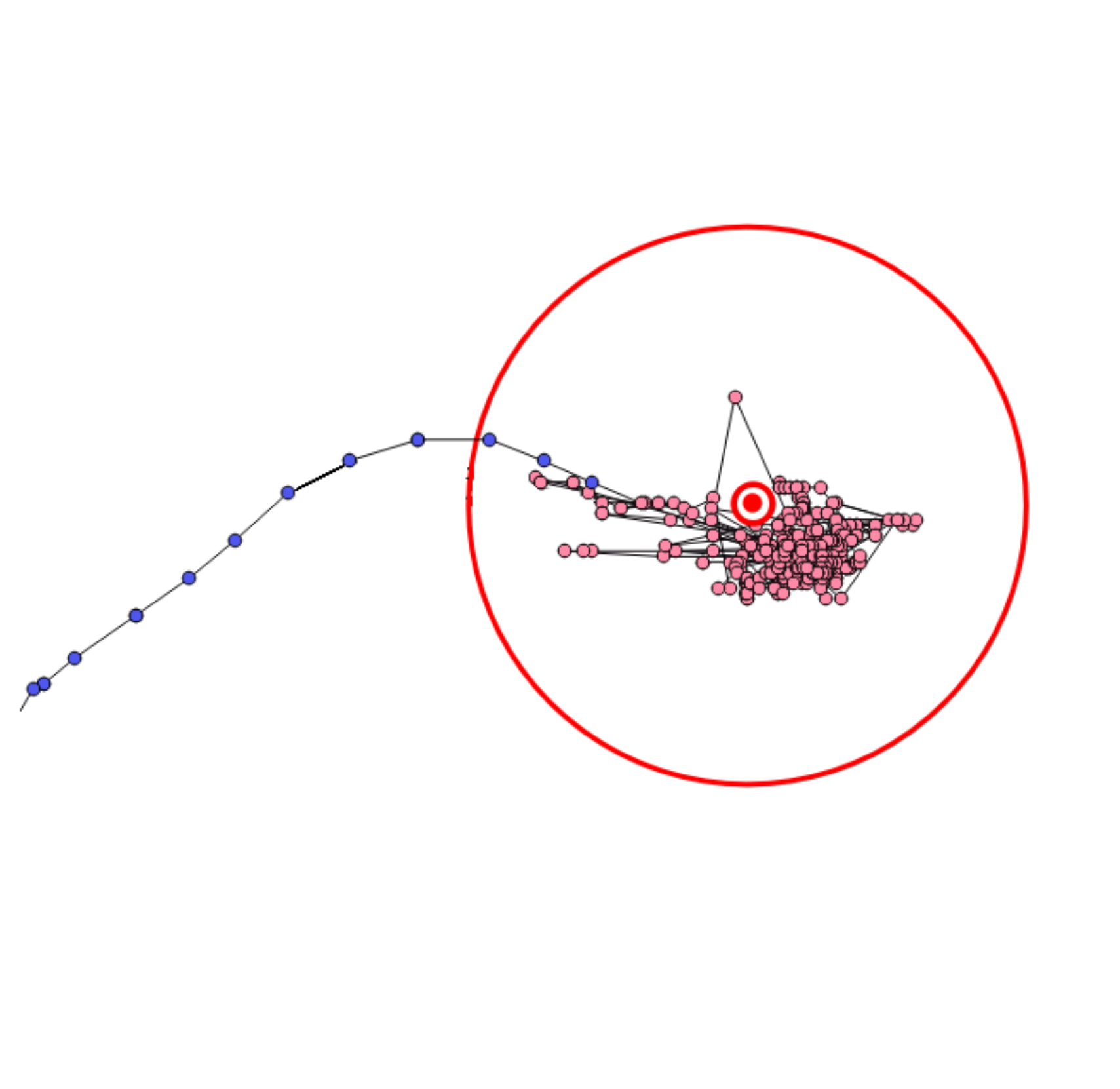}}}\label{long_stop}
\subfigure[Slow motion] {\fbox{\includegraphics[width=33mm]{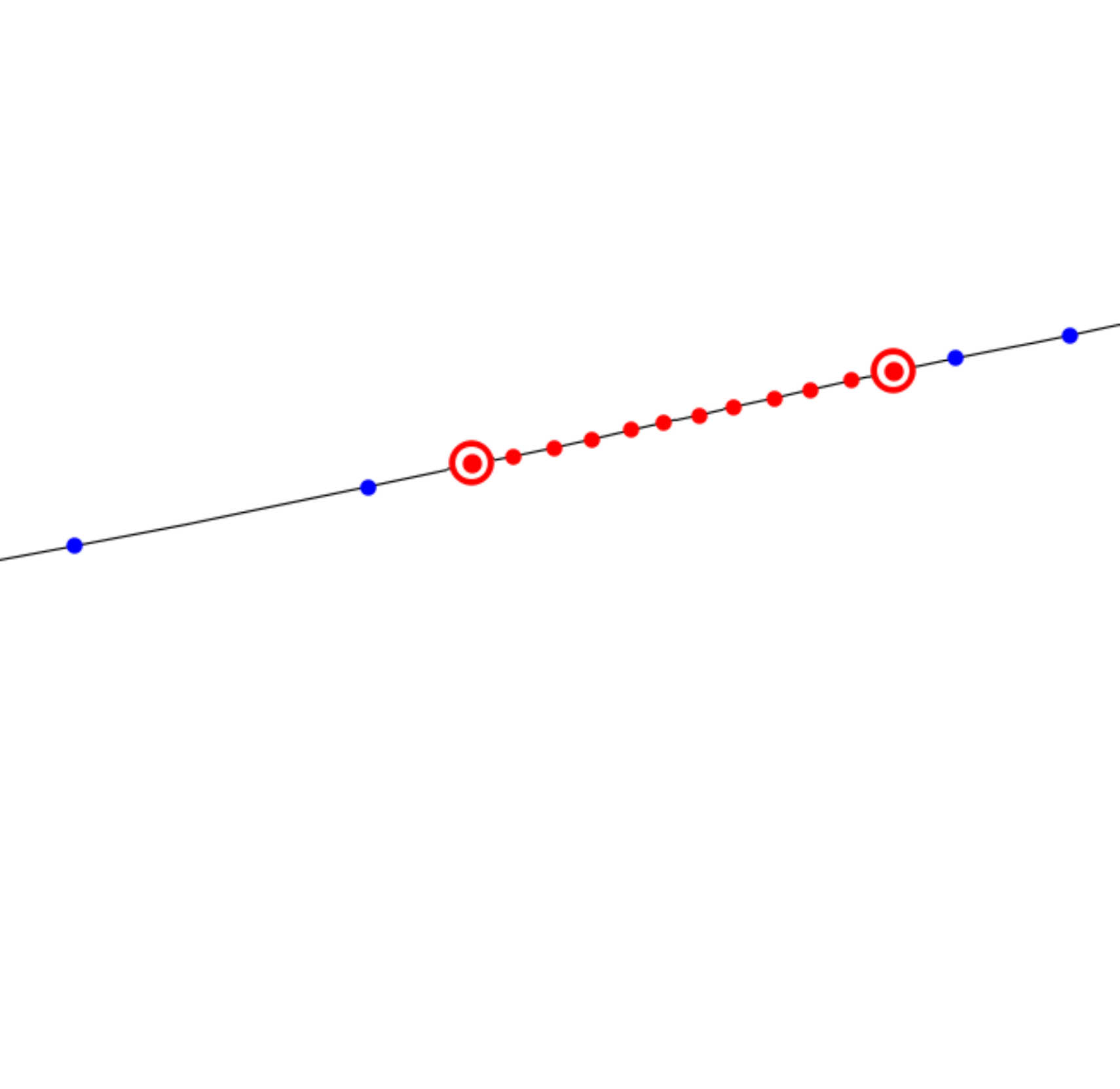}}}\label{slow_motion}
\subfigure[Smooth turn] {\fbox{\includegraphics[width=33mm]{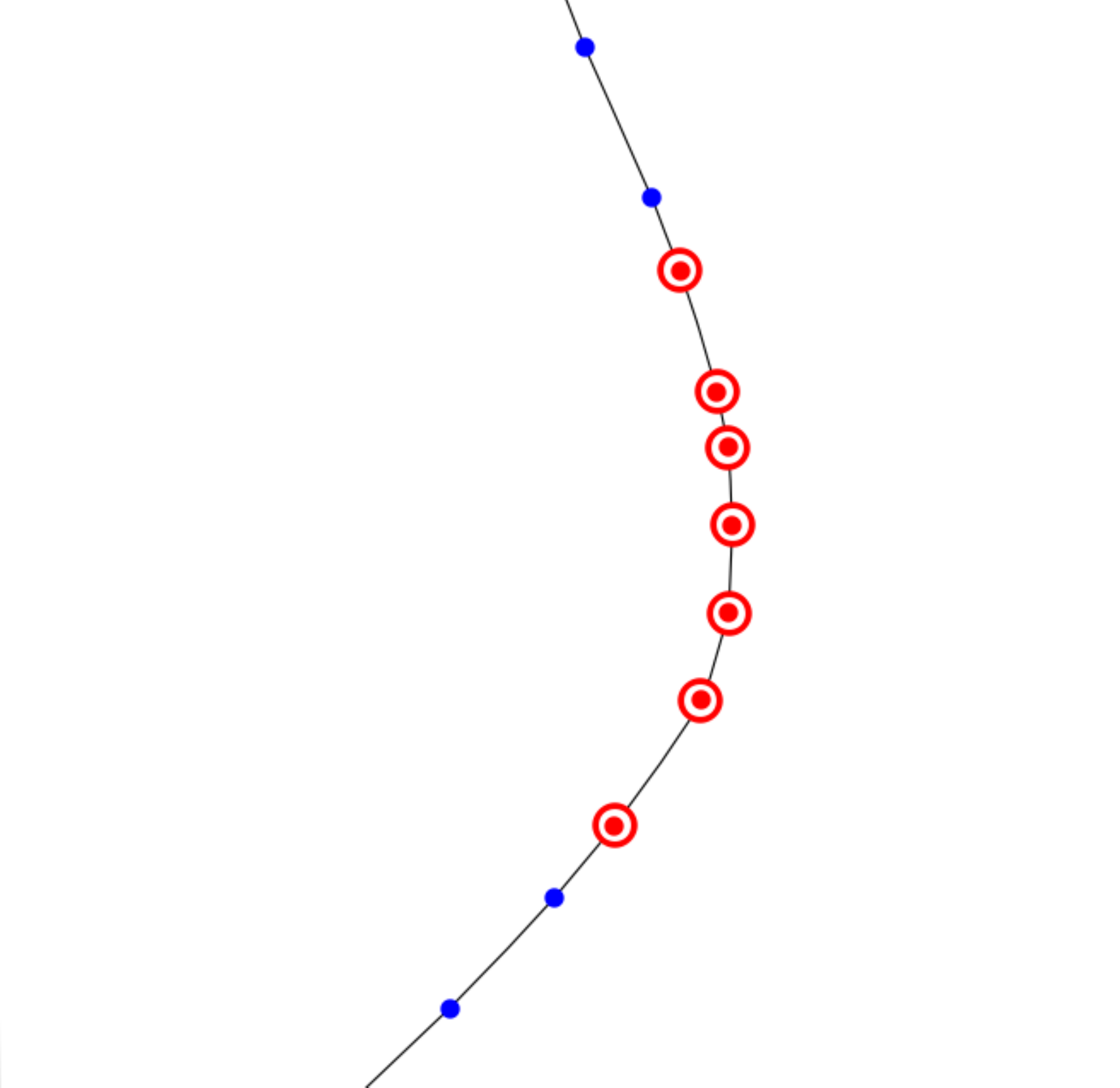}}}\label{smooth_turn}
\caption{Instantaneous and long-lasting trajectory events.} \label{longterm}
\end{figure}

Once potentially noisy positions are cleared, the {\em mobility tracker} can instantly deduce a variety of {\em instantaneous} events by examining the trace of each vessel alone:

\begin{enumerate}
\item[i)] {\em Pause} indicates that a vessel is temporarily halted, once its instantaneous speed $v_{now}$ is below a suitable threshold $v_{min}$. For example, if $v_{now}$ is currently less than $v_{min}=1$ knot, then the ship seems idle. For the vessel shown in Figure~\ref{longterm}(a), the red bullets indicate several pause events; apparently, the ship is anchored at the port and such small displacements may be due to GPS errors or sea drift.
\item[ii)] {\em Speed change} occurs once current $v_{now}$ deviates by more than $\alpha\%$ from the previously observed speed $v_{prev}$. Given a  threshold $\alpha$, the formula $ |\frac{v_{now}-v_{prev}}{v_{now}}| > \frac{\alpha}{100}$ indicates whether the vessel has just decelerated or accelerated. This normally happens when approaching to or departing from a port, as depicted in Figure~\ref{longterm}(b).
\item[iii)] {\em Turn} is spotted when heading in $\overrightarrow{v}_{now}$ has just changed by more than a given angle $\Delta\theta$; e.g., there is a difference of $\geq 15^o$ from the previous direction. 
\end{enumerate}

No critical point gets immediately issued upon detection of any such simple events. An instantaneous pause or turn may be serendipitous and is not meaningful out of context, because a series of such events may signify that the ship is stopped for some time. To avoid iteratively probing these locations later, we simply attach a {\em bitmap} to each stream tuple, using one bit for each particular instantaneous event. Note that multiple bits may be set at each location; e.g., the vessel may have just taken a sudden turn and also changed its speed above the respective thresholds, hence two distinct bits must be set to 1.

By buffering these instantaneous events within the window, we then can detect spatiotemporal phenomena of some duration. Examination of such {\em long-lasting trajectory events} is carried out in the following order. Note that if a certain long-lasting event has just been detected at a location, then checking if it also qualifies for another event is skipped altogether.

\begin{enumerate}
\item {\em Gap} in reporting is examined first. This event is spotted when a vessel has not emitted a message for a time period $\Delta T$, e.g., over the past 10 minutes. This may occur when the vessel sails in an area with no AIS receiving station nearby, or because the transmission power of its transponder allows broadcasting in a shorter range. Then, its course is unknown during this period, as it occurs between the two red bullets in Figure~\ref{longterm}(c). Reporting that contact was lost is important not only for online monitoring, but also for safety reasons, e.g., a suspicious move near maritime boundaries, or a potential intrusion of a tanker into a marine park. A pair of critical points signify when contact was lost ({\em gapStart} annotates the previously reported location) and when it was restored ({\em gapEnd} for current location).
\item Checking for a {\em long-term stop} is only fired if the vessel is noticed to move ($v_{now}>v_{min}$)  just after a pause. If current location is preceded by at least $m$ consecutive instantaneous pause or turn events  in the buffer, and they are all within a predefined radius $r$ (e.g., 250 meters), then a long-term stop is identified. In Figure~\ref{longterm}(d), the red points inside the circle succeed one another and indicate such immobility, so they are collectively approximated by a single critical point (their centroid) annotated as {\em stopped} with their total duration.
\item {\em Slow motion} means that a vessel consistently moves at very low speed ($\leq v_{min}$) over its $m$ most recent messages, as in Figure~\ref{longterm}(e). If those buffered positions have not already qualified as a long-term stop by the previous rule (because they did not fall inside a small circle), then they probably succeed each other slowly along a path. The first and the last of these positions are both reported as critical, respectively annotated as {\em lowSpeedStart} and {\em lowSpeedEnd}.
\item {\em Smooth turns} are examined last. Due to their large size and maritime regulations, vessels normally report a series of locations when they change course. By checking whether the cumulative change in heading over buffered previous positions exceeds a given angle $\Delta\theta$, a series of such critical {\em turning} points may be emitted, as illustrated with the red points in Figure~\ref{longterm}(f).
\end{enumerate}

Thus, critical points are emitted from each detected long-lasting trajectory event, and this relies heavily on efficient noise reduction (cf. Section~\ref{sec:noise}. For instance, an outlier breaking the subsequence of instantaneous pause events could prevent characterization of a long-term stop, and instead yield two successive such stops very close to each other. Moreover, in case that no long-lasting trajectory event was identified at this location, we check its associated bitmap with these additional rules:   

\begin{enumerate}
\item[5.] If the bit for `turn' is set, we check whether the current heading in $\overrightarrow{v}_{now}$ also deviates more than $\Delta\theta$ from mean velocity $\overrightarrow{v}_{m}$ of the vessel. If true, we emit a critical {\em turning} point. Note that this could possibly affect only raw AIS  locations that are not qualified as erroneous. In fact, noisy positions as those in Figure~\ref{noise}(c) have already been discarded by the Noise Reduction module.
\item[6.] If the bit for `speed change' is set (Figure~\ref{longterm}(b)) and current speed $v_{now}$ also deviates by more than $\alpha\%$ from the mean speed $v_m$ of this vessel, then a critical point must be emitted and annotated as {\em speedChange}. It signifies that this instantaneous event was not caused by fluctuations in the measured speed due to delayed messages, but that such change in speed is probably valid.   
\end{enumerate}

Clearly, this detection process can only lead to a {\em single annotation} for each critical point. For instance, if a vessel disappeared for long and is suddenly found anchored somewhere, this event will be spotted either as a gap or a stop, but not both. We have deliberately chosen such a `crisp' classification allowing a single characterization per detected point, as our goal is to achieve a concise trajectory representation, by dropping superfluous locations. In future work, we plan to introduce a fuzzy, probabilistic scheme of multiple annotations per critical point at diverse confidence margins.

Each critical point is issued along with a {\em velocity vector} (comprising instantaneous speed and heading), as an indicator of the short-term course of that particular vessel. This measurement may be useful for further analysis, e.g., in order to identify complex maritime events as explained  in Section \ref{sec:cer}.

\begin{figure}[t]
\centering
{\fbox{\includegraphics[width=10cm]{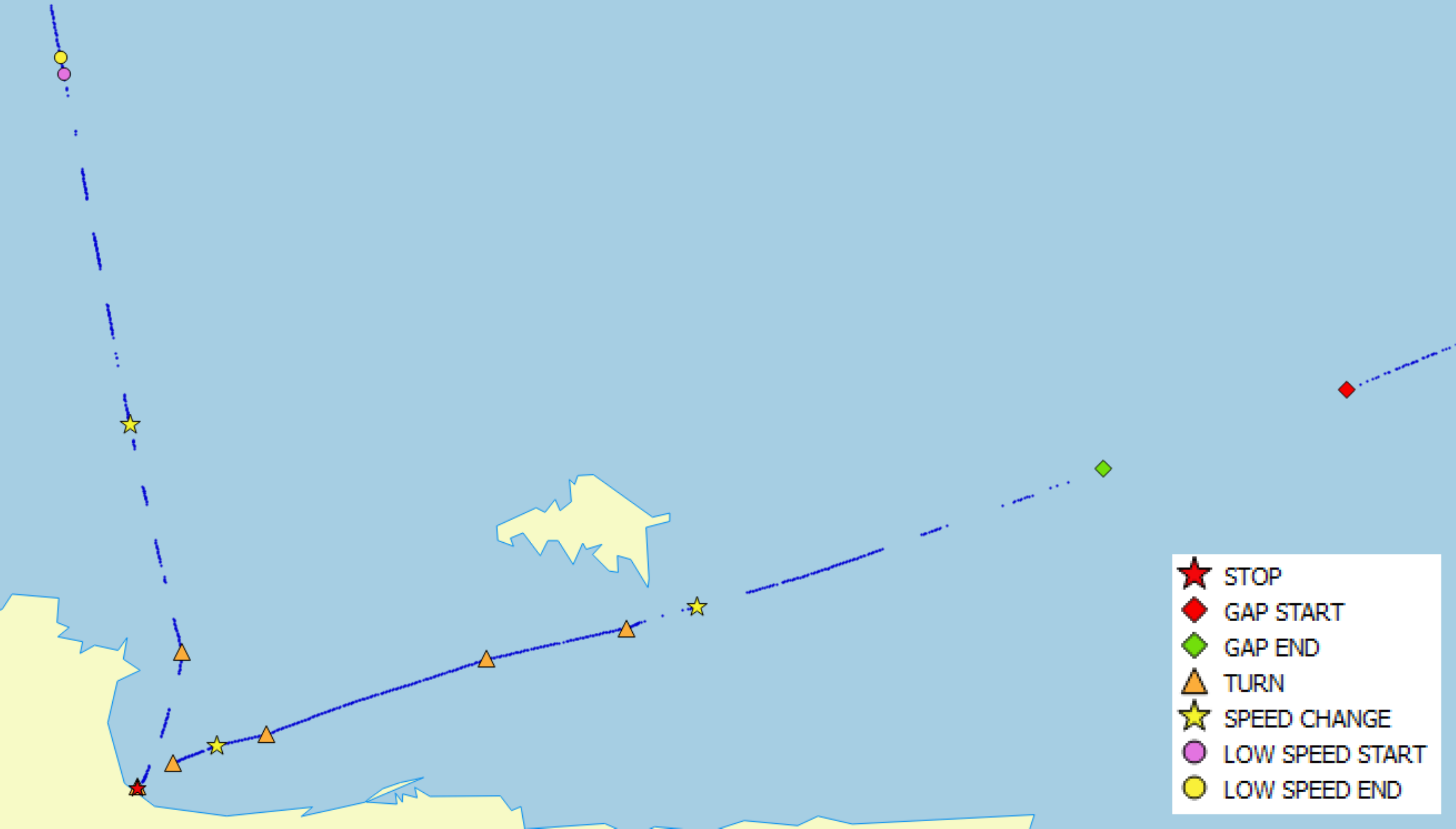}}}
\caption{Critical points identified along a vessel trajectory (raw AIS positions are shown as blue dots).} \label{trajectories}
\end{figure}

The example trajectory in Figure~\ref{trajectories} illustrates the data compression gains achieved when retaining critical points only. Obviously, such filtering greatly depends on proper choice of parameter values, which is a trade-off between reduction efficiency and approximation accuracy. For a suitable calibration of these parameters, apart from consulting maritime domain experts (our partners in the AMINESS project), we have also conducted several exploratory tests on randomly chosen vessels from AIS data in the Aegean Sea. For instance, setting $\Delta\theta = 5^o$ instead of $\Delta\theta = 15^o$ may even double the amount of critical points, because more raw AIS locations would qualify as turning points due to sea drift and discrepancies in GPS signals. Since our analysis is mostly geared towards data reduction, for our empirical study (Section~\ref{sec:empirical_trajectory}) we have chosen an aggressive parametrization (values listed in bold in Table~\ref{tbl:calibration}), which yields quite tolerable accuracy. With more relaxed parameter values, additional events can be detected, capturing slighter changes along each trajectory.

The complexity for detecting instantaneous events and communication gaps is $O(1)$ per incoming positional tuple, since only the two latest locations are examined per vessel. The cost for the remaining long-lasting events is $O(b)$, where $b$ is the number of buffered positions that need inspection by each such rule. This may involve just a few points in case of smooth turns, but $b$ may be higher (sometimes, a few hundred) if a series of positions qualify for  a stop or slow motion. However, we stress that checks for stop of slow motion are fired rather infrequently, only once a vessel starts moving after a certain period of idleness. Thus, the overall cost is more than affordable and detection is near real-time, as we empirically verified (Section~\ref{sec:empirical_trajectory}).

By taking advantage of those online annotations at critical points along trajectories, lightweight, succinct {\em synopses} can be retained per vessel over the recent past. Then, the {\em compressor} module simply evicts point locations that have not been detected as critical. Instead of resorting to a costly simplification algorithm, we opt to reconstruct vessel traces approximately from already available critical points. This summarization depends on the annotation of detected trajectory events (stop, turn, gap, etc.), so as to refresh each trajectory accordingly. This main-memory process affects trajectory portions currently within the sliding window. Of course, resulting synopses may be also archived via the {\em Trajectory Exporter} module for offline use as files (e.g., KML, CSV) or in a database by incrementally emitting {\em ``delta'' batches} of critical points as they get identified at each slide of the window.

The aforementioned rules for detecting trajectory events are suitably defined in the mobility tracker, which allows fast, in-memory maintenance of movement features. Note that additional events can be detected by simply enhancing the mobility tracker with extra conditions. In future work, we plan to complement this methodology so as to capture more features, such as traveled distance from a given origin (e.g., a port). Nonetheless, even with this set of filters, we can figure out the mutability in each trajectory and distinctly characterize its course across time. Most importantly, these spatiotemporal features can serve as a basis to recognize more complex maritime events, as we discuss next.

\section{Complex Event Recognition}\label{sec:cer}

The trajectory detection module compresses a vessel position stream to a stream of critical events, including the instantaneous events $\mathit{gapStart}$ and $\mathit{gapEnd}$, indicating communication gaps, $\mathit{lowSpeedStart}$, $\mathit{lowSpeedEnd}$, $\mathit{speedChange}$ and $\mathit{turn}$, and the durative event $\mathit{stopped}$. Each such event is accompanied by the coordinates and velocity (speed and heading) of the corresponding vessel. This data stream, hereafter Movement Event (ME) stream, is transmitted to the complex event (CE) recognition module, which combines it with the locations of ports and protected areas, in order to recognize potentially suspicious or dangerous maritime situations, for the benefit of marine authorities.

The CE recognition module is based on the `Event Calculus for Run-Time reasoning' (RTEC)~\cite{DBLP:journals/tkde/ArtikisSP15}. The Event Calculus \cite{kowalski86} is a logic programming action language. RTEC has a formal, declarative semantics---CE patterns in RTEC are (locally) stratified logic programs \cite{local-strat}. In contrast, almost all complex event processing languages, including \cite{brenna07, li05, agrawal08}, and several data stream processing languages, such as ESL \cite{bai06} that extends CQL \cite{arasu06}, lack a rigorous, formal semantics \cite{cugola10}. Reliance on informal semantics constitutes a serious limitation for maritime monitoring, where validation and traceability of the effects of events are crucial. Moreover, the semantics of event query languages and production rule languages often have an algebraic and less declarative flavor \cite{eckert10, paschke09ruleml}.
In the following sections we present RTEC and illustrate its use for maritime monitoring. 
 
\subsection{Event Calculus for Run-Time reasoning}

RTEC has a linear temporal model including integer time-points. Following Prolog's convention, variables start with an upper-case letter, while predicates and constants start with a lower-case letter. For a \emph{fluent} $F$---a property that is allowed to have different values at different points in time---the term $F \val V$ denotes that fluent $F$ has value $V$. 
$\holdsFor(F \val V, I)$ denotes that $I$ is the list of the maximal intervals for which $F\val V$ holds continuously. RTEC is interval-based and thus avoids the related logical problems of time-point-based event processing approaches (see \cite{paschke05} for a discussion of these problems).
$\holdsAt(F \val V, T)$ represents that fluent $F$ has value $V$ at some time-point $T$. \holdsAt\ and \holdsFor\ are defined in such a way that, for any fluent $F$, \linebreak\holdsAt$(F \val V, T)$ if and only if $T$ belongs to one of the maximal intervals of $I$ for which \holdsFor$(F \val V, I)$. 

An {\em event description} in RTEC includes rules that define the \emph{event instances} with the use of the \happensAt\ predicate, the \emph{effects of events} with the use of the \initiatedAt\ and \terminatedAt\ predicates, and the \emph{fluent values} with the use of the \holdsAt\ and \holdsFor\ predicates, as well as other, possibly atemporal, constraints. 
Table~\ref{tbl:ec} presents a fragment of the predicates available to the event description developer. 

\begin{table}[t]
\caption{RTEC Predicates.}\label{tbl:ec}
\begin{center}
\renewcommand{\arraystretch}{0.5}
\setlength\tabcolsep{3.6pt}
\begin{tabular}{ll}
\hline\noalign{\smallskip}
\multicolumn{1}{c}{\textbf{Predicate}} & \multicolumn{1}{c}{\textbf{Meaning}}  \\
\noalign{\smallskip}
\hline
\noalign{\smallskip}
\holdsAt$(F \val V,\ T)$ & The value of fluent $F$ is $V$ at time $T$\\[4pt]

\holdsFor$(F \val V,\ I)$ & $I$ is the list of the maximal intervals for which $F\val V$ holds continuously\\[4pt]

\happensAt$(E,\ T)$ & Event $E$ occurs at time $T$  \\[4pt]                           
                           
\initiatedAt$(F \val V,\ T)$ & At time $T$ a period of time for which $F\val V$ is initiated \\[4pt]

\terminatedAt$(F \val V,\ T)$ & At time $T$ a period of time  for which $F\val V$ is terminated \\[4pt]


\intersectall$\mathit{(L, I)}$ & $\mathit{I}$ is the list of maximal intervals produced by the intersection of \\
& the lists of maximal intervals of list $\mathit{L}$\\

\hline
\end{tabular}
\end{center}
\end{table}


For a fluent $F$, $F\val V$ holds at a particular time-point $T$ if $F \val V$ has been \emph{initiated} by an event that has occurred at some time-point earlier than $T$, and has not been \emph{terminated} at some other time-point in the meantime. This is an implementation of the \emph{law of inertia}. To compute the \emph{intervals} $I$ for which $F\val V$, i.e. $\holdsFor(F \val V, I)$, we find all time-points $T_s$ at which $F\val V$ is initiated, and then, for each $T_s$, we compute the first time-point $T_f$ after $T_s$ at which $F\val V$ is terminated. As an example, consider the formulation below:
%
\begin{align} 
& \label{eq:gap}
\begin{mysplit}
\initiatedAt\mathit{(gap(Vessel)\val\true,\ T) \leftarrow} \\
\qquad	\happensAt\mathit{(gapStart(Vessel),\ T),} \\
\qquad	\holdsAt\mathit{(coord(Vessel)\val(Lon,Lat),\ T),} \\
\qquad  \nbf\ \mathit{nearPorts(Lon,Lat)}\\
\terminatedAt\mathit{(gap(Vessel)\val\true,\ T) \leftarrow} \\
\qquad	\happensAt\mathit{(gapEnd(Vessel),\ T)}
\end{mysplit}
\end{align}
$\mathit{gap(Vessel)}$ is a Boolean fluent denoting a communication gap for some $\mathit{Vessel}$, i.e.~the $\mathit{Vessel}$ stops transmitting AIS messages. In some cases, the absence of AIS messages is suspicious and thus we need to record it. $\mathit{gapStart(Vessel)}$ and\linebreak $\mathit{gapEnd(Vessel)}$ are instantaneous MEs indicating, respectively, the time-points in which a $\mathit{Vessel}$ stops and resumes sending AIS messages. $\mathit{coord}$ is a fluent reporting the coordinates of a vessel. Like MEs, this type of information is provided by the trajectory detection module. $\mathit{nearPorts(Lon,Lat)}$ is an atemporal predicate that becomes true when the point $\mathit{(Lon,Lat)}$ is close to a port. `\nbf' is negation-by-failure \cite{clark78}.   
Rule-set \eqref{eq:gap} states that $\mathit{gap(Vessel)\val\true}$ is initiated if the trajectory detection module reports a $\mathit{gapStart}$ ME for the $\mathit{Vessel}$, and the $\mathit{Vessel}$ is far from the ports of the area under surveillance. 
Given rule-set \eqref{eq:gap}, RTEC computes the list of maximal intervals during which $\mathit{gap(Vessel)\val\true}$ holds continuously.

\subsection{Spatial Indexing}
\label{subsec:spatial-pre}

CE recognition for maritime surveillance requires various types of spatial operation \cite{Garcia11, Laere09}. For instance, we need to determine whether a point---a vessel's location---lies inside a polygon indicating an area of interest, such as a protected area, or whether it is near another point, such as a port. Moreover, we need to detect the vessels that are in close proximity (heading towards each other). In our approach to CE recognition, the availability of the full power of logic programming is one of the main attractions of employing RTEC as the temporal formalism. It allows CE patterns to include not only temporal constraints but also (complex) atemporal constraints. Recall e.g.~the use of the atemporal predicate $\mathit{nearPorts}$ in the specification of communication $\mathit{gap}$ in rule-set \eqref{eq:gap}. This is in contrast to various state-of-the-art CE recognition approaches, such as \cite{dousson07, kramer09, cugola10, DBLP:conf/sigmod/ZhangDI14}, which support very limited atemporal reasoning, thus being unsuitable for maritime monitoring.

\begin{figure}[t]
\centering
{\fbox{\includegraphics[height=.35\textwidth]{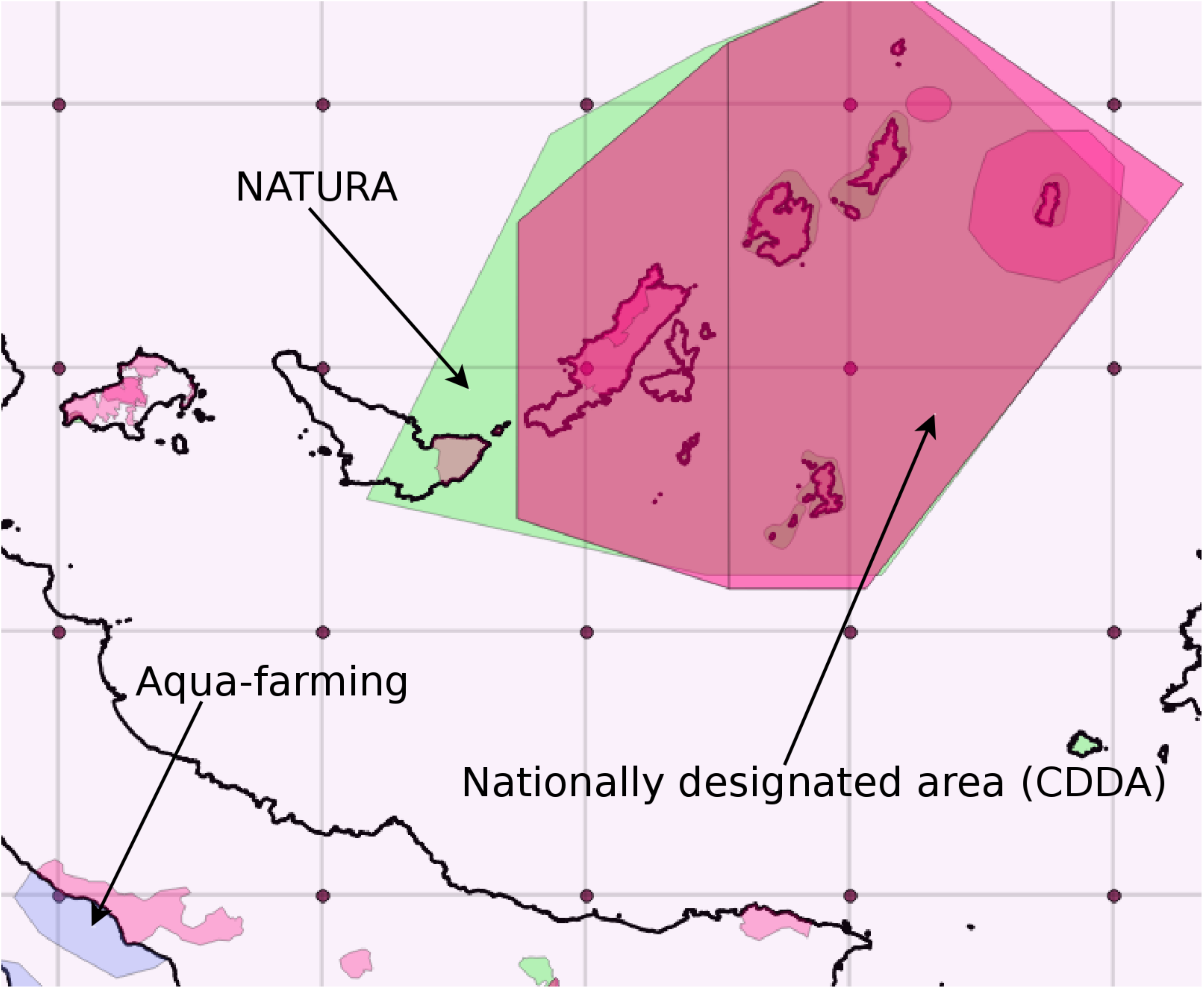}}}
\caption{Grid partitioning: different polygon types indicate different types of (overlapping) area of interest.} 
\label{fig:grid}
\end{figure}

For efficient spatial reasoning, we adopt a grid partitioning scheme which divides the surveillance area into equally sized cells (see Figure \ref{fig:grid}). Each area of interest and port is assigned only to those cells with which it overlaps. This assignment is performed off-line and provided as background knowledge to RTEC. The use of a grid enables us to quickly determine, through a simple calculation on the coordinates, the cell inside which a vessel is located. The task of determining each vessel's cell is performed before each CE recognition query. This way, we can efficiently compute the number of vessels in close proximity and check whether a vessel is inside an area of interest, by performing calculations (e.g.~using the ray crossings algorithm \cite{orourke_computational_1998} for determining whether a point lies inside a polygon) \emph{only} for those vessels/areas in the same or adjacent cells.

\subsection{Specifying Complex Maritime Events}
\label{subsec-scenarios}

Given the critical ME stream produced by the trajectory detection module, and a set of protected areas, RTEC recognizes a set of CEs for the benefit of maritime authorities. The choice of CEs and their patterns were specified in collaboration with the domain experts of the AMINESS project. Below we present a fragment of our CE patterns. The complete list may be found in \cite{alevizos15}.

\textbf{Suspicious Vessel Delay.} 
Some vessels, such as those passing through protected areas in order to minimize trip length 
and fuel consumption, switch off their transmitters and stop sending position signals. 
But sailing through a protected area is not the only reason for switching off an AIS transmitter. To investigate the behavior of vessels during a communication gap, we formulated the CE below:
\begin{align} 
& \label{eq:suspicious-delay}
\begin{mysplit}
\holdsFor\mathit{(suspiciousDelay(Vessel)\val\true,\ I)} \leftarrow \\
\qquad    \holdsFor\mathit{(gap(Vessel)\val\true,\ I_{gap}),}\\
\qquad 	  \mathit{extendedDelays(Vessel,\ I_{gap},\ I)}
\end{mysplit}
\end{align}
Recall that $I$ in $\holdsFor(F \val V, I)$ is the list of the maximal intervals for which $F\val V$ holds continuously (see Table \ref{tbl:ec}). $\mathit{I_{gap}}$ in \holdsFor$\mathit{(gap(Vessel)\val\true,\ I_{gap})}$, therefore, is the list of maximal intervals during which a $\mathit{Vessel}$ stops transmitting AIS signals while at open sea (see rule-set \eqref{eq:gap} for the $\mathit{gap}$ fluent).
$\mathit{extendedDelays(Vessel, I', I)}$ selects the maximal intervals $I$ of the list $\mathit{I'}$ for which the highest possible speed of the $\mathit{Vessel}$ is below a threshold. 
We estimate the highest possible speed of a vessel in a simplified way: we assume that the vessel moved along a straight line from the point of $\mathit{gapStart}$ to that of $\mathit{gapEnd}$. Under this assumption, its speed cannot have been greater than the one determined by dividing this shortest path by the time spent to travel it. 
Rule \eqref{eq:suspicious-delay} thus states that a very low vessel speed combined 
with a communication gap occurring at open sea is to be treated as a suspicious delay.

A more refined implementation would estimate the highest possible speed of a vessel during a communication gap even when it is impossible (due to e.g.~terrestrial areas) or unlikely (due to weather conditions) to sail along a straight line. 


\textbf{Vessel Rendezvous.}  
`Suspicious delay' allows us to define additional types of suspicious activity; consider the rule below:
\begin{align} 
& \label{eq:possible-rendezvous}
\begin{mysplit}
\holdsFor\mathit{(possibleRendezvous(Vessel_{1},Vessel_{2})\val\true,\ I)} \leftarrow \\
\qquad    \holdsFor\mathit{(in(Vessel_{1}, Cell)\val\true,\ I_1),}\\
\qquad    \holdsFor\mathit{(in(Vessel_{2}, Cell)\val\true,\ I_2),}\\
\qquad    \holdsFor\mathit{(suspiciousDelay(Vessel_{1})\val\true,\ I_3),}\\
\qquad    \holdsFor\mathit{(suspiciousDelay(Vessel_{2})\val\true,\ I_4),}\\
\qquad 	  \intersectall\mathit{([I_{1},I_{2},I_{3},I_{4}],\ I)}
\end{mysplit}
\end{align}
$\mathit{in(Vessel, Cell)}$ indicates the $\mathit{Cell}$ of the grid in which the $\mathit{Vessel}$ is located. The value of this fluent is set prior to each CE recognition query, 
as described in Section \ref{subsec:spatial-pre}. $\intersectall$ is a built-in RTEC predicate which calculates the intersection of a list of lists of maximal intervals (see Table \ref{tbl:ec}). According to rule \eqref{eq:possible-rendezvous}, if two vessels simultaneously exhibit a $\mathit{suspiciousDelay}$ and are located in the same area, then this could indicate that they had arranged for a rendezvous. 
Note that, since we do not have information about the vessels' locations during communication gaps, the above rule cannot capture the precise place and time of the rendezvous, if any. 

Maritime activities form hierarchies, in the sense that the formulation of one activity is also used to define other, higher-level activities. E.g.~vessel rendezvous is specified in terms of suspicious vessel delay. In contrast to many state-of-the-art CE recognition systems, such as Esper\footnoteremember{esper}{\url{http://www.espertech.com/esper/}} and SASE\footnote{\url{http://sase.cs.umass.edu/}}, RTEC can naturally express hierarchical knowledge by means of well-structured specifications.

\textbf{Fast Approach.} 
Another dangerous situation may arise when a vessel is rapidly moving towards some other vessel(s). 
Such a behavior could indicate a vessel pursuit or even imminent collision. Consider the formalization below:
\begin{align} 
& \label{eq:fast-approach}
\begin{mysplit}
\happensAt\mathit{(fastApproach(Vessel),\ T)} \leftarrow \\
\qquad    \happensAt(\mathit{speedChange(Vessel),\ T),}\\
\qquad	  \holdsAt\mathit{(velocity(Vessel)\val Speed,\ T),} \\
\qquad 	  \mathit{Speed > 20\ knots},\\
\qquad	  \holdsAt\mathit{(coord(Vessel)\val (Lon,Lat),\ T),} \\
\qquad    \nbf\ \mathit{nearPorts(Lon,Lat),}\\
\qquad    \holdsAt\mathit{(headingToVessels(Vessel)\val\true,\ T)}
\end{mysplit}
\end{align}
$\mathit{fastApproach(Vessel)}$ and $\mathit{speedChange(Vessel)}$ are instantaneous CE and ME respectively. $\mathit{velocity}$ is a fluent indicating the speed of a vessel. This information, as well as a vessel's heading, is provided by the trajectory detection module and accompanies every detected ME. $\mathit{headingToVessels(Vessel)}$ is a fluent that becomes true whenever a $\mathit{Vessel}$'s direction of movement is towards at least one nearby vessel. According to rule \eqref{eq:fast-approach}, a `fast approach' is recognized when a $\mathit{Vessel}$ changes its speed at open sea, the new speed is above 20 knots, and there is at least one other nearby vessel towards which it is heading. The value of 20 knots was chosen by domain experts. 




\textbf{Package Picking.}  
Another possible interaction between two vessels is when one of them drops a package at some area and another vessel appears later in order to pick it up. One way of formulating this type of interaction is the following:
\begin{align} 
& \label{eq:package-picking}
\begin{mysplit}
\happensAt\mathit{(possiblePicking(Vessel_1,Vessel_2),\ T_{pick})} \leftarrow \\
\quad    \happensAt\mathit{(\endE(stopped(Vessel_1)\val\true), T_{drop}),}\\
\quad    \holdsAt\mathit{(in(Vessel_1)\val Cell,\ T_{drop}),}\\
\quad    \happensAt\mathit{(\startE(stopped(Vessel_2)\val\true), T_{pick}),}\\
\quad    \holdsAt\mathit{(in(Vessel_2)\val Cell,\ T_{pick}),}\\
\quad    \mathit{T_{pick} - T_{drop} < 1 \ hour,}\\
\quad    \holdsAt\mathit{(coord(Vessel_1)\val(Lon_1,Lat_1),\ T_{drop}),}\\
\quad    \holdsAt\mathit{(coord(Vessel_2)\val(Lon_2,Lat_2),\ T_{pick}),}\\
\quad    \mathit{distance((Lon_1,Lat_1),(Lon_2,Lat_2),\ Dist),}\\
\quad    \mathit{Dist < 0.5 \ km}
\end{mysplit}
\end{align}
$\mathit{stopped(Vessel)}$ is a Boolean fluent indicating that a $\mathit{Vessel}$ has stopped at open sea. The definition of this fluent is based on the information provided by the trajectory detection module, which reports the list of maximal intervals during which a vessel has stopped. From this list, we keep only those intervals where the vessel is not in any port. 
$\startE(F\val V)$ (respectively $\endE(F\val V)$) is a built-in RTEC event taking place at each starting (ending) point of each maximal interval for which $F\val V$ holds continuously. 
Thus e.g.~$\startE\mathit{(stopped(Vessel)\val\true)}$ takes place at the starting point of each maximal interval for which the $\mathit{Vessel}$ has stopped at at open sea. Rule \eqref{eq:package-picking} describes a scenario where a vessel had stopped at some area and started moving at time $\mathit{T_{drop}}$, then, after no more than an hour, another vessel arrived and stopped at the same area, and the Haversine distance between the two stop locations, as calculated by the $\mathit{distance}$ predicate, was no more than half a kilometer. 

\subsection{Recognizing Complex Maritime Events}\label{sec:caching}

\begin{figure}%
\centering
\includegraphics[width=0.5\columnwidth]{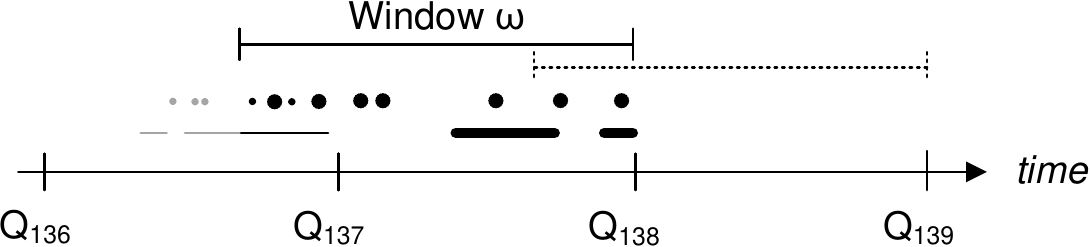}%
\vspace{0.5em}
\caption{CE recognition in RTEC.}%
\label{fig:window_example}%
\end{figure}



RTEC performs CE recognition by means of continuous query computation. At each query time $Q_i$, the MEs that fall within a specified sliding window $\omega$ are taken into consideration. All MEs that took place before or at $Q_i{-}\omega$ are discarded. 
At $Q_i$, the CE intervals computed by RTEC are those that can be derived from MEs that occurred in the interval $(Q_{i}{-}\omega, Q_{i}]$, as recorded at time $Q_i$. When the range $\omega$ is longer than the slide step $\beta$, it is possible that an ME occurs in the interval $(Q_{i}{-}\omega, Q_{i-1}]$ but arrives at RTEC only after $Q_{i-1}$; its effects are taken into account at query time $Q_i$. This is illustrated in Figure~\ref{fig:window_example}. Occurrences of MEs are displayed as dots line segments. For CE recognition at $Q_{138}$, only the events marked in black are considered, whereas the greyed out events are neglected. Assume that all events marked in bold arrived only after $Q_{137}$. Then, we observe that two MEs were delayed, i.e., they occurred before $Q_{137}$, but arrived only after $Q_{137}$. In our setting, the window range $\omega$ is larger than the slide step. Hence, these events are not lost but considered as part of the recognition process at $Q_{138}$. Further details about the reasoning engine of RTEC may be found at \cite{DBLP:journals/tkde/ArtikisSP15}.

\section{Empirical Evaluation}
\label{sec:empirical}

Our maritime surveillance system has a modular design with loosely coupled components.
The mobility tracker for online trajectory detection\footnote{\scriptsize Source code is publicly available at \url{http://www.dblab.ece.ntua.gr/~kpatro/tools/streamAIS/}.} is developed in GNU C++ and runs entirely on main memory for efficiently coping with massive, volatile, streaming locations.
RTEC\footnote{\scriptsize  \url{https://github.com/aartikis/RTEC}.}, the complex event (CE) recognition component, is implemented in Prolog\footnote{\scriptsize The patterns of the complex maritime events are  available at \url{http://users.iit.demokritos.gr/~a.artikis/aminess.tar.gz}.}. 

We conducted experiments against a real AIS dataset containing 23GB of AIS messages spanning from 1~June~2009 to 31~August~2009 for $N\val$ 6,425 vessels in the Aegean, the Ionian, and part of the Mediterranean Sea. Not all vessels were actually on the move at all times, since a considerable part (chiefly cargo ships) were just passing by, and thus  tracked for a limited period (days or even hours). But most vessels were frequently sailing, e.g., passenger ships or ferries to the islands. When decoded and cleaned from corrupt messages, the dataset yielded 168,240,595 raw timestamped positions\footnote{\scriptsize This anonymized dataset (for privacy, each original MMSI has been replaced by a sequence number) is publicly available at \url{http://chorochronos.datastories.org/?q=content/imis-3months}}.

\begin{table}[t]
\centering\caption{Experimental settings.} \label{tbl:simulation}
\begin{scriptsize}
\begin{tabular}{|l|c|}
\hline
\textbf{Parameter} & \textbf{Value} \\
\hline
\hline Vessel count $N$ & {\bf 6,425}; 128,000; 1,280,000 \\
\hline Window range $\omega$ & 10min; 1h; 2h; {\bf 6h}; 9h; 24h \\
\hline Window slide $\beta$ & {\pbox{20cm}{1min; 5min; 10min; 15min; \\ 20min; 30min; {\bf 1h}; 90min; 2h; 4h}} \\
\hline Position stream rate $\rho$ (positions/sec) & {\bf original}; 1K; 2K; 5K; 10K \\
\hline Protected areas & 3,966 polygons with 78,418 edges \\
\hline
\end{tabular}
\end{scriptsize}
\end{table}

\begin{table}[t]
\centering \caption{Mobility tracking parameters.} \label{tbl:calibration}
\begin{scriptsize}
\begin{tabular}{|l|c|}
\hline
\textbf{Parameter} & \textbf{Value} \\
\hline
\hline Minimum speed $v_{min}$ for asserting movement & 1 knot ($\cong$1.852 km/h) \\
\hline Maximum rate $\alpha$ of speed change between successive locations & 25\% \\
\hline Minimum gap period $\Delta T$ (minutes) & 5; {\bf 10}; 15; 30 ; 60 \\
\hline Turn threshold $\Delta\theta$ (degrees) & $2^o$; $3^o$; $5^o$; $10^o$; {\bf 15$^o$}; $20^o$\\
\hline Radius $r$ to determine long-term stops & 250 meters\\
\hline Minimal number $m$ of inspected positions & 10 \\
\hline
\end{tabular}
\end{scriptsize}
\end{table}

We simulated a streaming behavior by consuming this positional data little by little, i.e., reading small chunks periodically according to window specifications. We examine sliding windows with varying ranges $\omega$ and slide steps $\beta$ based on timestamps from the original AIS messages. 
Thus, we replay this stream and the window keeps in pace with the  reported timestamps and not the actual time of each simulation. The arrival rate of positions is fluctuating throughout this 3-month period and varies widely among vessels; none of them reports at a fixed frequency, whereas there are ships inactive for large intervals. If we only consider the activity period of each vessel (i.e., when it actually relays positions, either moving or not), then it reports every two minutes on average, which translates into a mean arrival rate $\rho\approx$50 positions/sec from the entire fleet. For consistency with the real-world scenario, we consume the original stream ``as is'' in some simulations, even though this is a very low rate for a streaming application.
Moreover, we performed additional experiments at artificially increased rates so as to stress test our system and verify its efficiency and robustness. 
For the CE recognition component, the artificially enlarged datasets include 1,2M vessels and 3,2B MEs.
The simulation settings are listed in Table~\ref{tbl:simulation}, whereas the calibrated settings for online mobility tracking are given in Table~\ref{tbl:calibration}; default values are shown in bold.

Next, we report indicative results from these experiments. The trajectory event detection component operated on a server running Debian Linux ``Wheezy'' 7.5 amd64 with 48GB of RAM and two Intel Xeon X5675 processors at 3.07GHz. The CE recognition component RTEC operated on a computer with Intel Xeon CPU E5-2630 v2@2.60GHz$\times$12 processors and 256GB RAM, running Ubuntu Linux 14.04 and SWI Prolog 7.2.1.

\subsection{Assessment of Trajectory Detection}
\label{sec:empirical_trajectory}

\subsubsection{Performance of online mobility tracking} 

\begin{figure*}[t]
\begin{minipage}[h]{1\linewidth}
\centering
\subfigure[Small window ranges]
{\includegraphics[width=0.48\textwidth]{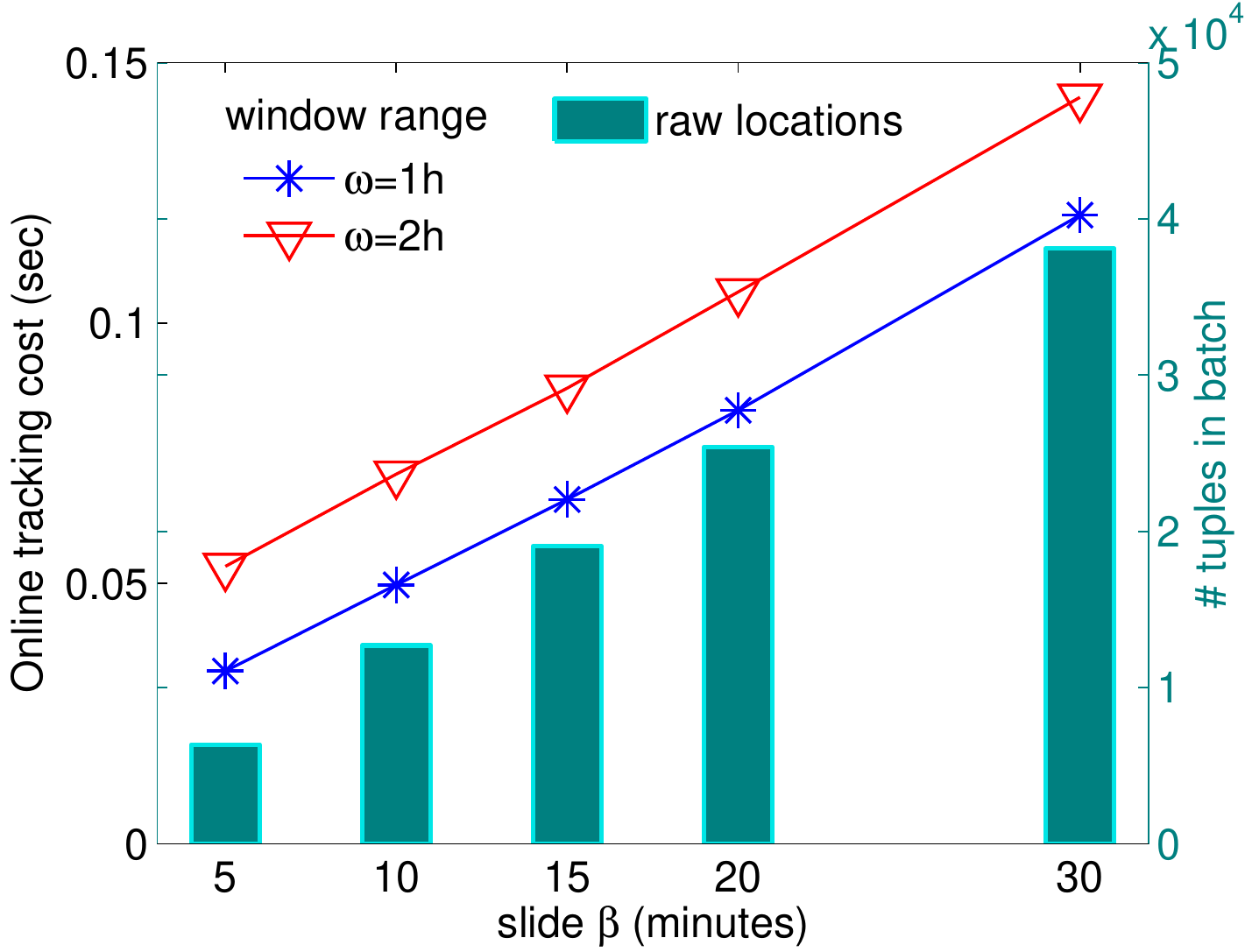}}
\subfigure[Large window ranges]
{\includegraphics[width=0.48\textwidth]{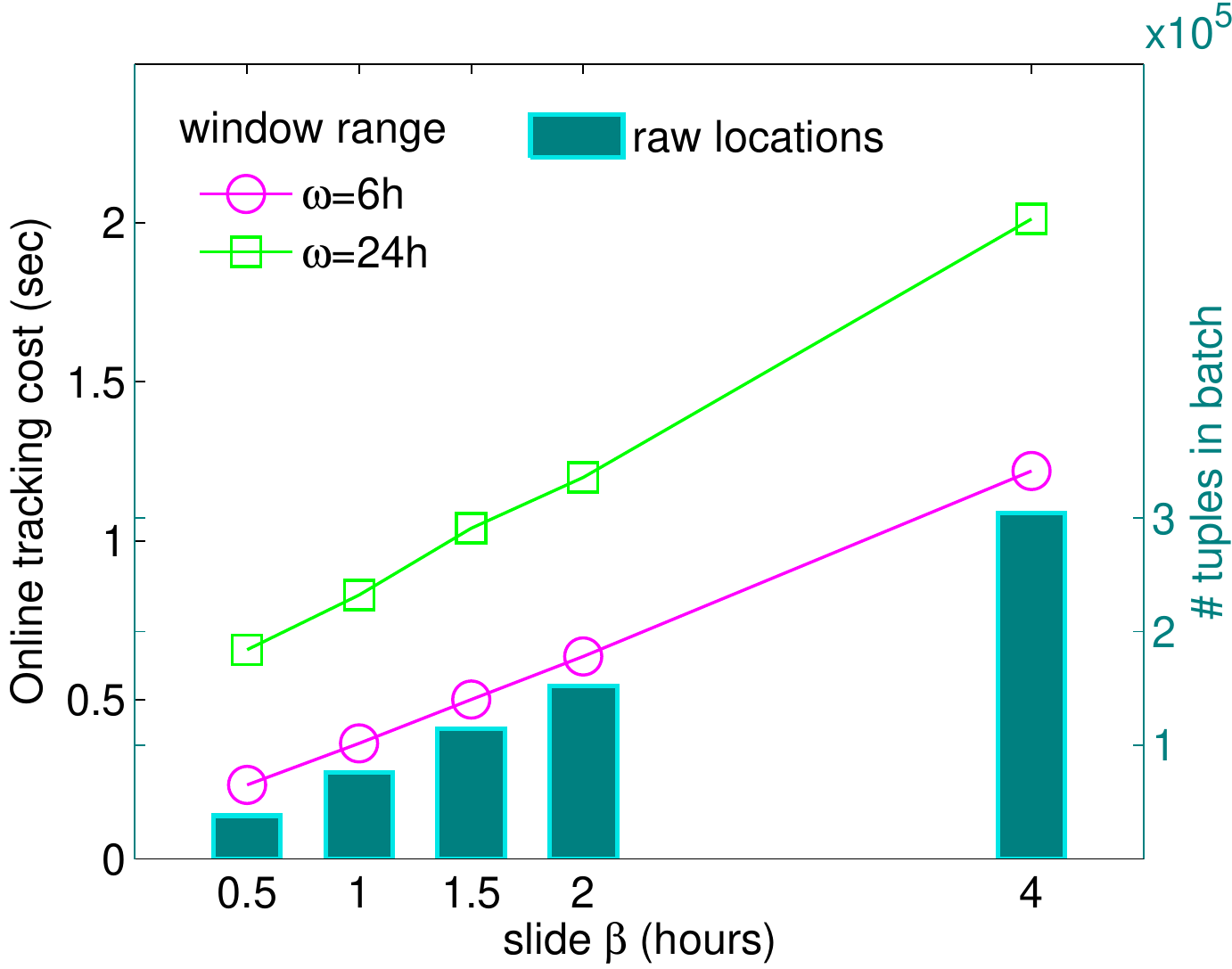}}
\caption{Online mobility tracking cost per window slide.} \label{graph:traj_win}
\end{minipage}
\end{figure*}

First, we examine performance of online detection concerning trajectory movement events using simulations at the original arrival rate. These experiments have been performed employing window specifications with varying ranges $\omega$ and slide steps $\beta$, and we measure the total time it takes to update a window with a fresh batch of raw AIS locations, evict expired ones, detect trajectory events, and report critical points. Then, we calculate averages of these time values over the total count of window instantiations, hence obtaining the per slide cost for window maintenance and identification of any trajectory events therein. Figure~\ref{graph:traj_win} plots this average execution cost per window for monitoring the entire fleet.

From Figure~\ref{graph:traj_win}(a) it turns out that our mobility tracker provides results instantly for smaller $\omega$ up to 2 hours. In the worst case,  it takes less than 150ms to track down any critical points per incoming batch of raw positional tuples. Quite expectedly, the cost grows linearly with an increasing slide $\beta$, as the window slides forward less often and thus each batch contains more input locations (illustrated with bars in this plot) directly proportional to the sliding step $\beta$. 

For larger windows with range $\omega$ up to 24 hours shown in Figure~\ref{graph:traj_win}(b), the cost is greater. Increased by almost an order of magnitude compared to the execution times in Figure~\ref{graph:traj_win}(a), the cost still remains linear with the size of wider sliding steps. Again, this is due to the larger amount of accumulated raw locations per batch (depicted with the bar plots). For the larger window tested ($\omega=24$ hours and $\beta=4$ hours), critical points can be reported in less than 2 seconds per batch, even though the mobility tracker has to validate almost 30,000 fresh raw positions each time. This clearly testifies the robustness and timeliness of the online tracking process.

\begin{figure*}[t]
\centering
\begin{minipage}[h]{0.48\linewidth}
{\includegraphics[width=1\textwidth]{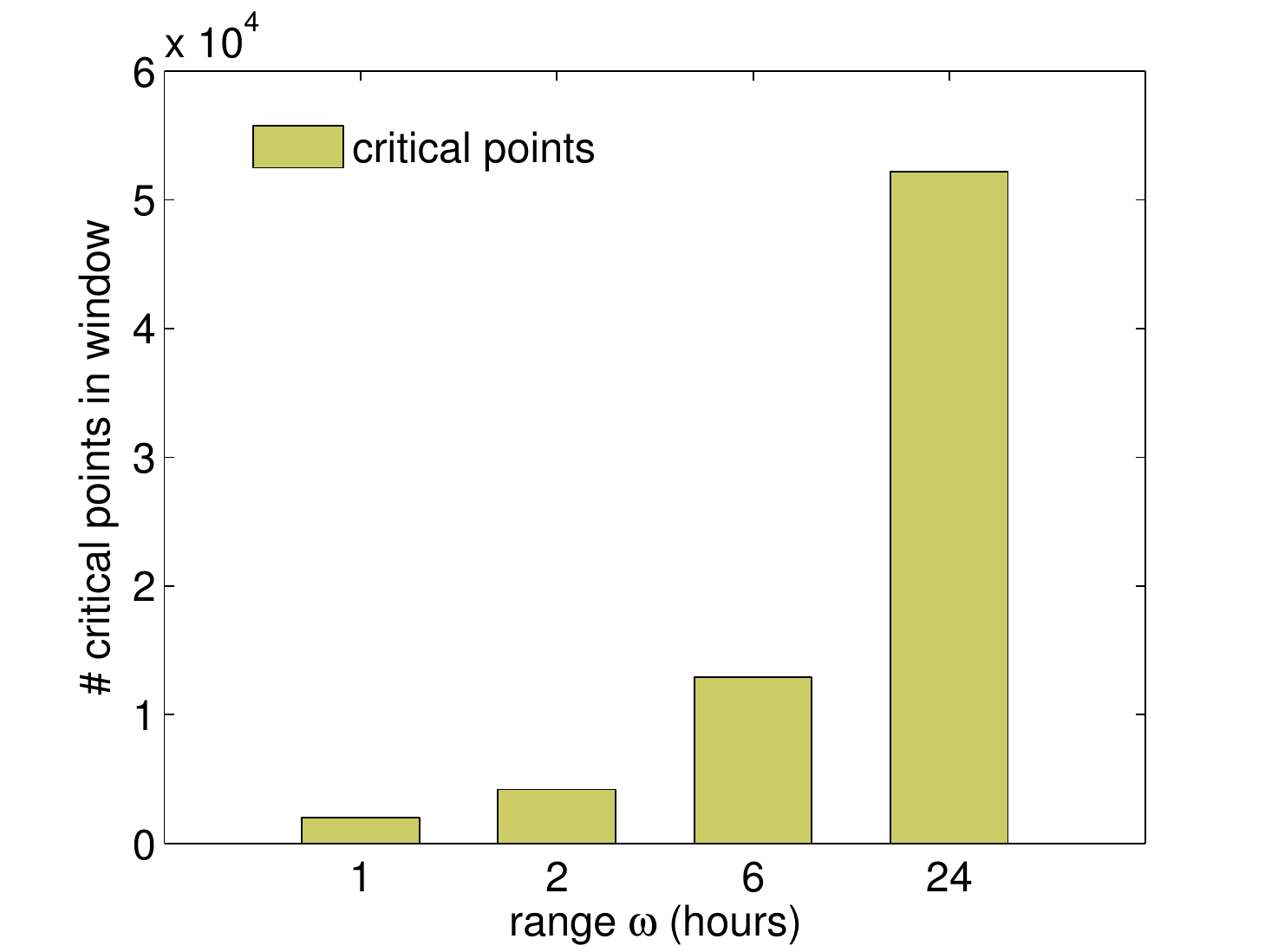}}
\caption{Window states for varying ranges.}\label{graph:win_state}
\end{minipage}
\begin{minipage}[h]{0.48\linewidth}
{\includegraphics[width=1\textwidth]{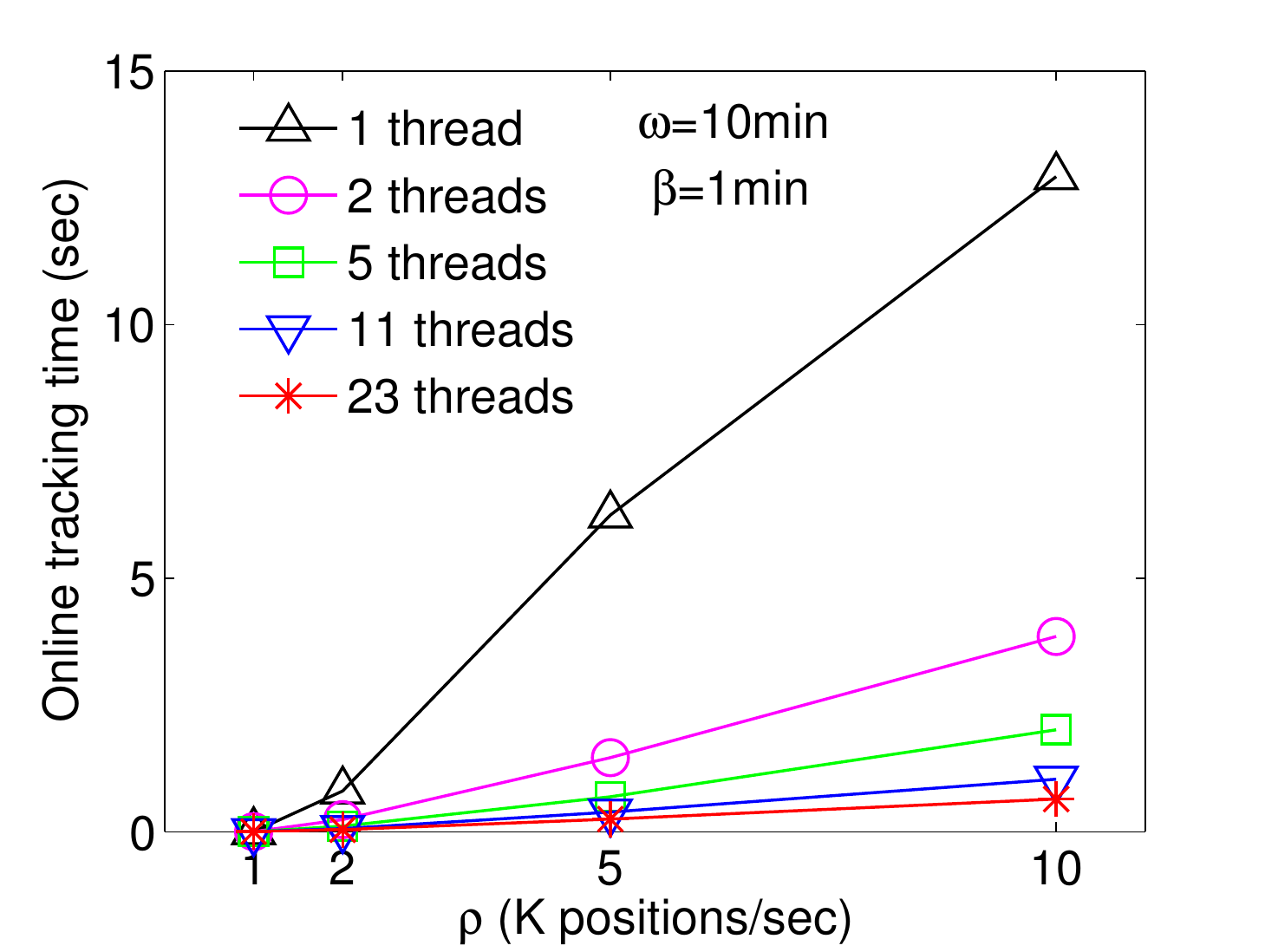}}
\caption{Parallelized online detection.}\label{graph:threads}
\end{minipage}
\end{figure*}

We should also stress that these execution costs are drastically reduced compared to our previous performance study in \cite{[PAK+15]}. Apart from better memory management in the implementation of the method, this improvement should be also attributed to the extra module for noise reduction. Since each position qualified as noise is filtered out without further processing, this incurs no more checks against the rest of the positions reported per vessel. Moreover, it reduces the size of the trajectory synopsis (i.e., critical points retained out of the raw positions per vessel), since such erroneous deviations from the known course are suppressed. Figure~\ref{graph:win_state} plots the average amount of critical points retained per window state for several window ranges $\omega$. It is no wonder that the number of critical points in window are proportional to its range, as this is actually the memory footprint of the maintained trajectory synopses. Space consumption is provably lightweight, since the trajectories of all vessels within the latest $\omega=24$ hours can be approximately reconstructed from the almost 52,000 critical point locations maintained in the respective window state.

\subsubsection{Performance under varying arrival rates} 

Admittedly, such swift processing of raw positions is largely due to the low arrival rate of the original AIS stream (on average $\rho\approx$50 positions/sec). Hence, for a more stringent assessment of the online mobility tracking module, we performed some extra simulations, 
by admitting bigger chunks of data for processing at considerably increased arrival rates up to $\rho=$10,000 positions/sec. 
Then, given the fleet size $N$, every ship appears as reporting almost twice per second; although quite improbable in practice, this makes sense as a stress test. 

As our objective is timeliness, the window for the simulations in Figure~\ref{graph:threads} was set with range $\omega=10$ minutes and slide $\beta=1$ minute. We first discuss performance when employing a {\em single processor} to tackle the entire trajectory detection process. 
Observe that critical points are still issued promptly for $\rho=1,000$ positions/sec, but the latency grows with increasing rates. 
Note that this cost includes reporting time for the resulting critical points (i.e., after detection), and this adds a significant overhead at higher arrival rates, as greater chunks of AIS updates inevitably generate more critical points. In the worst case tested with $\rho=10,000$ positions/sec, 
the online mobility tracker accepts 600,000 fresh raw positions every minute; yet, it can output results in less than 13 seconds, well before the next window slide. 

The fact that  the mobility tracker updates and maintains each trajectory in isolation from the rest, offers great opportunities for more advanced scalability. Indeed, the trajectory detection process can be parallelized by using {\em multiple processors}: each one is responsible to monitor a distinct subset of vessels. In future work, we plan to study advanced parallelization schemes for trajectory detection. However, as a proof of concept, we implemented a simple scheme with a varying number of concurrent threads for the mobility tracker. Each thread consumes a substream of the incoming raw positions that correspond to the particular vessels it has been assigned to monitor. For simplicity, this subdivision is based on simple hashing over the $MMSI$ identifier of the vessel, such that its positions are always propagated to the same thread to establish consistency in trajectory maintenance. The system load may not be evenly balanced among the threads and cannot account to fluctuations in the arrival rate, but still the burden of processing can be shared and thus boost performance. In this case, the overall tracking cost per window is considered the maximum of costs incurred by each of the concurrent threads in order to emit results. Obviously, this cost differs depending on the size of the incoming batch consumed by a thread in each window instantiation. Figure~\ref{graph:threads} plots the average tracking cost per window slide when multiple threads are used (due to hashing, a prime number of threads was specified). There are significant savings even when employing two threads only; the original stream is halved into two substreams, but the cost drops by almost two thirds as each thread exploits better the available system resources. With more threads the cost still drops, although at a lower pace due to the overhead from context switching and contention for system resources. Overall, even this simplified approach confirms that the trajectory detection process is capable of handling scalable volumes of streaming vessel positions and has great potential for parallelization and advanced load balancing.

\subsubsection{Approximation error}

\begin{figure*}[t]
\centering
\begin{minipage}[h]{0.48\linewidth}
{\includegraphics[width=1\textwidth]{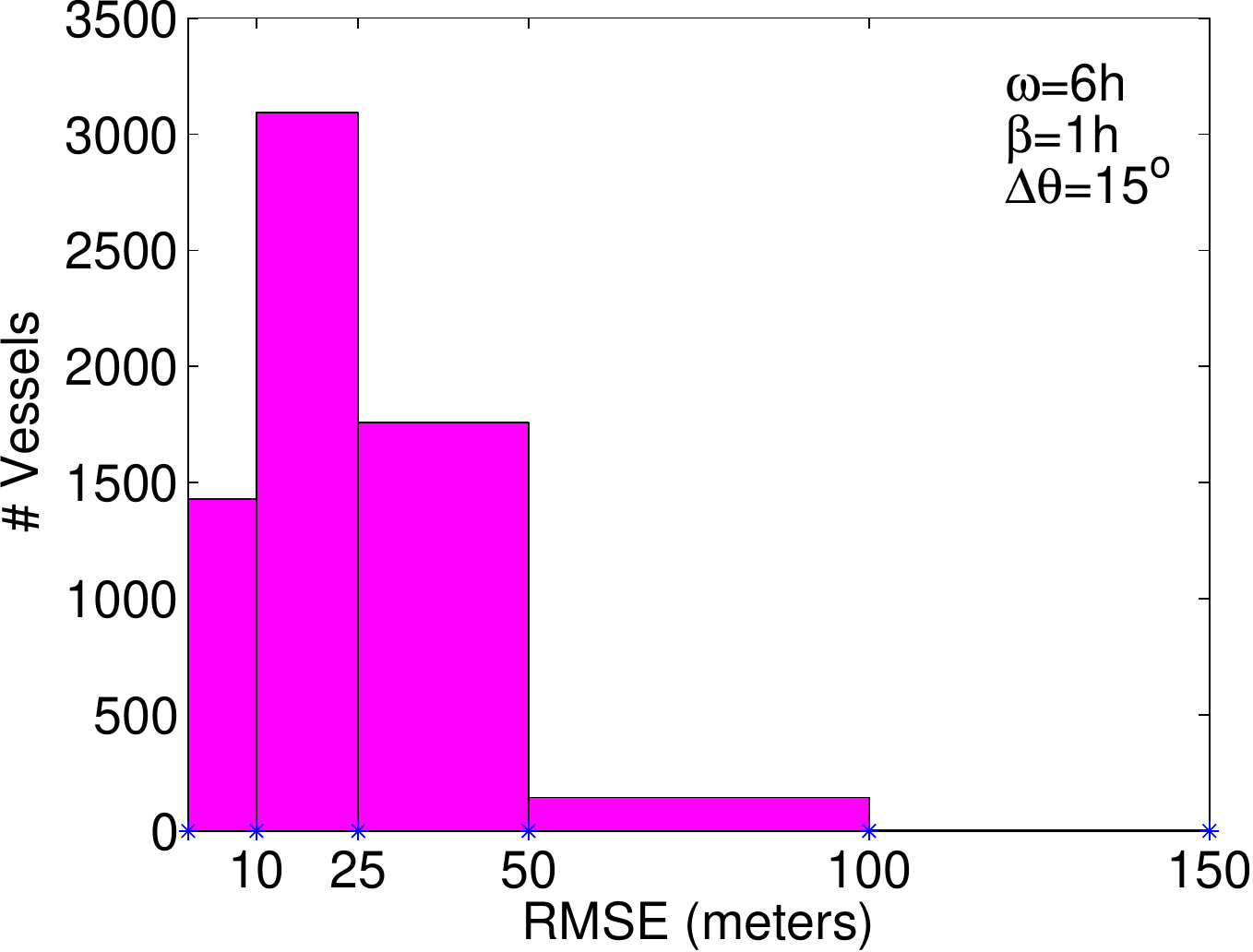}}
\caption{Breakdown of RMSE for $\Delta\theta=15^o$.}\label{graph:RMSE_breakdown}
\end{minipage}
\begin{minipage}[h]{0.48\linewidth}
{\includegraphics[width=1\textwidth]{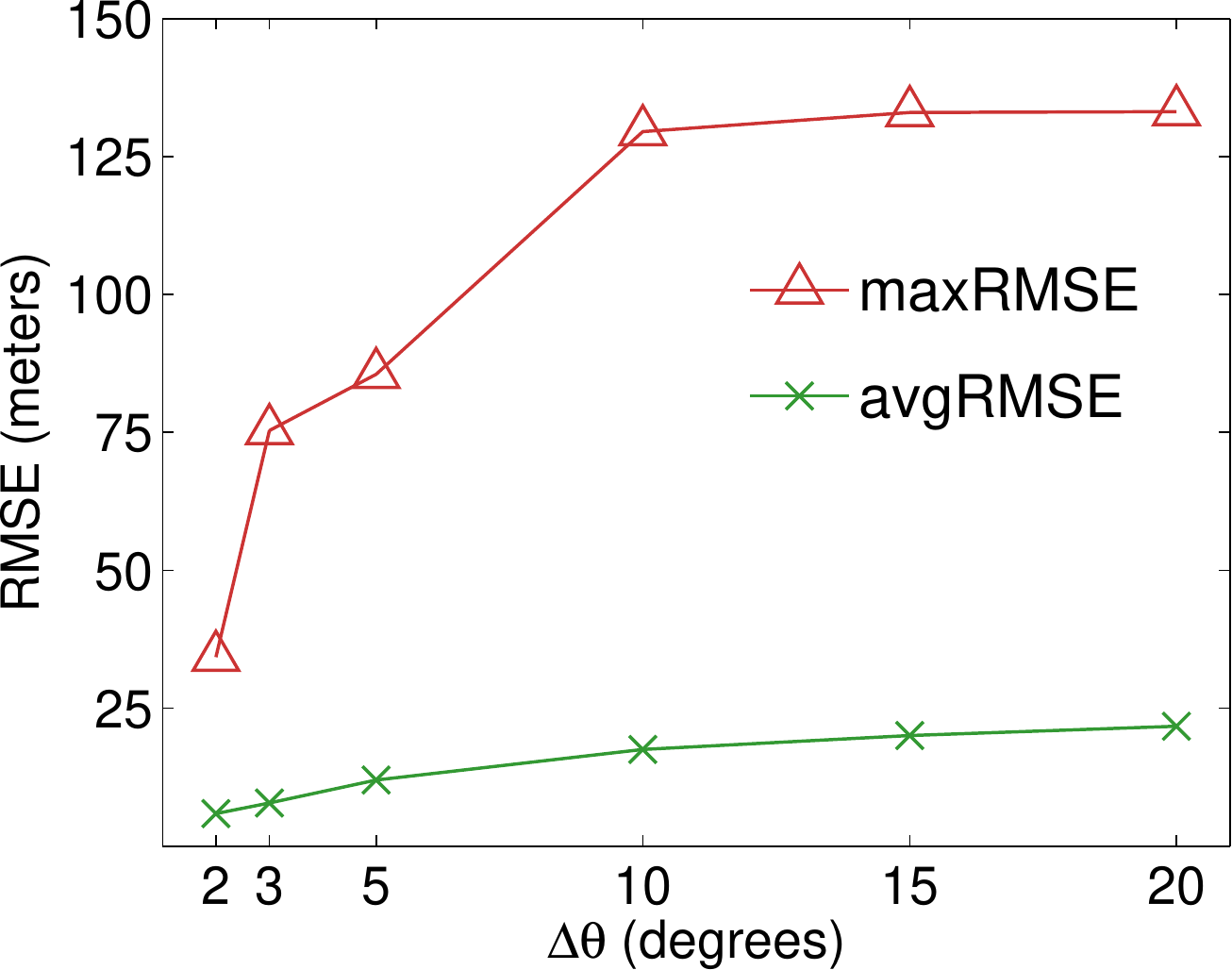}}
\caption{Trajectory approximation quality.}\label{graph:RMSE}
\end{minipage}
\end{figure*}

Preserving only critical points incurs a lossy approximation in trajectory representations. To assess the quality of those compressed trajectories, we estimated their deviation from the original ones (i.e., without discarding any raw positions except for those qualified as noise).
Deviation can be computed from the pairwise distance between {\em synchronized} locations from the original and the compressed trajectory. If an original AIS point $p_i$ at time $t_i$ has been evicted as non-critical, then its corresponding time-aligned $p'_i$  in the compressed trace can be estimated using linear interpolation along the path that connects the two critical points before and after $t_i$. For each vessel that has reported $M$ raw positions, 
we estimated the root mean square error ($\mathit{RMSE}$) between its original and synchronized sequences of its locations as:
\[ \mathit{RMSE} = \sqrt{\frac{1}{M} \cdot \displaystyle\sum_{i=1}^{M} (H(p_i, p'_i))^{2} } \]
\noindent which returns one RMSE estimate (in meters) per vessel trajectory and employs Haversine distance $H$ between geographic coordinates.
Figure~\ref{graph:RMSE_breakdown} illustrates the number of vessel trajectories for certain intervals of these RMSE estimates. For example, RMSE is between 10 and 25 meters for trajectories of 3093 vessels (almost half of the fleet), whereas RMSE less than 10 meters occurs for another 1428 vessels. In contrast, only 3 vessels were found with RMSE above 100 meters. Although parametrization of the mobility tracker is common for all vessels, this result proves that it can offer a correct (or at least fairly trustful) approximation in almost all cases.

Figure~\ref{graph:RMSE} plots the {\em average} and {\em maximum} $\mathit{RMSE}$ over the entire fleet for several values of turn threshold $\Delta\theta$, which is used to recognize significant changes in heading. As discussed in Section~\ref{sec:tracking}, the degree of trajectory approximation is mostly sensitive to parameter $\Delta\theta$ compared to the rest in Table~\ref{tbl:calibration} and this is reflected on the plot. Both error estimates escalate as this angle tolerance gets more relaxed. In the worst case for $\Delta\theta=20^o$, average RMSE is only 22 meters and the maximum $\mathit{RMSE}$ ever observed is 133 meters, which are negligible compared to the much larger size of open-sea vessels, and also considering the discrepancies inherent in GPS positioning and AIS transmissions. In practice, a moderate threshold of $10^o$ or $15^o$ may be adequate for balancing compression efficiency without losing important details in vessel mobility. Therefore, the suggested method can provide quite acceptable accuracy and can capture most, if not all, critical changes along each vessel's course.

\subsubsection{Compression efficiency} 

\begin{figure*}[t]
\centering
\begin{minipage}[h]{0.48\linewidth}
{\includegraphics[width=1\textwidth]{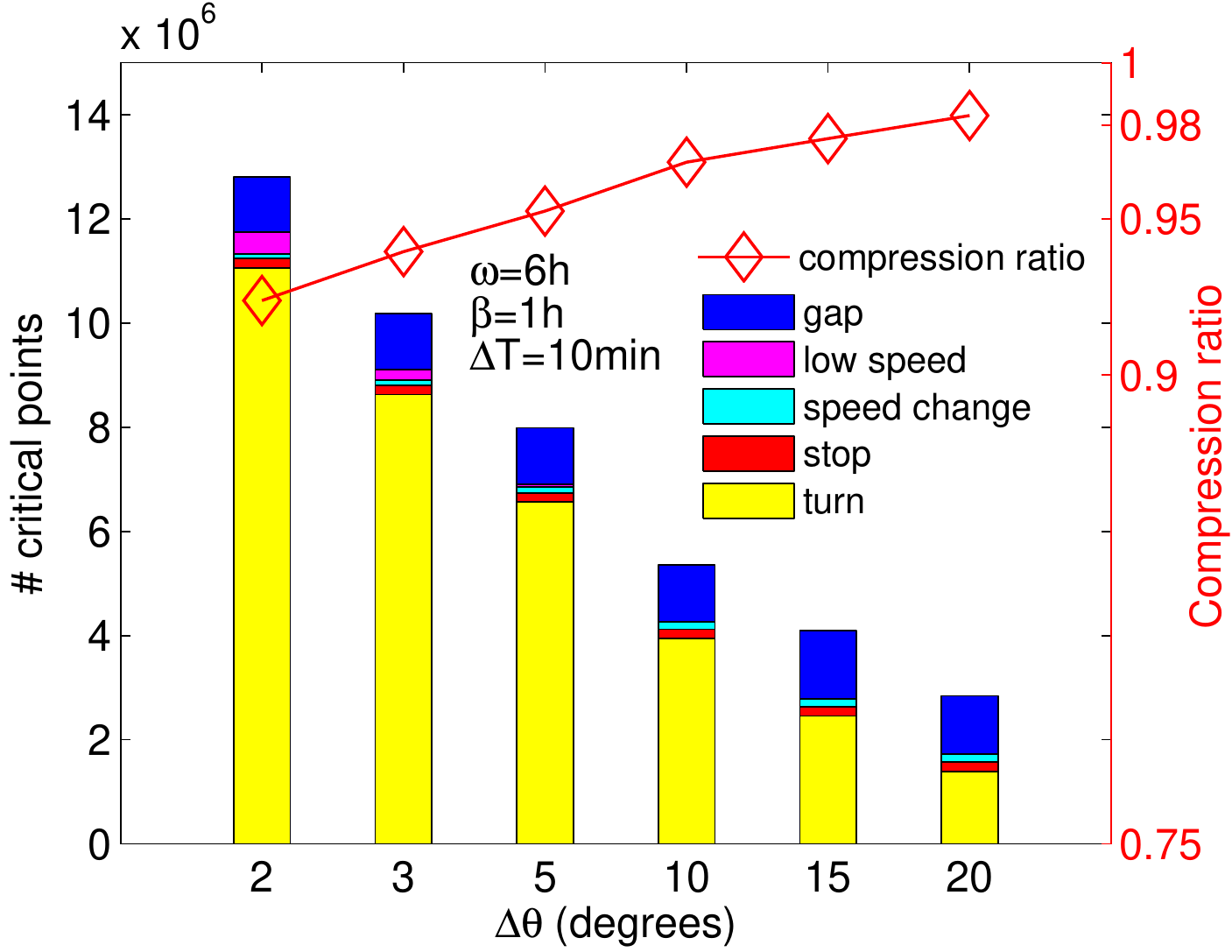}}
\caption{Compression for varying $\Delta\theta$.}\label{graph:reduction_angle}
\end{minipage}
\begin{minipage}[h]{0.48\linewidth}
{\includegraphics[width=1\textwidth]{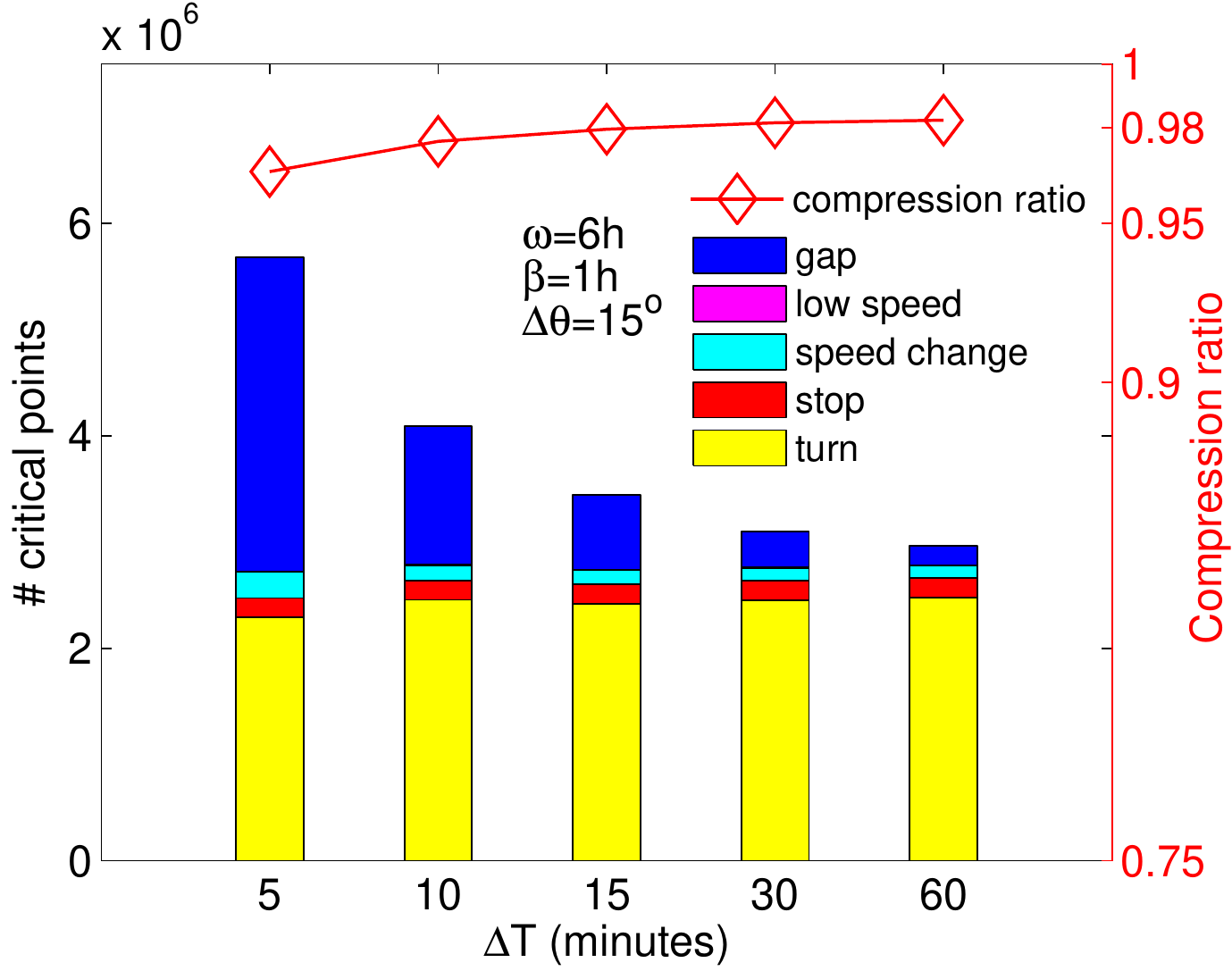}}
\caption{Compression for varying period $\Delta T$.}\label{graph:reduction_gap}
\end{minipage}
\end{figure*}

In this experiment, we examine the efficiency of our prototype in keeping only major trajectory characteristics as critical points and discard the rest. 
In order to measure the {\em compression ratio} accomplished by online trajectory tracking, we compared the amount of discarded points against the originally relayed locations per vessel. A compression ratio close to 1 signifies stronger data reduction, as the vast majority of original locations are dropped. 
The red line plot in Figure \ref{graph:reduction_angle} depicts measurements of this ratio with varying tolerance angles for detecting changes in heading. 
With a lower $\Delta\theta$, even slight deviations in vessel direction can be spotted, and thus extra critical points get reported. Bar charts  in Figure~\ref{graph:reduction_angle} illustrate the amount of critical points in each class (gap, low speed, speed change, stop, turn) retained from the entire dataset. Clearly, every further increase in threshold $\Delta\theta$ suppresses more and more turning points and only marginally affects the share of other classes, incurring extra reduction in the total amount of emitted critical points. Hence, relaxing this parameter value leads to a more intense compression. 
Most importantly, compression ratio always remains above 92\%, and with a more relaxed $\Delta\theta$ it reaches as much as 98\%. In this latter case, only 2\% of the original locations survive as critical, mostly by eliminating local manoeuvres of little impact on vessel's course. Eliminating noise also plays an important role in data reduction, as erroneous deviations are dropped and no points need be retained.

A similar pattern regarding reduction efficiency can be observed in Figure~\ref{graph:reduction_gap} with respect to varying periods $\Delta T$ for detecting gaps in communication. Not surprisingly, it is the amount of critical points marking those gap periods that gets reduced with increasing thresholds $\Delta T$. This time, reduction ratio is never below 96\%, even though many more points are required to keep track that contact was lost even for 5 minutes. In a streaming context, such high compression ratios may lead to reduced system load in subsequent stages of the analysis, without sacrificing quality, as discussed earlier.


\subsubsection{Quality of synopses} 

\begin{figure*}[t]
\centering
\begin{minipage}[h]{0.48\linewidth}
{\includegraphics[trim=1cm 1.5cm 1cm 0, width=1\textwidth]{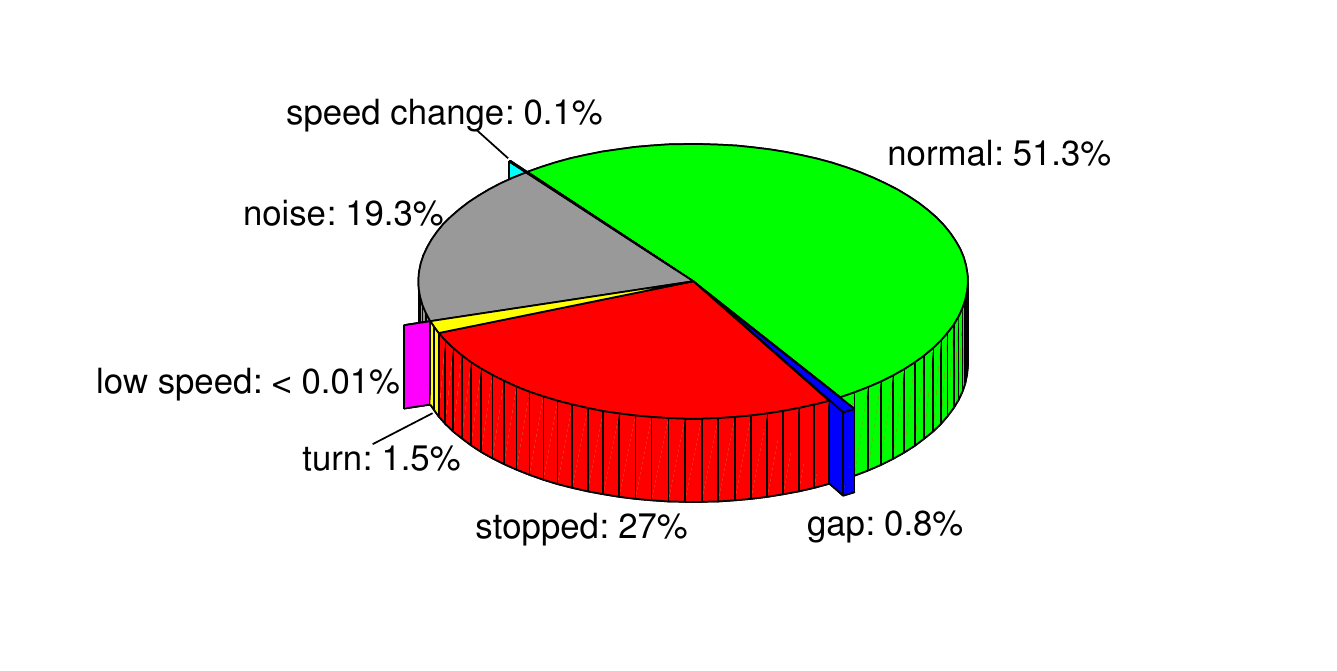}}
\caption{Classified raw locations of all vessels.}\label{graph:classified_points}
\end{minipage}
\begin{minipage}[h]{0.48\linewidth}
{\includegraphics[trim=0 1.15cm 0 0, width=1\textwidth]{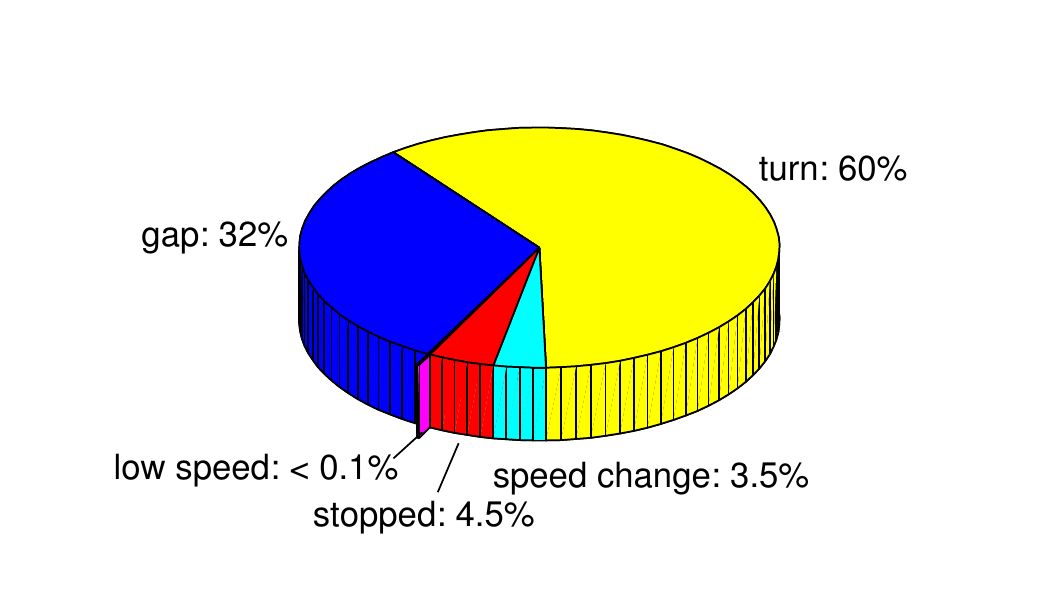}}
\caption{Characterization of critical points.}\label{graph:classified_critical_points}
\end{minipage}
\end{figure*}

As raw AIS locations pass through the trajectory detection module in successive window instantiations, they get characterized according to their significance on vessel mobility. Figure~\ref{graph:classified_points} illustrates a breakdown of the resulting classifications after the input stream was exhausted and all critical points were detected for the entire 3-month period. More than half of the relayed raw positions indicate a ``normal'' course, i.e., a vessel moves according to its known velocity vector with a steady speed and heading. Thus, apart from notifying on the current position of each vessel, such points practically do not alter its trajectory and can be safely discarded without any further consideration. In addition, almost one out of five original positions is classified as noise for reasons explained in Section~\ref{sec:noise}. It must be stressed that the vast majority of such points are not really ``off-course'' positions from the reporting vessel, but they actually fall along its route. However, due to their delayed arrival, these locations falsely indicate the vessel as moving back and forth in an agitating manner with no obvious reason, hence they should be purged altogether.

But it is the remaining vessel locations, after eliminating redundancy and noise, which get actually classified as critical points and can be used in trajectory summarization. As Figure~\ref{graph:classified_points} testifies, there are relatively very few cases (about 0.1\% of the raw data) in which vessels either move at low speed or they change their speed considerably. This is to be expected, since vessels usually follow their planned course and they typically manoeuvre when arriving to or departing from ports. Gaps in communication account for about 0.8\% of the raw data. Although we observed that this phenomenon is rather frequent in practice, we note that at most two points are used to delimit this period and thus indicate the loss of contact with the respective vessel. Locations indicating turns are very important and must be certainly considered in trajectory representation. In fact, these points are roughly 1.5\% of the total, since only significant deviations (over 15$^o$) from the known course qualify for a turn, even though possibly emitting a series of such critical points in case of smooth turns as depicted in Figure~\ref{longterm}(b). Finally, about 27$\%$ of the raw positions are emitted when vessels are idle, most usually when anchored in a port. As discussed in Section~\ref{sec:trajectory}, instead of keeping all these points, we collect successive ``pause'' events occurring within a small distance and merge them into a collective ``stop'' event located at their centroid. This incurs huge savings in the resulting synopses, leading into much more concise trajectory representations that can be highly usable in subsequent query processing and offline analytics.

This dramatic effect on summarization is much more evident in Figure~\ref{graph:classified_critical_points}, which plots a breakdown of the accumulated critical points after processing the entire dataset. The resulting trajectories mainly consist of turning points between stops with some occasional changes in speed, but rather frequent communication gaps. Indeed, 60\% of critical points are turning points, which are only 1.5\% of the original AIS locations (Fig.~\ref{graph:classified_points}). Points indicating gap periods are almost 32\% of the critical points, testifying the frequent loss of contact with vessels on the move. In contrast, long-term stops comprise a meagre 4.5\% in the resulting trajectory synopses, since locations that belong to the same stop event are compressed into a single centroid that suffices to designate that the vessel is idle during this period. Apart from redundant ``normal'' points and eliminating the inherent noise, condensing these stops really adds much to the reduction efficacy of the trajectory detection module, offering a valuable semantic interpretation of the motion features.

\begin{table}[t]
\centering \caption{Statistics from post-processing of compressed trajectories.} \label{tbl:reconstructed}
\begin{scriptsize}
\begin{tabular}{|l|c|}
\hline Critical points in reconstructed trips & 3,895,112 \\
\hline Critical points in open-ended trips & 196,131 \\
\hline Average trips per vessel & 20 \\
\hline Average number of critical points per trip &  48 \\
\hline Average travel time per trip & 07:24:48 \\
\hline Average traveled distance per trip & 131.513km \\
\hline
\end{tabular}
\end{scriptsize}
\end{table}

As an offline, post-processing step, we have reconstructed trajectories from the entire sequence of critical points accumulated per vessel. In effect, the long motion history of a ship can be broken up into shorter ``trips'' between identified stop points, which usually indicate anchorage at ports. Table~\ref{tbl:reconstructed} lists representative statistics from these approximate trajectories, and offers insight on a possible offline usage of the results from trajectory summarization. It turns out that a typical trip spanning several hours over a long distance (more than 131km) can be approximated with 48 points only; once more, this confirms the strong reduction effect of the method. Note that almost 5\% of the detected critical points belong to ``open-ended'' trips, as certain vessels were only spotted while traversing the Aegean without anchoring there.

\subsection{Assessment of Complex Event Recognition}
\label{sec:empirical-cer}

The trajectory detection module compresses a vessel position stream to a stream of critical movement events (ME)s. Each such event is represented by predicates expressing the activity of the vessel, its coordinates and its velocity (see Section \ref{sec:cer}). This way, the ME stream given to RTEC includes 15,884,253 predicates.  In addition to this stream, RTEC makes use of real data consisting of protected areas, such as NATURA areas, represented as polygons, and ports, represented as points, across the Greek seas. The dataset has 3,966 protected areas with a total of 78,418 edges, and 64 ports. The size of the grid is 720$\times$900 km$^2$. Given this combination of event stream and static geographical information, RTEC recognizes the following CEs: illegal shipping, suspicious vessel delay, vessel rendezvous, suspicious areas, vessel pursuit, and package picking. 

\subsubsection{Grid partitioning}

\begin{figure*}[t]
\centering
\subfigure[Average CE recognition times.]
{\includegraphics[width=0.4\textwidth]{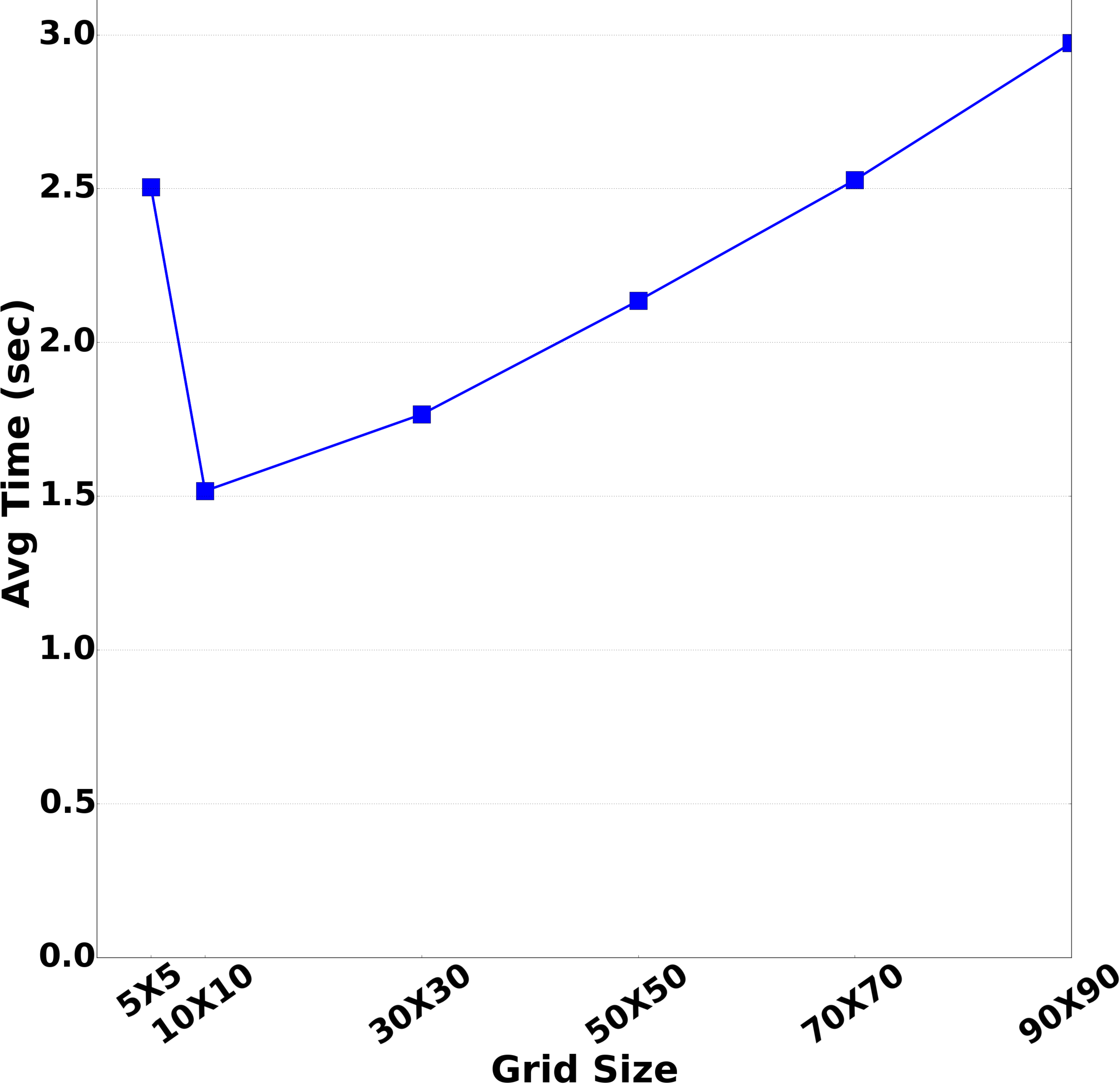}}$\qquad$
\subfigure[Average number of CEs.]
{\includegraphics[width=0.4\textwidth]{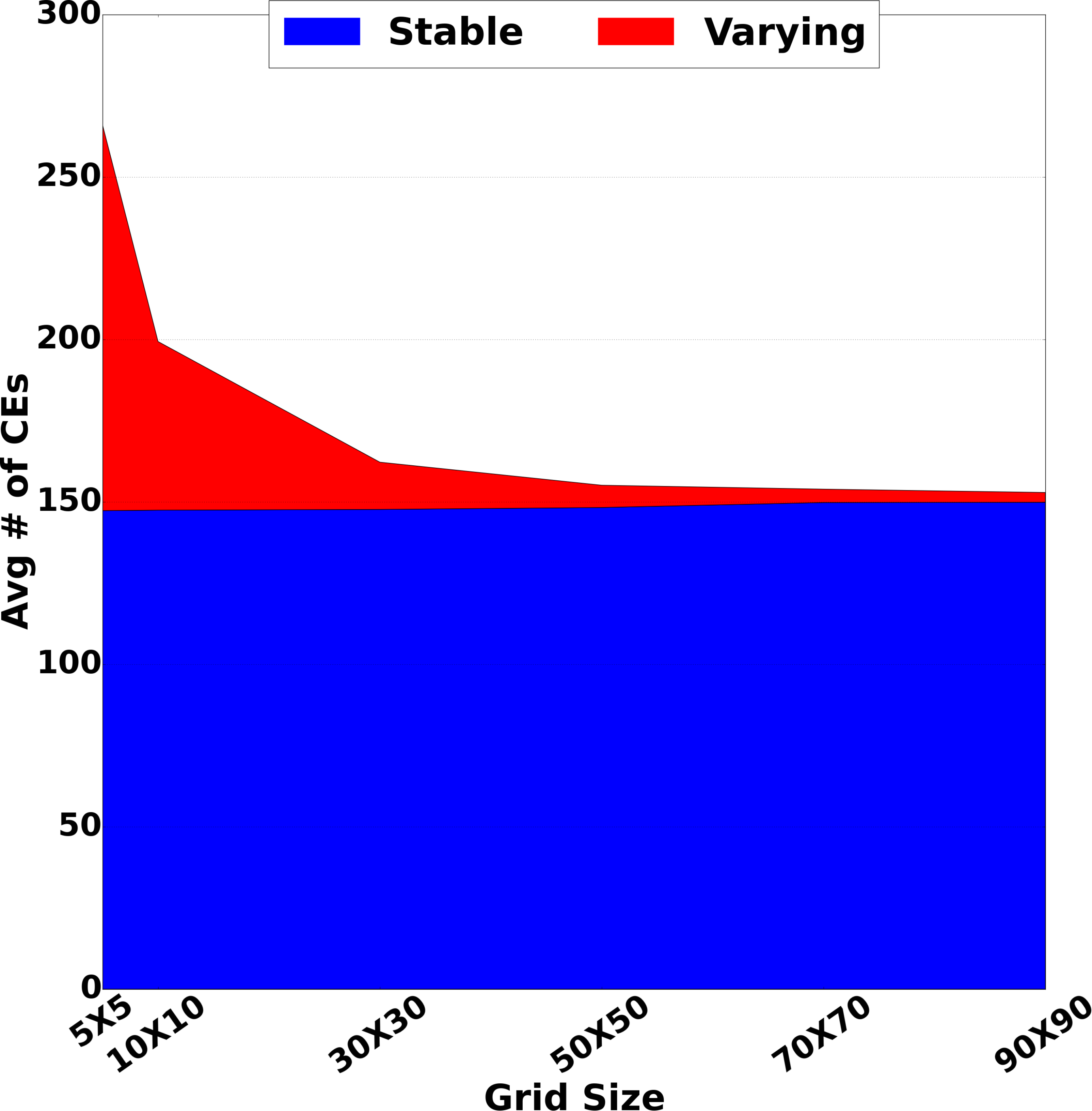}}
\centering
\caption{CE recognition for different grid cell sizes.} 
\label{fig:cer_grids}
\end{figure*}

Figure \ref{fig:cer_grids} shows the results from a first set of experiments in which we attempted to determine the optimal grid granularity/cell size. Starting with a grid having\linebreak $5{\times}5\val 25$ cells (with a cell size of $138{\times}170$ km$^2$), we increased the number of cells along each dimension, up to $90{\times}90\val$ 8,100 cells (each cell being $9{\times}11$ km$^2$ wide). Figure \ref{fig:cer_grids}(a) shows the average CE recognition times in CPU seconds for each different grid. Both the window $\omega$ and the slide $\beta$ are set to 1 hour. 
The worst grid configuration ($90{\times}90$) is almost two times slower than the best ($10{\times}10$), but in all cases the average time is within the same order of magnitude (and less than 3 seconds). Figure \ref{fig:cer_grids}(b) shows the average number of recognized CEs for each different grid as a stack plot. The number of recognized CEs shows a decreasing trend initially, with a tendency to stabilize after grid configuration $30{\times}30$. The reason for the difference in the number of detected CEs lies in the way the respective patterns are defined. Most of the CE patterns are grid-independent. Therefore, the number of recognized CEs remains stable across all grid configurations. On the other hand, $\mathit{possibleRendezvous}$, as defined by rule \eqref{eq:possible-rendezvous}, depends on the size of the grid cells. Recall that the place of vessel rendezvous cannot be determined precisely, since vessels stop transmitting AIS signals during the time of the meeting. Therefore, we can only define $\mathit{possibleRendezvous}$ in terms of the cells in which the vessels in question stopped (respectively resumed) transmitting AIS signals. As a result, when the size of the cells increases, more $\mathit{possibleRendezvous}$ CEs are being recognized.
For this reason, instead of choosing the $10{\times}10$ grid for the remaining experiments,
we opted for the $30{\times}30$ one, which has comparable time performance and a stable number of detected CEs.

\subsubsection{Performance under varying window sizes and distributed configurations}

\begin{figure}[t]
\centering
\subfigure[Average CE recognition times.]
{\includegraphics[width=0.32\textwidth]{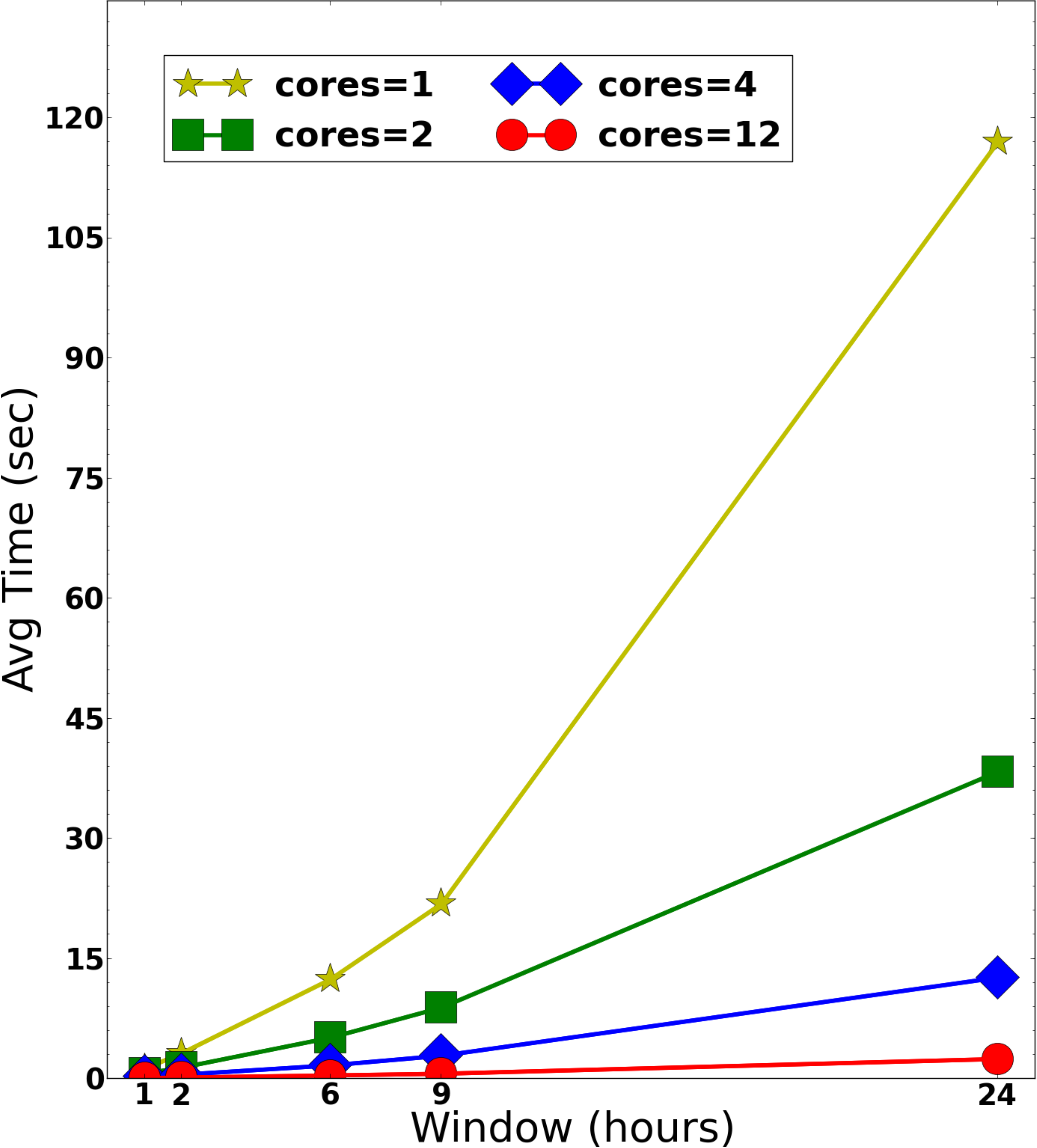}}
\subfigure[Average number of MEs.]
{\includegraphics[width=0.32\textwidth]{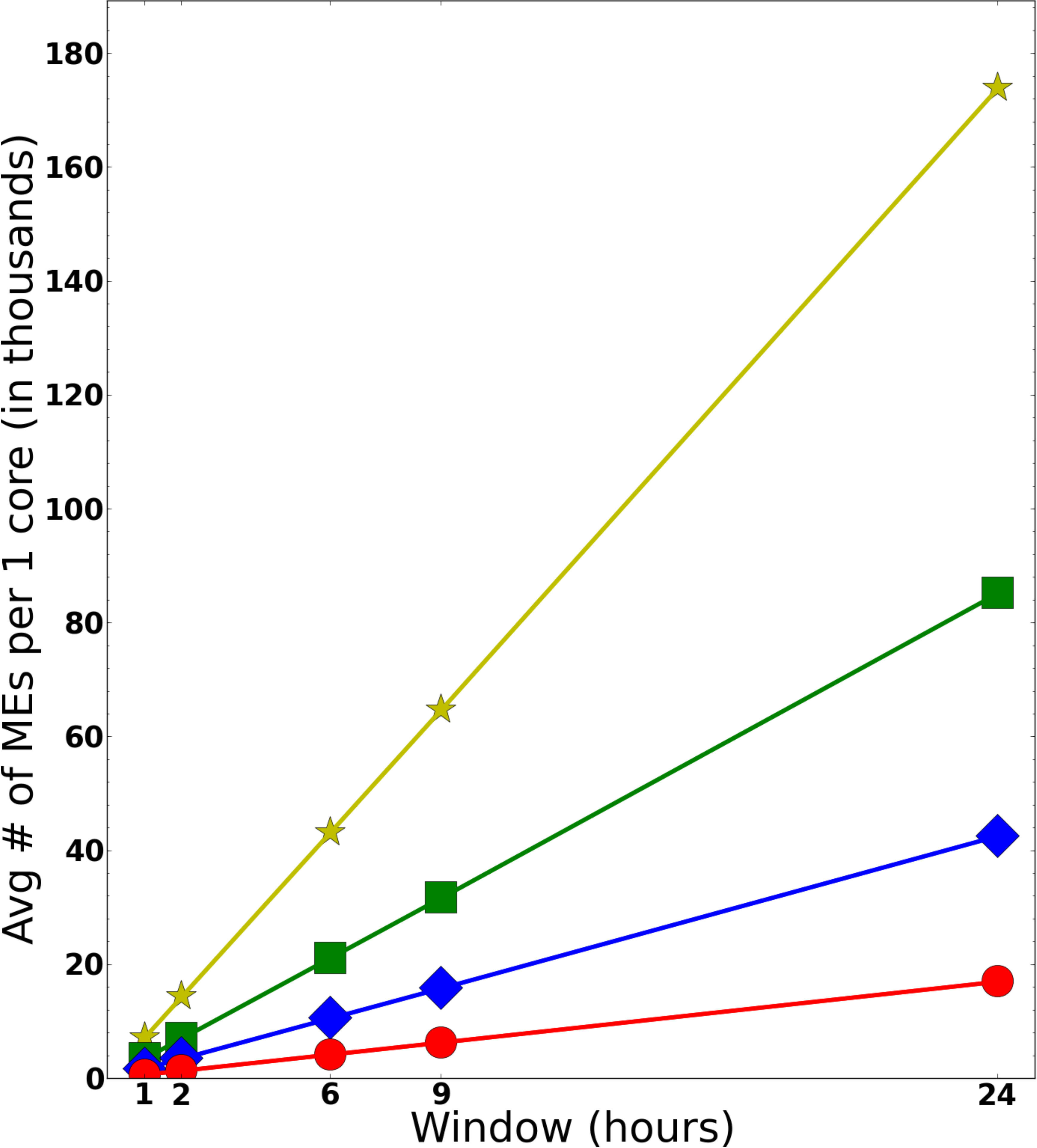}}
\subfigure[Average number of CEs.]
{\includegraphics[width=0.32\textwidth]{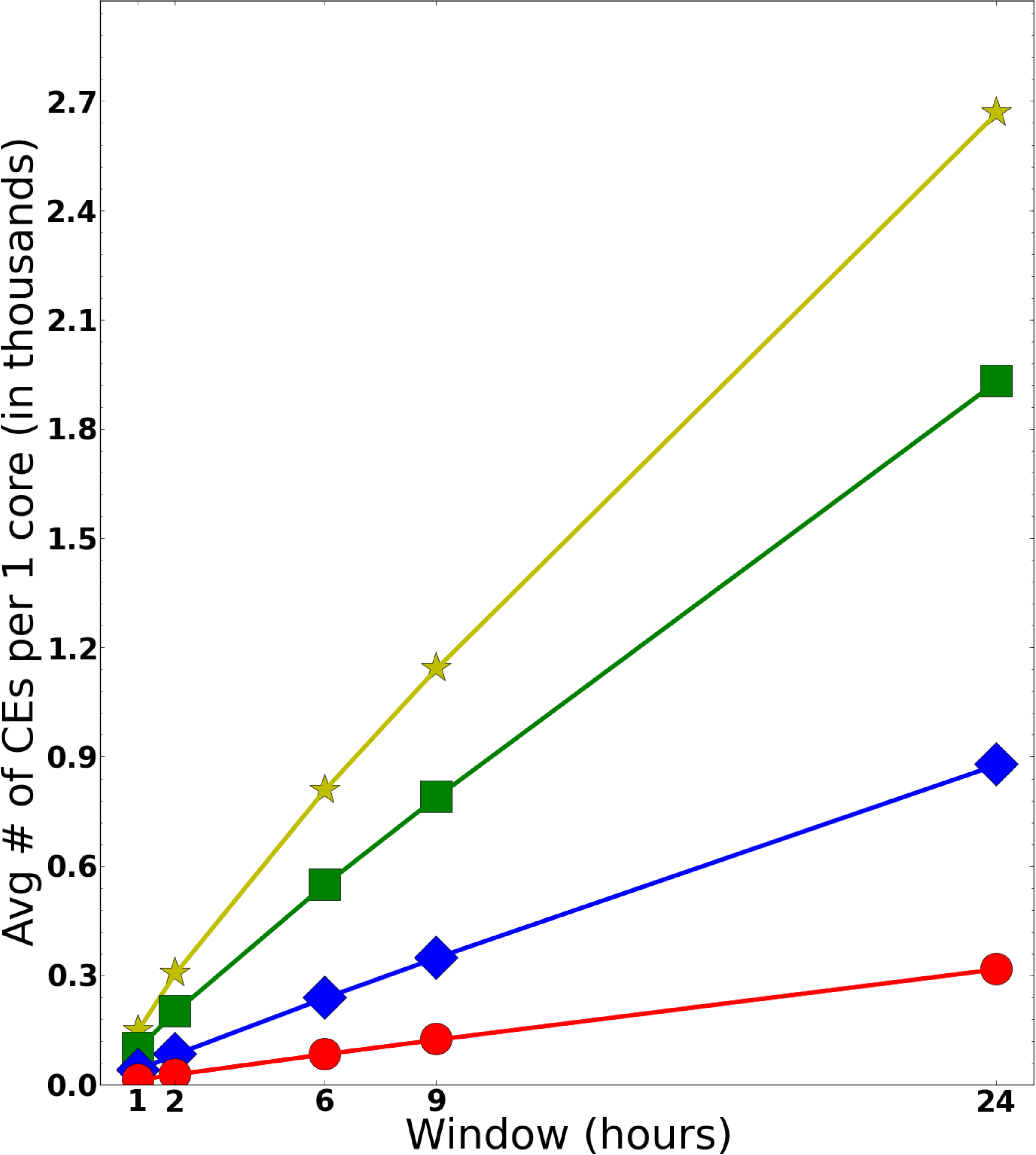}}
\centering
\caption{CE recognition under varying window sizes and distributed configurations.} 
\label{fig:cer}
\end{figure}

Next, we proceed with a more thorough analysis of the performance of RTEC. Figure \ref{fig:cer} shows the results of experiments under various window sizes and distributed configurations. First, we used a single processor to perform CE recognition for all 6,425 vessels, 3,966 areas and 64 ports. We subsequently employed multiple processors on which RTEC operated in parallel, by following a data partitioning scheme. We divided the grid covering the surveillance area into multiple sub-grids (groups of adjacent cells) whose number was equal to that of the processors used in parallel. Each processor was responsible for the areas and ports located in, and the vessels passing through its assigned sub-grid. We used three distributed settings: performing CE recognition on two, four and twelve processors. We made an attempt to evenly distribute the load of MEs among the different processors, by exhaustively searching for the best configuration. The different sub-grids were required to be compact rectangles without dispersed cells. As a result, we did not take into account solutions with sub-grids of arbitrary shapes and the load distribution was thus not the best possible. This is an off-line process that takes place only once.

Figure \ref{fig:cer}(a) shows the average CE recognition times, including the time taken for spatial indexing (see Section \ref{subsec:spatial-pre}). The slide $\beta$ is 1 hour while the window $\omega$ ranges from 1 hour to 24 hours. 
Figure \ref{fig:cer}(b) shows the average number of MEs for each setting. 
In the case of a single processor, the window ranges from $\approx$7,200 MEs (1 hour) to 175,000 MEs (24 hours). In the distributed settings---two, four and twelve processors for CE recognition---
the input MEs are forwarded to the appropriate processor according to vessel location. 
For instance, when twelve processors are used in parallel, each one of them processes $\approx$700 MEs for the 1 hour window, and 17,000 MEs for the 24 hour window. Figure \ref{fig:cer}(c) shows the average number of CEs for each setting. This number also depends on the window size. 
In the case of a single processor, e.g., for 1 hour windows approximately 150 CEs are recognized, 
while for 24 hour windows RTEC recognizes around 2,900 CEs.   
We do not show memory consumption figures because memory usage is negligible and stable.

Figure \ref{fig:cer}(a) shows that we can achieve a significant performance gain by running RTEC in parallel. As the window size increases, the gain becomes more pronounced. Furthermore, Figure \ref{fig:cer}(a) shows that RTEC supports real-time CE recognition. E.g.~for a window of 6 hours, RTEC recognizes all CEs in 14~sec when a single processor is used, and in 0.4~sec when twelve processors are used in parallel. 


\subsubsection{Tolerance to irrelevant events}

\begin{figure*}[t]
\centering
\subfigure[Average CE recognition times and memory usage.]
{\includegraphics[width=0.48\textwidth]{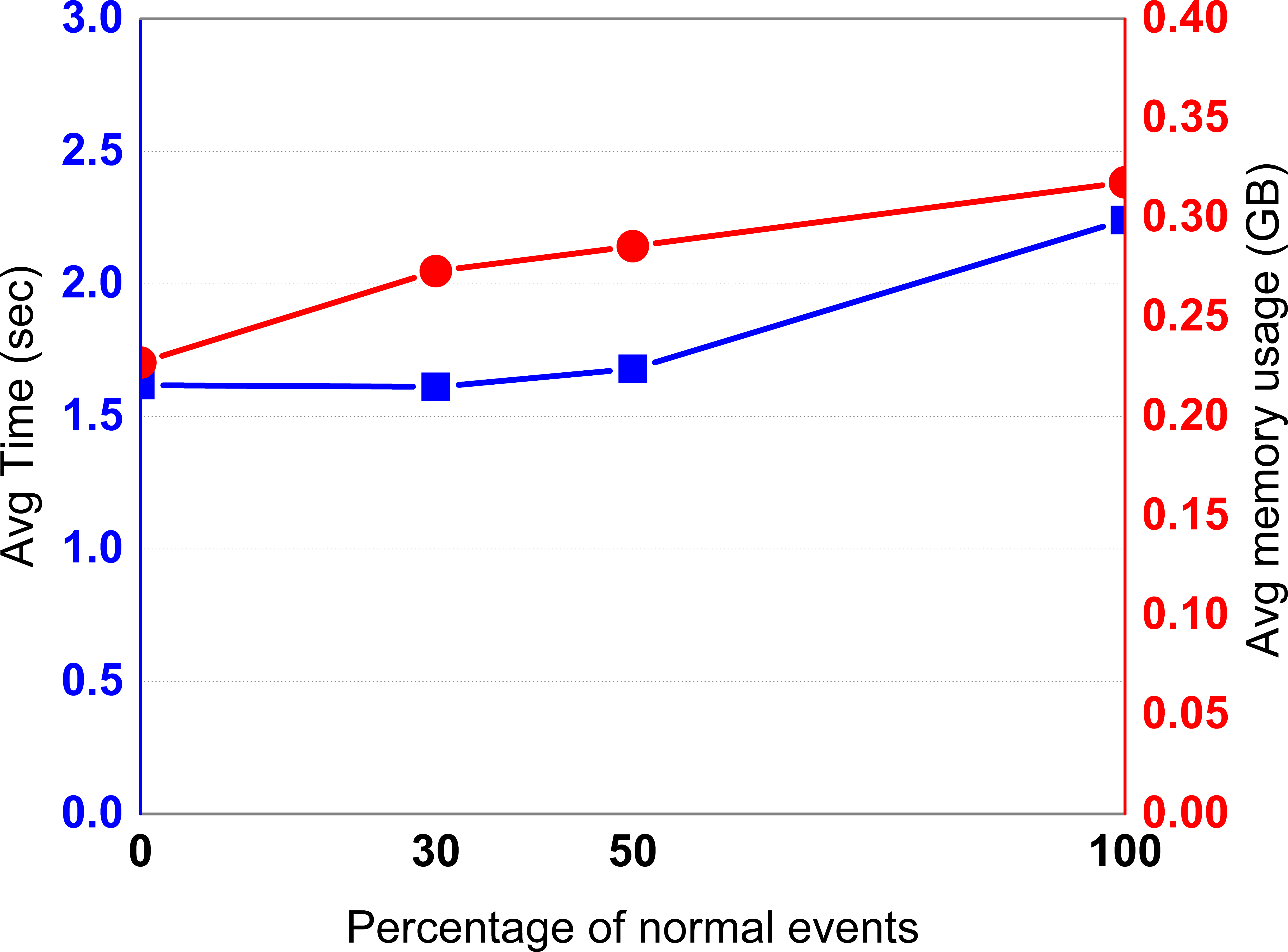}}$\quad$
\subfigure[Average number of input (critical and normal) events and CEs.]
{\includegraphics[width=0.48\textwidth]{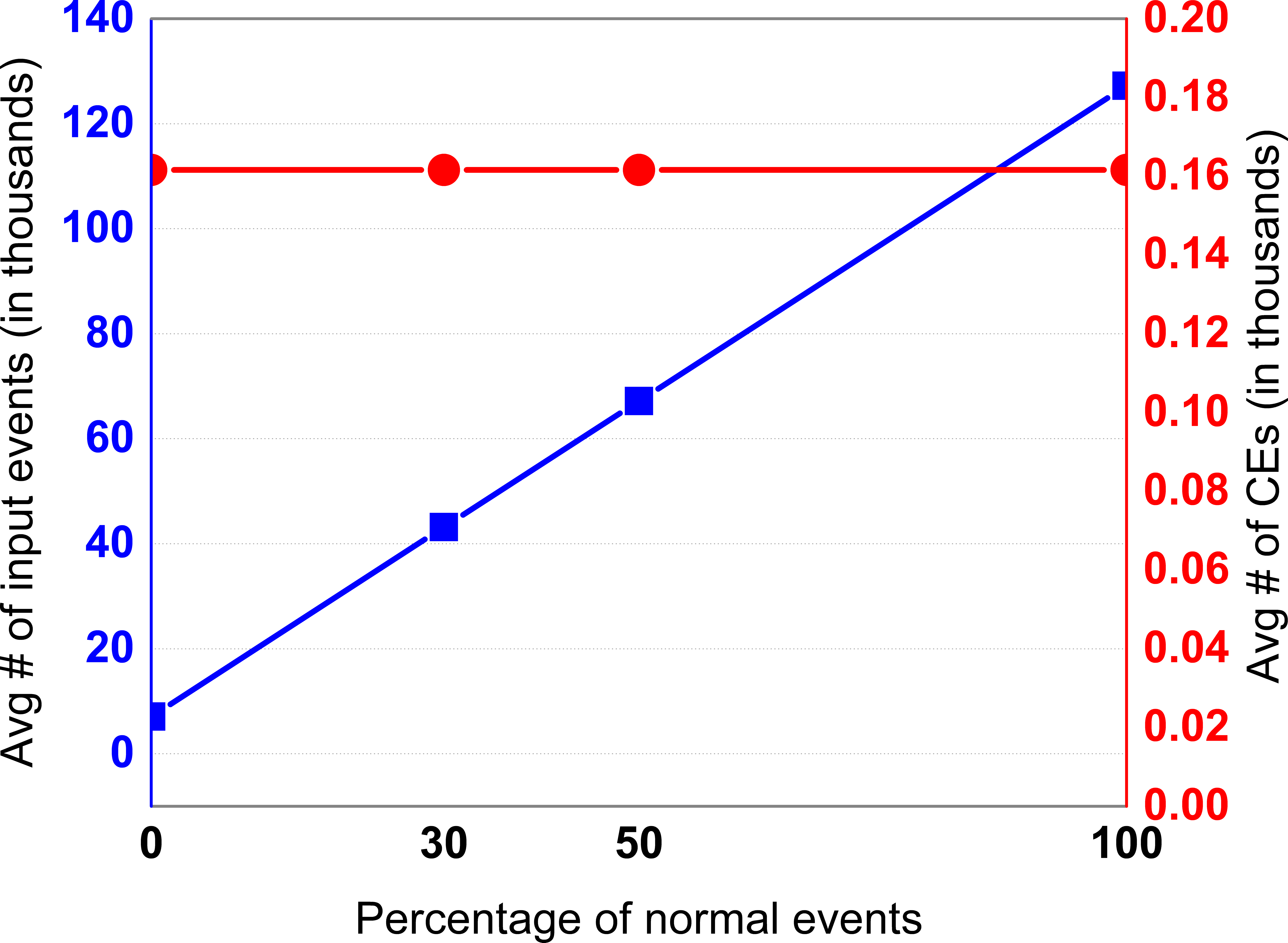}}
\centering
\caption{CE recognition on 16M critical MEs and 79M--262M normal events.} 
\label{fig:cer_nc_same_rules}
\end{figure*}

\begin{figure*}[t]
\centering
\subfigure[Number of input (critical and normal) events at each recognition step.]
{\includegraphics[width=0.7\textwidth]{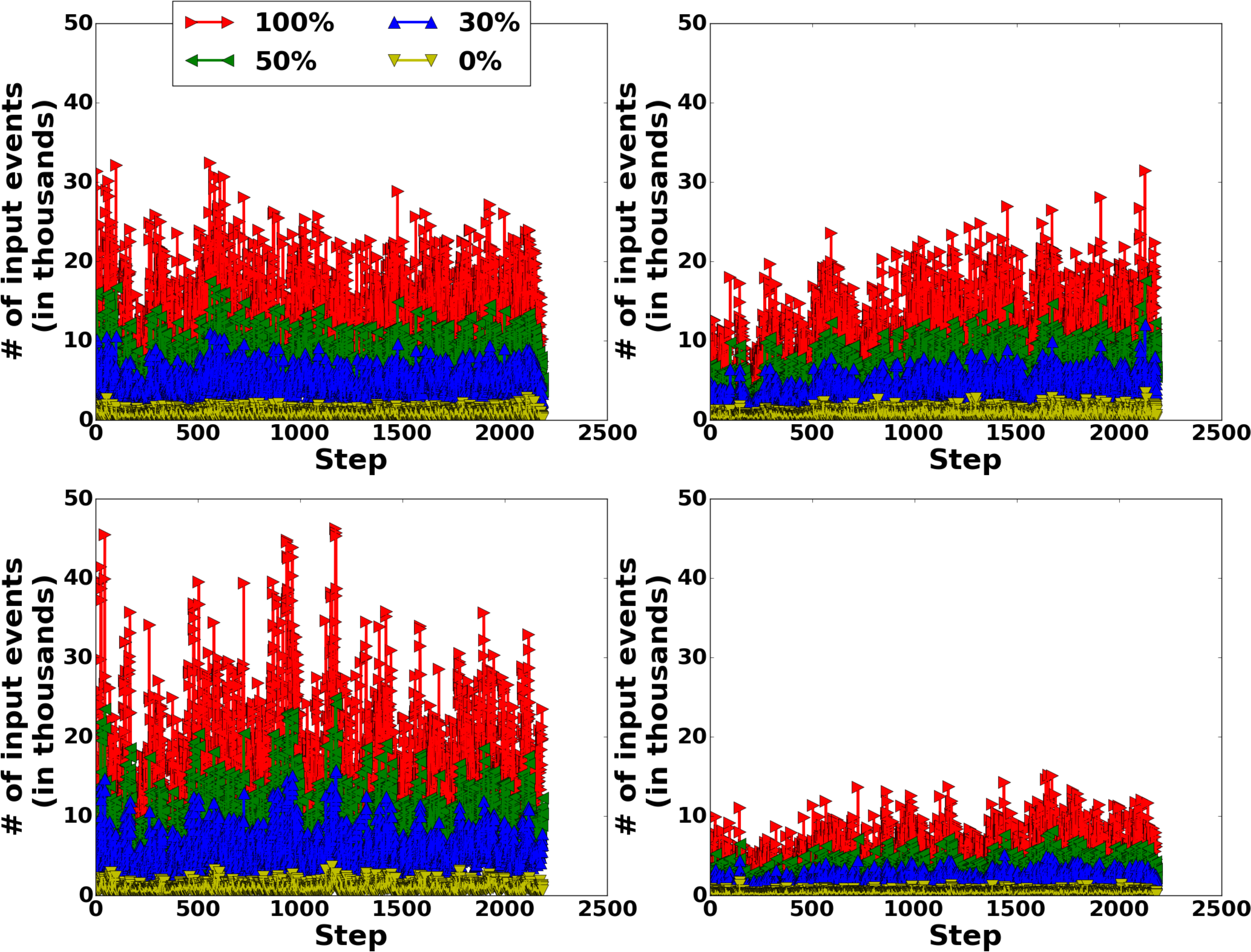}}\\
\subfigure[Number of CEs at each recognition step.]
{\includegraphics[width=0.7\textwidth]{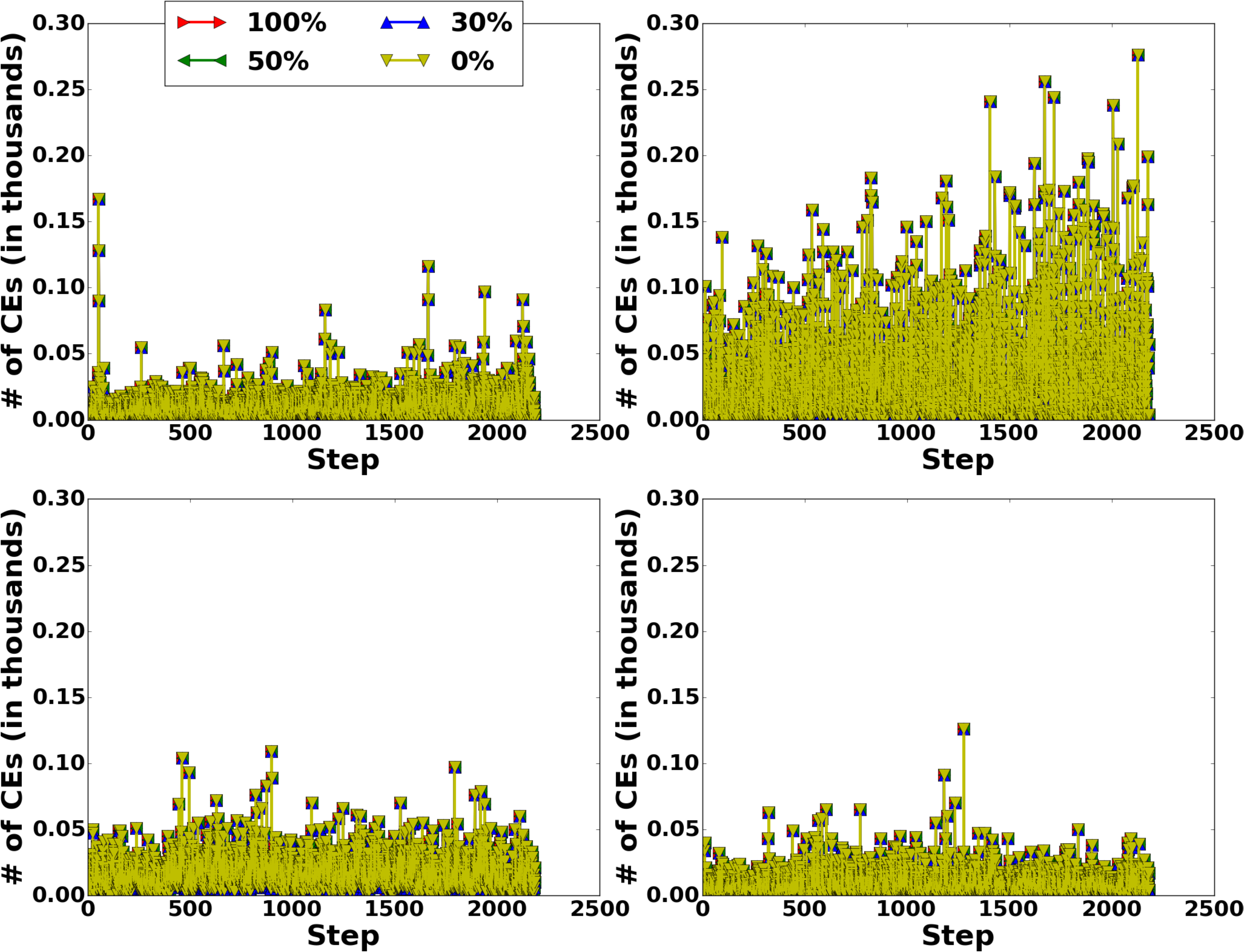}}
\centering
\caption{CE recognition on four out of twelve cores in the presence of irrelevant input.} 
\label{fig:cer_nc_same_rules-actual}
\end{figure*}

In the experiments presented so far, the input stream consisted of {\em critical} MEs, i.e.~all events of the input stream were part of the CE patterns and thus activated the recognition process. 
In several applications, however, the CE recognition module has to deal with input streams including several events that are irrelevant to CE pattern matching. To test the tolerance of RTEC to irrelevant input events, we performed experiments in which, in addition to the critical MEs, we retained a percentage of the normal events, i.e.~the ones not annotated as critical by the trajectory detection module. In these experiments, RTEC has to skip over the normal events in order to find the critical MEs expressing the CE patterns. Figures \ref{fig:cer_nc_same_rules} and \ref{fig:cer_nc_same_rules-actual} shows the results from this set of experiments; the percentage of normal events ranges from $0\%$ (the input stream consists only of the 16M critical MEs) to $100\%$ (the input stream consists of all 262M normal events and the 16M critical MEs). Both the window $\omega$ and the slide $\beta$ are set to 1 hour. 
The system performance, both with respect to CE recognition times and memory usage is quite robust (see Figure \ref{fig:cer_nc_same_rules}(a)), even though the average number of input events, as shown in Figure \ref{fig:cer_nc_same_rules}(b), increases linearly. Predictably, the average number of recognized CEs remains the same. These experiments were run on twelve cores. 

Figures \ref{fig:cer_nc_same_rules-actual}(a)--(b) show the actual number of input events and recognized CEs respectively, for four of the twelve cores. As already mentioned, the grid partitioning is not perfectly balanced, and thus some cores have a higher load in terms of input events. Figure \ref{fig:cer_nc_same_rules-actual}(b) shows that the number of recognized CEs does not depend only on the number of input events. For example, the top right core produces consistently more CEs, although it accepts more or less the same number of critical MEs as the top left and the bottom left cores.
More importantly, Figure \ref{fig:cer_nc_same_rules-actual}(a) illustrates the effect of the compression achieved by the trajectory detection module. Without this compression, RTEC would have to treat each incoming event as potentially critical, thus taking into consideration all input events shown in this figure (see the 100\% setting), as opposed to just considering the events annotated as critical by the trajectory detection module (see the 0\% setting). 


\subsubsection{Artificially enlarged data streams}

\begin{figure}[tp]
\centering
\subfigure[Average CE recognition times.]{ 
\includegraphics[width=0.46\textwidth]{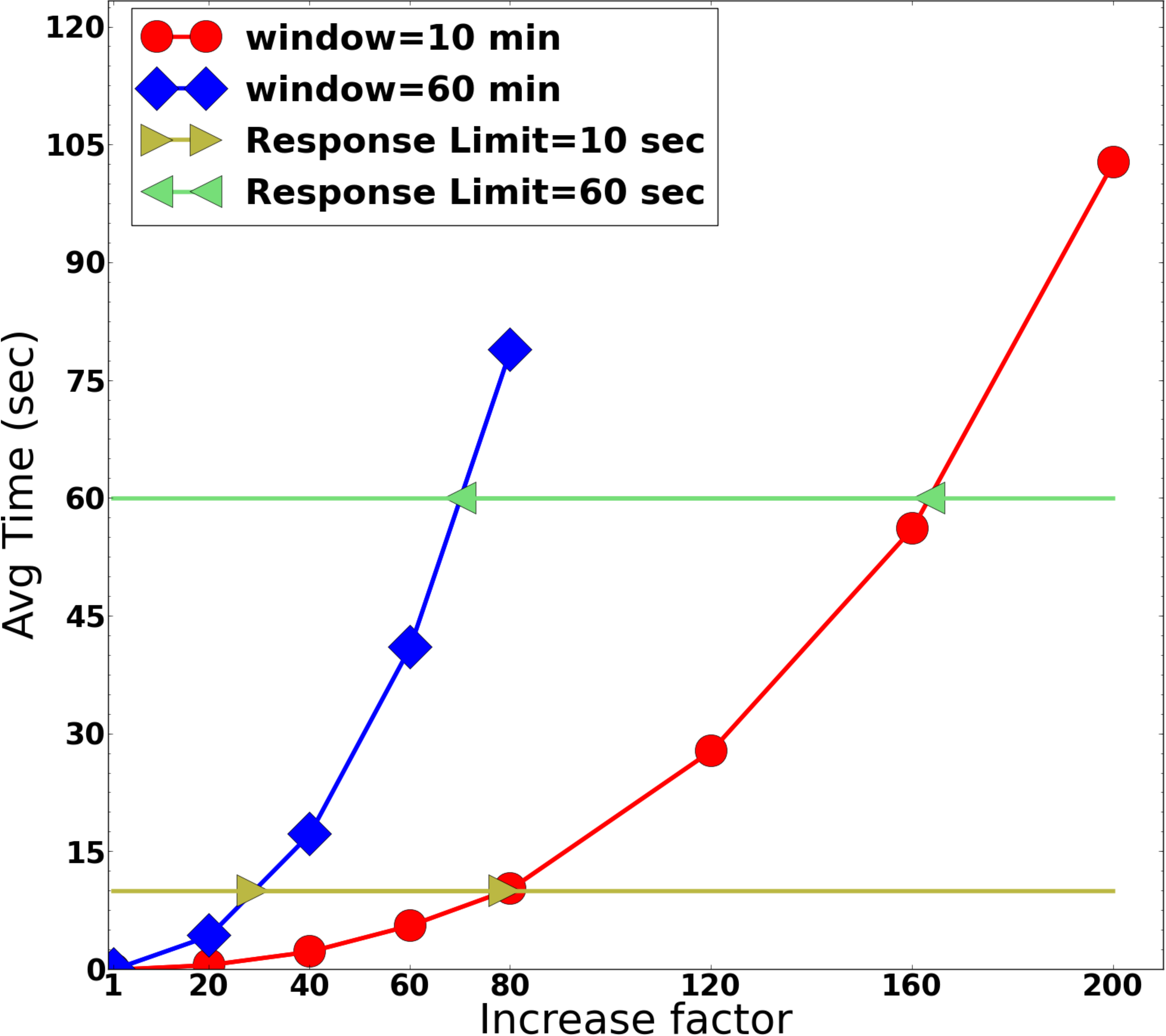}
}
\subfigure[Average memory usage.]{
\includegraphics[width=0.46\textwidth]{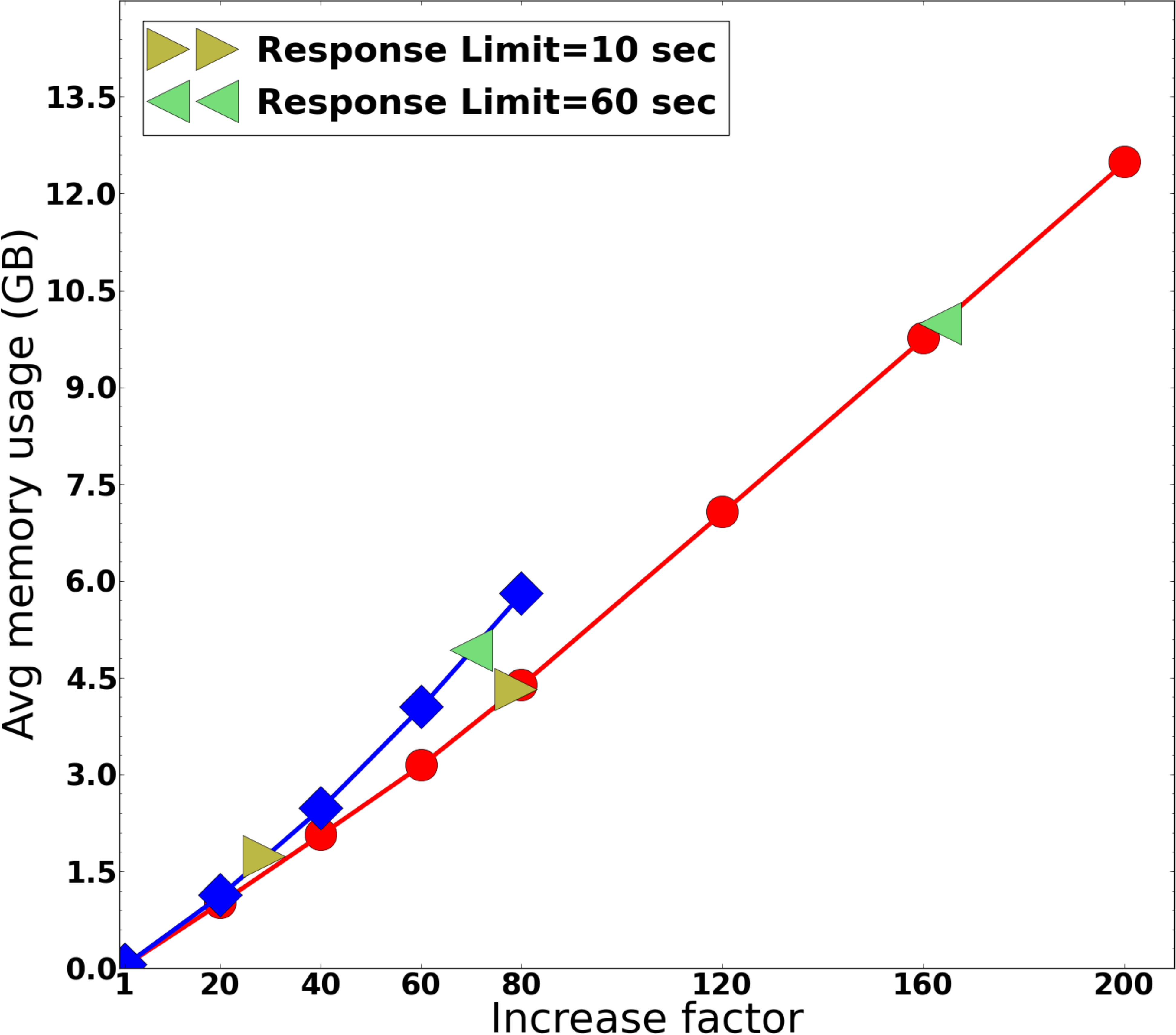}
}
\subfigure[Average number of MEs.]{
\includegraphics[width=0.46\textwidth]{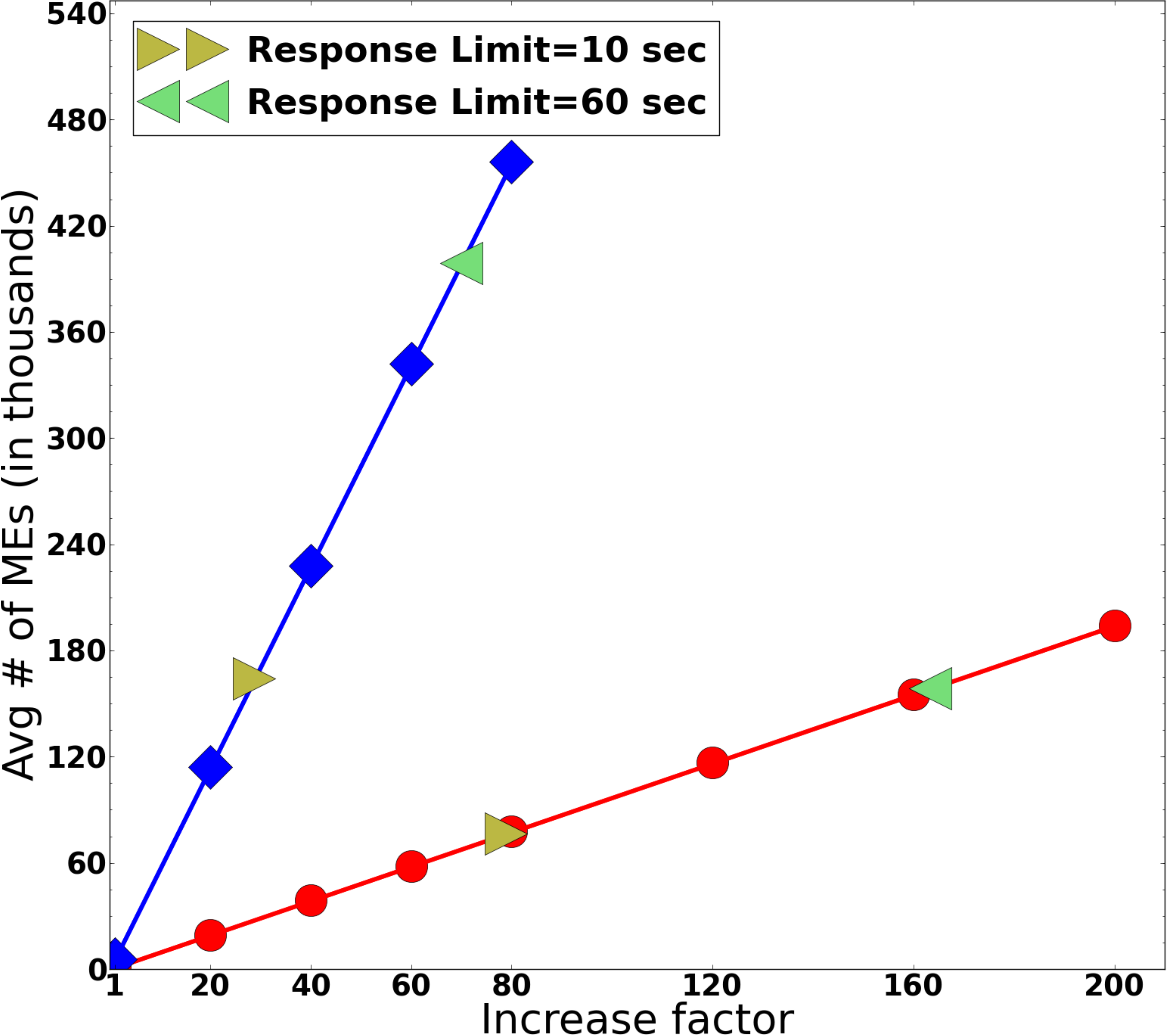}
}
\subfigure[Average number of CEs.]{
\includegraphics[width=0.46\textwidth]{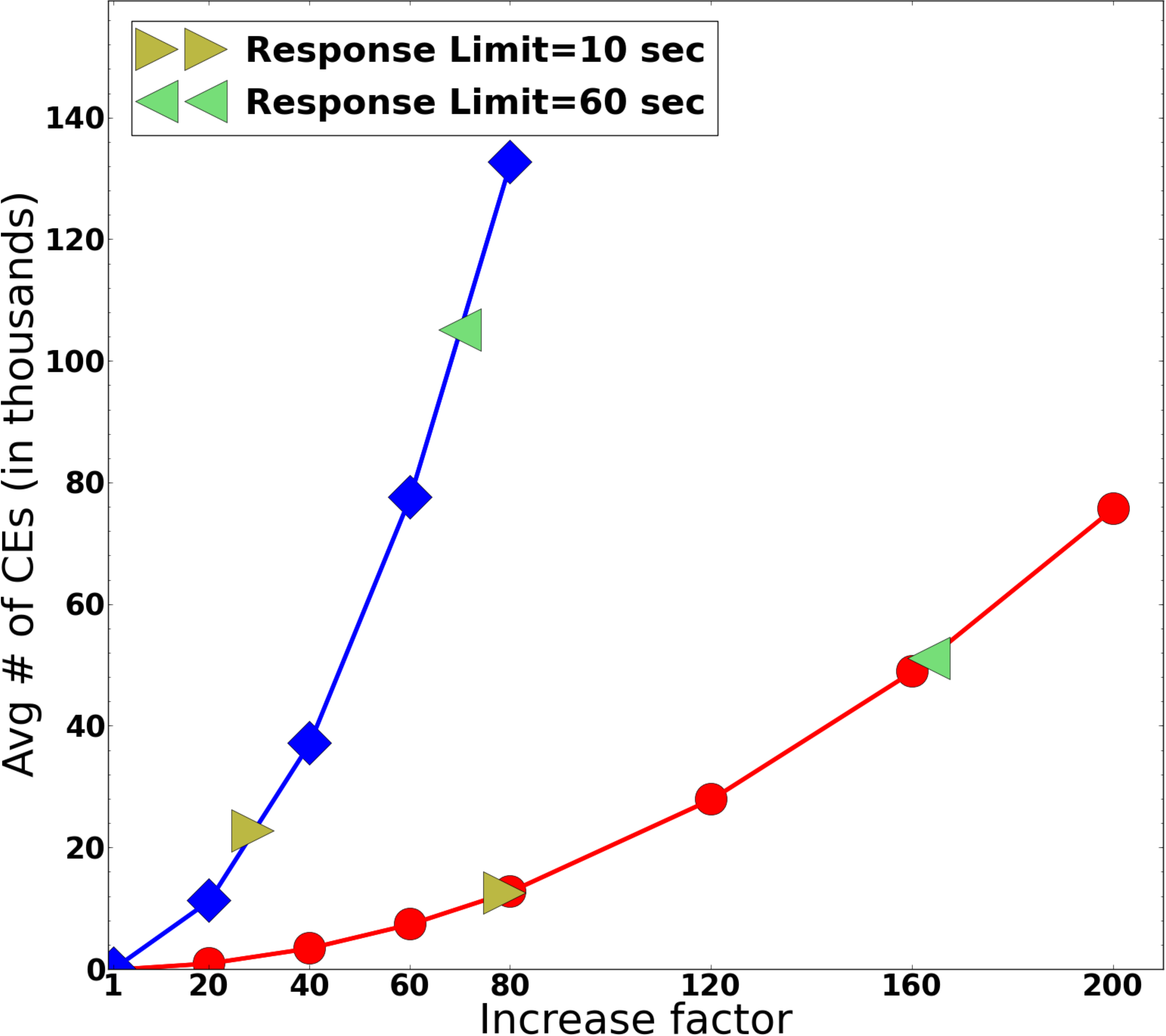}
}
\subfigure[CEs/MEs Ratio.]{
\includegraphics[width=0.46\textwidth]{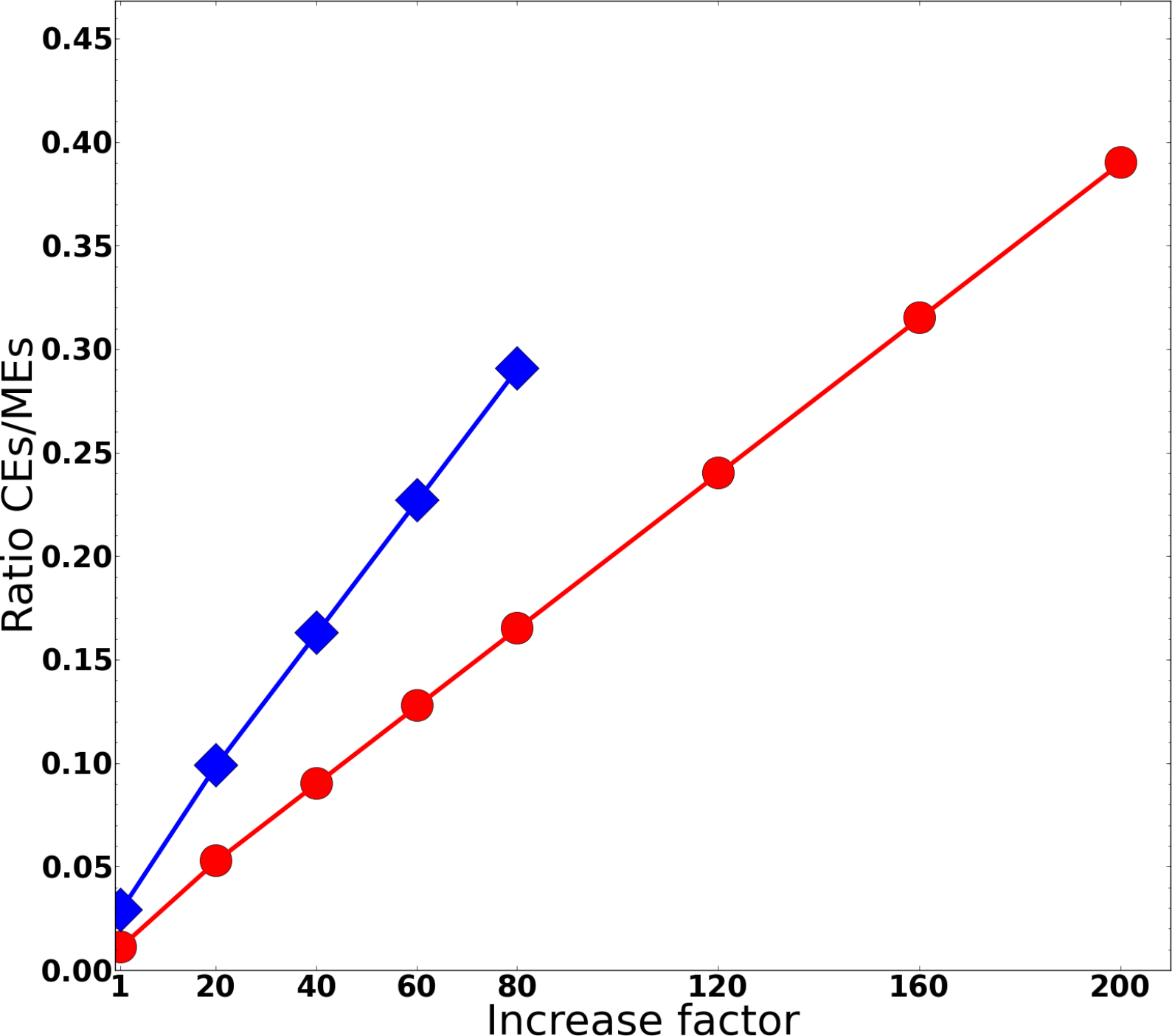}
}
\subfigure[Average number of internal list instances.]{
\includegraphics[width=0.46\textwidth]{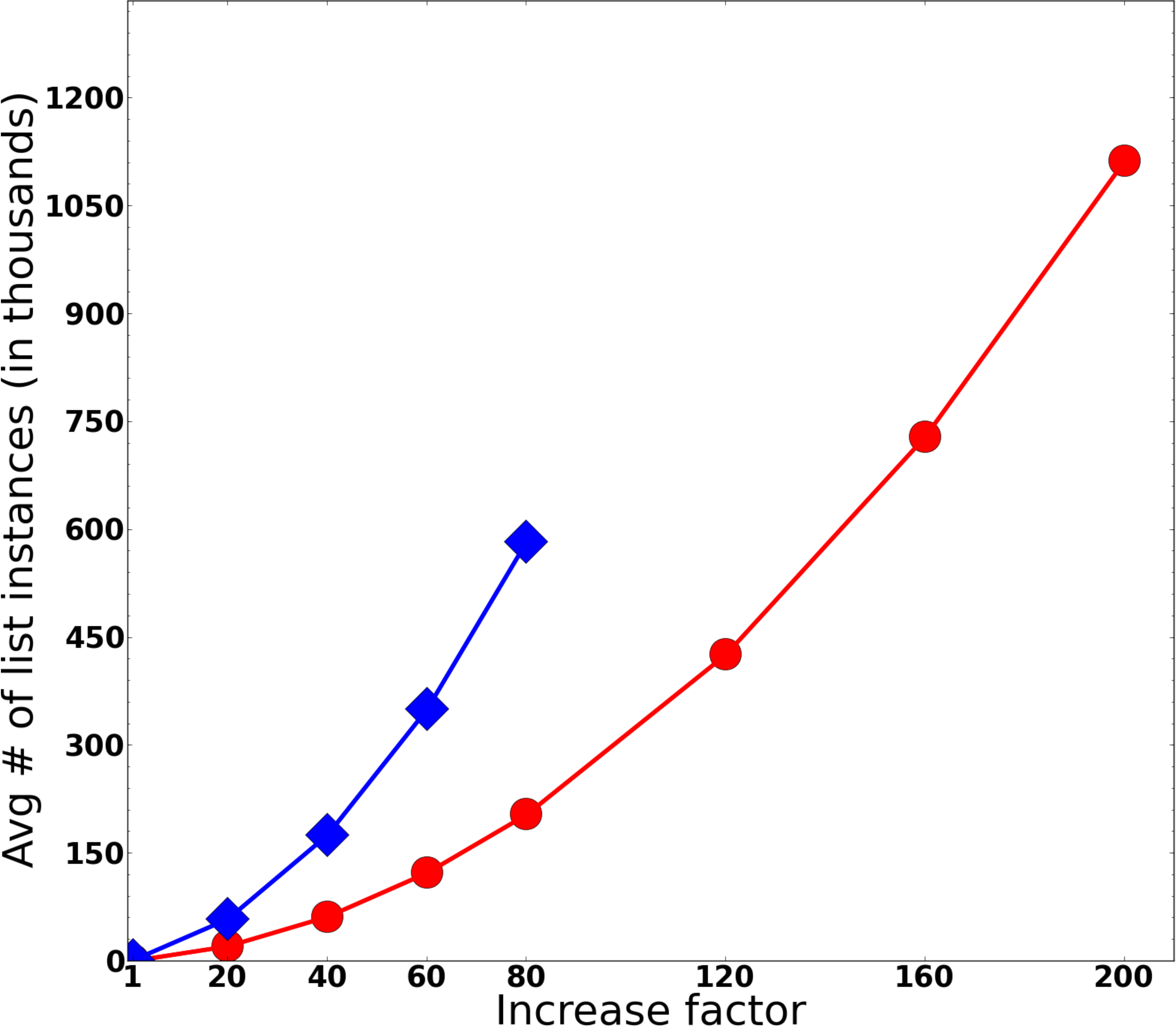}
}
\caption{CE recognition for approximately 128K--1,28M vessels and 320M--3,2B MEs. Response limits are indicated with lines in Figure \ref{fig:cer_replic}(a) and with points in Figures \ref{fig:cer_replic}(b)--(d).}
\label{fig:cer_replic}
\end{figure}

To test further the performance of RTEC, we created enlarged data streams by inserting extra critical MEs to the real dataset. No normal events were added for these experiments. For each vessel, we replicated its trajectories by a specified number of times (increase factor). The replicated trajectories were assigned to new vessels which do not exist in the original dataset, thus adding more vessels as well. An increase factor $X$ implies that we have exactly $X$ times as many critical MEs and $X$ times as many vessels, compared to the original dataset of $\approx$16,000,000 MEs and $\approx$6,500 vessels. We varied the increase factor from $20$, producing datasets of $\approx$128,000 vessels and 320,000,000 MEs, to $200$, creating data of $\approx$1,280,000 vessels and 3,200,000,000 MEs.
To the best of our knowledge, this is the most comprehensive evaluation of CE recognition techniques in the maritime domain. Compared to e.g.~\cite{terroso-saenz_complex_2015}, our surveillance area, number of vessels and data volume are substantially larger. Moreover, CE recognition needs to consider (a large number of) protected areas.

Figure \ref{fig:cer_replic} displays the experimental results. Twelve processors are used in parallel, the slide $\beta$ is set to 1 min, and the window sizes $\omega$ are 60 min (1 hour) and 10 min. We set two `response limits' to 10 sec and 60 sec---we stopped performing tests once the response limit of 60 sec was exceeded. 

Figure \ref{fig:cer_replic}(a) shows the average CE recognition times while Figure \ref{fig:cer_replic}(b) the average memory usage. As expected, a smaller window produces both lower recognition times and lower memory usage. Figures \ref{fig:cer_replic}(c) and \ref{fig:cer_replic}(d) display respectively the average number of MEs and CEs. We also show how RTEC performs with respect to the two response limits of 10 and 60~sec. Assume, e.g.~that we want to guarantee that RTEC responds within 60 sec.  In this case, for a window of 60~min, RTEC can handle data streams that are $\approx$70 times larger ($\approx$450,000 vessels)  than the original one (see Figure \ref{fig:cer_replic}(a)). This means that a window may include up to approximately 400,000 MEs, 
while recognizing around 105,000 CEs (see the 60 sec response limit markers on the 60~min window lines in Figures \ref{fig:cer_replic}(c) and \ref{fig:cer_replic}(d) respectively). Memory consumption is 4,9 GB (see the 60 sec response limit marker on the 60~min window line in Figure \ref{fig:cer_replic}(b)). Similarly, for a window of 10~min, RTEC is guaranteed to respond within 60~sec even in data streams that are 160 times larger than the original ($\approx$1,000,000 vessels). 
 
Comparing the 60~min~window/80~increase~factor setting against the 10~min~window/200~increase~factor setting, we see that the latter has fewer MEs and CEs (see the right-most marks of the window lines in Figures \ref{fig:cer_replic}(c) and \ref{fig:cer_replic}(d)). However, both the CE recognition times and memory usage are higher (see the right-most marks of the window lines in Figures \ref{fig:cer_replic}(a) and \ref{fig:cer_replic}(b)). One reason for this behavior is that the complexity of the recognition task is higher in the 10~min~window/200~increase~factor setting. 
A simple measure of complexity in CE recognition is the ratio of CEs to MEs. This ratio is depicted in Figure \ref{fig:cer_replic}(e). 

As a further step towards understanding RTEC's behavior, we investigated how it stores some of its internal structures. During CE recognition,  RTEC maintains several lists of time intervals for different types of event and fluent. In Figure \ref{fig:cer_replic}(f), we plot the average number of instances of these lists. The results show that these lists are mostly responsible for the increased recognition times and memory usage, as the increase factor grows. More specifically, the lists concerning $\mathit{possibleRendezvous}$ and $\mathit{possiblePicking}$ event 
(see rules \eqref{eq:possible-rendezvous} and \eqref{eq:package-picking} respectively), 
require that all combinations of vessels within some area are checked. Therefore, when the number of vessels increases, as is the case with replicated trajectories, the number of possible combinations is increased by the square of the number of vessels, hence the parabolic lines in Figure \ref{fig:cer_replic}(f). On the other hand, a window increase does not result in many more vessels being considered. Many of the added MEs may refer to extra messages transmitted by the same vessel over the longer duration of a larger window. Note that this is a different kind of complexity than the one reflected in the ratio of CEs to MEs. In the case where we have more combinations to check, this does not necessarily imply that proportionally more CEs will be recognized, 
since their definitions might not be satisfied. RTEC has to check all possible combinations though, 
thus incurring a higher latency and memory consumption.

\section{System Deployment}
\label{sec:deployment}

\begin{figure*}[t]
\centering
{\includegraphics[width=\textwidth]{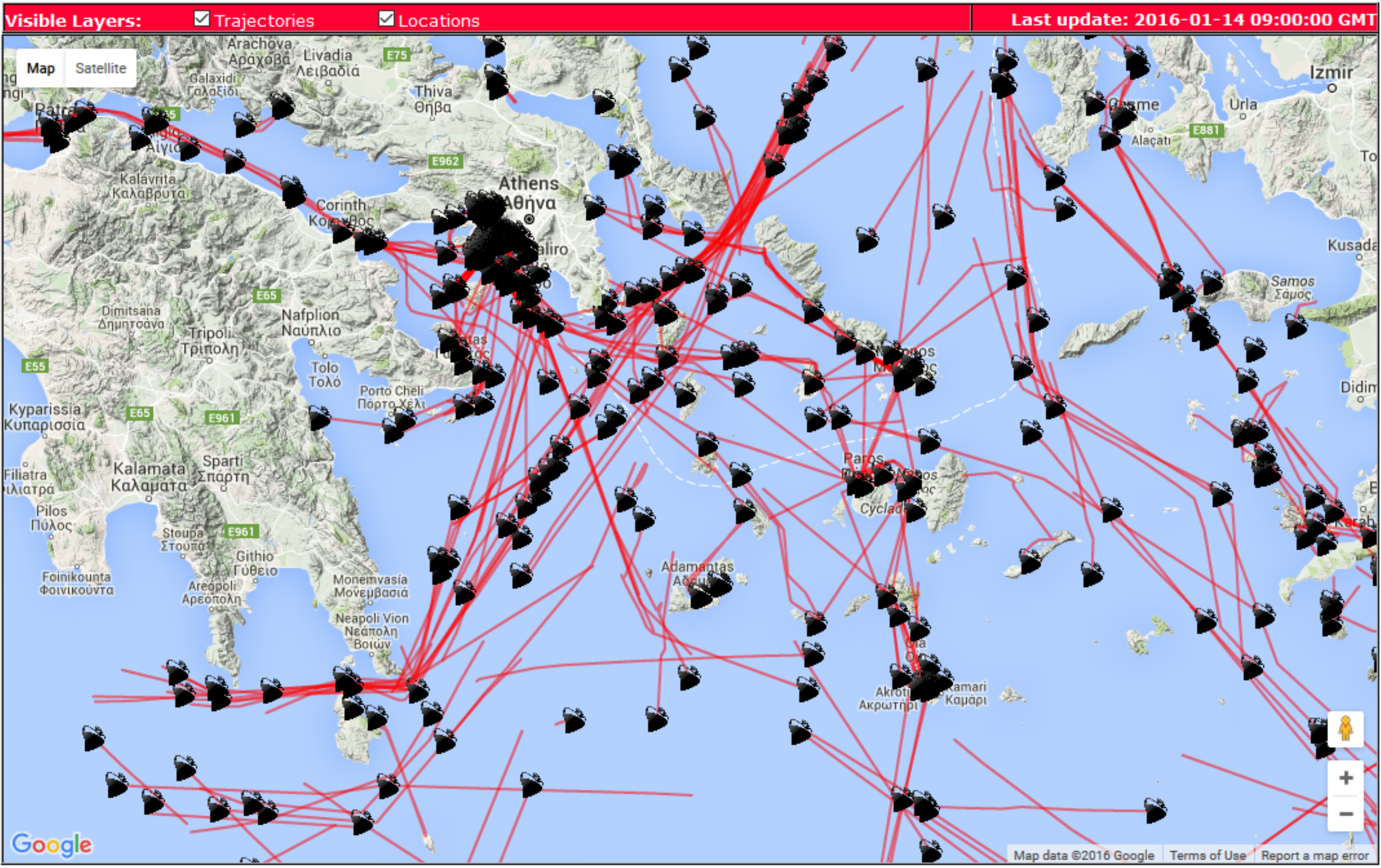}}
\caption{Monitoring real-time AIS positions and vessel trajectories in the Aegean.} \label{demo}
\end{figure*}

In order to verify its operational capabilities, we have set up our system to monitor {\em live} AIS feeds from vessels across the Aegean Sea. Fresh positions are periodically fetched from a database (maintained by the University of the Aegean) that receives any transmitted AIS messages. Each retrieved record only includes vessel identification, geographic position, and a timestamp. Currently, fresh input is fetched every hour (the sliding step $\beta$ of the window employed in this instance), although a different period may be chosen. Typically, we observed that up to 40,000 fresh records can be fetched per hour, which is a rather moderate amount of geostreaming data compared with the arrival rates simulated in our empirical validation (cf. Section \ref{sec:empirical}).

Once each batch of fresh data arrives, it is being consumed by the online mobility tracker in order to identify any critical points in each vessel's course. The resulting points with their annotations (turn, stop, etc.) are then archived into a PostgreSQL database. But the Trajectory Exporter module also converts them into KML files for map visualization in a web application, as illustrated in Figure~\ref{demo}. Critical points may be exported upon detection, but in order to avoid duplication of data across the modules, we opt to export them once they get evicted from the sliding window (currently set to a range of $\omega=$6 hours). Hence, exported results currently have a lag of 6 hours from the current time, but this is only a configuration parameter.


This deployment against real-time AIS messages collected across the Aegean has been activated since April 2015. This confirms that our system can integrate with a precious source of online data, offering to marine experts and authorities the means to instantly locate, recognize, and correlate events from real-time vessel traces.

\section{Summary \& Future Work}
\label{sec:summary}

In this paper, we introduced a system that monitors activity of thousands of vessels and can instantly recognize events with a potentially serious impact on the environment and on safe navigation at sea. The system can sustain massive streaming messages from vessels and can filter out noise and redundant positions along their course. Hence, it can retain only succinct synopses of vessel trajectories, drastically reducing the original path into few critical points that convey major motion features. As empirically validated, with a proper parametrization our suggested trajectory summarization may incur a compression ratio of almost 98\%, with tolerable error in the resulting approximation. Moreover, trajectory detection is highly scalable and can be easily parallelized in order to sustain very high arrival rates in the input stream. Furthermore, this reduced information may be readily analyzed online for Complex Event (CE) recognition. Equipped with efficient pattern matching algorithms, this module correlates critical trajectory positions with static geographical data, and detects suspicious or dangerous situations, such as illegal shipping, suspicious vessel delay, vessel rendezvous, suspicious area, vessel pursuit and package picking. We showed that the CE recognition module performs in real-time using real data as well as synthetically enlarged datasets that include up to 1,28M vessels and 3,2B critical positions. 


We plan further extensions and improvements in the existing implementation. First, since trajectory detection may be sensitive to parameters, we intend to study advanced methods for adaptive, auto-calibrated parameterization depending on the size, the type, and the motion patterns of vessels. 
Also, creating CE patterns manually is painstaking and error-prone. To facilitate the process of CE pattern construction, we plan to employ a recent framework for incremental structure learning that takes advantage of Big Data in order to construct Event Calculus programs \cite{katzourisML}. 
Besides, maritime surveillance exhibits various types of uncertainty \cite{DBLP:journals/inffus/SnidaroVB15}. AIS messages are often corrupt, with incorrect or missing fields. Furthermore, maritime CE patterns do not account for all possible situations. To deal with these issues, we have been developing an Event Calculus dialect for Markov Logic Networks \cite{skarlatidisTOCL}. This way, we can use weight learning techniques \cite{Duchi:2011:ASM:1953048.2021068} for estimating the confidence values of CE patterns, and subsequently perform probabilistic inference.  
Maritime surveillance may also benefit from combining multiple data sources. For example, the use of heterogeneous data sources, as in \cite{terroso-saenz_complex_2015}, can help in constructing more refined CE patterns. 
We aim to increase our data sources in order to improve monitoring quality and possibly recognize additional events.
Last, but not least, it would be challenging to apply ideas from our methodology against other sources of big geostreaming mobility data, e.g., traces of aircrafts or vehicles. Although the particular definitions of events may differ, as well as their configurations, we expect that our methodology may still be valid.

\begin{acknowledgements}
This work was funded partly by the {\em``AMINESS: Analysis of Marine INformation for Environmentally Safe Shipping''} project, which was co-financed by the European Fund for Regional Development and from Greek National funds, and partly by the EU-funded H2020 datACRON project. We wish to thank IMIS Hellas, our partner in AMINESS, for providing the AIS dataset used in the experiments.
\end{acknowledgements}

\bibliographystyle{spmpsci}      

\bibliography{aabib,traj_ref}

\begin{thebibliography}{10}
\providecommand{\url}[1]{{#1}}
\providecommand{\urlprefix}{URL }
\expandafter\ifx\csname urlstyle\endcsname\relax
  \providecommand{\doi}[1]{DOI~\discretionary{}{}{}#1}\else
  \providecommand{\doi}{DOI~\discretionary{}{}{}\begingroup
  \urlstyle{rm}\Url}\fi

\bibitem{agrawal08}
Agrawal, J., Diao, Y., Gyllstrom, D., Immerman, N.: Efficient pattern matching
  over event streams.
\newblock In: SIGMOD (2008)

\bibitem{alevizos15}
Alevizos, E., Artikis, A., Patroumpas, K., Vodas, M., Theodoridis, Y., Pelekis,
  N.: How not to drown in a sea of information: An event recognition approach.
\newblock In: IEEE International Conference on Big Data (2015)

\bibitem{arasu06}
Arasu, A., Babu, S., Widom, J.: The {CQL} continuous query language: semantic
  foundations and query execution.
\newblock The VLDB Journal \textbf{15}(2), 121--142 (2006)

\bibitem{DBLP:journals/tkde/ArtikisSP15}
Artikis, A., Sergot, M.J., Paliouras, G.: An event calculus for event
  recognition.
\newblock {IEEE} Trans. Knowl. Data Eng. \textbf{27}(4), 895--908 (2015)

\bibitem{bai06}
Bai, Y., Thakkar, H., Wang, H., Luo, C., Zaniolo, C.: A data stream language
  and system designed for power and extensibility.
\newblock In: CIKM, pp. 337--346 (2006)

\bibitem{brenna07}
Brenna, L., Demers, A.J., Gehrke, J., Hong, M., Ossher, J., Panda, B.,
  Riedewald, M., Thatte, M., White, W.M.: Cayuga: a high-performance event
  processing engine.
\newblock In: SIGMOD, pp. 1100--1102 (2007)

\bibitem{[CWT06]}
Cao, H., Wolfson, O., Trajcevski, G.: Spatio-temporal data reduction with
  deterministic error bounds.
\newblock VLDB Journal \textbf{15}(3), 211--228 (2006)

\bibitem{clark78}
Clark, K.: Negation as failure.
\newblock In: H.~Gallaire, J.~Minker (eds.) Logic and Databases, pp. 293--322.
  Plenum Press (1978)

\bibitem{cugola10}
Cugola, G., Margara, A.: {TESLA}: a formally defined event specification
  language.
\newblock In: DEBS, pp. 50--61 (2010)

\bibitem{dindar11}
Dindar, N., Fischer, P.M., Soner, M., Tatbul, N.: Efficiently correlating
  complex events over live and archived data streams.
\newblock In: DEBS, pp. 243--254 (2011)

\bibitem{dousson07}
Dousson, C., Maigat, P.L.: Chronicle recognition improvement using temporal
  focusing and hierarchisation.
\newblock In: IJCAI, pp. 324--329 (2007)

\bibitem{Duchi:2011:ASM:1953048.2021068}
Duchi, J., Hazan, E., Singer, Y.: Adaptive subgradient methods for online
  learning and stochastic optimization.
\newblock J. Mach. Learn. Res. \textbf{12}, 2121--2159 (2011)

\bibitem{eckert10}
Eckert, M., Bry, F.: Rule-based composite event queries: the language
  xchange$^{\mbox{eq}}$ and its semantics.
\newblock Knowledge Information Systems \textbf{25}(3), 551--573 (2010)

\bibitem{[EKSX96]}
Ester, M., Kriegel, H., Sander, J., Xu, X.: A density-based algorithm for
  discovering clusters in large spatial databases with noise.
\newblock In: KDD, pp. 226--231 (1996)

\bibitem{Garcia11}
Garcia, J., Gomez-Romero, J., Patricio, M., Molina, J., Rogova, G.: On the
  representation and exploitation of context knowledge in a harbor surveillance
  scenario.
\newblock In: {FUSION}, pp. 1--8 (2011)

\bibitem{[GT13]}
Golab, L., Johnson, T.: Data stream warehousing {\em (tutorial)}.
\newblock In: ACM SIGMOD, pp. 949--952 (2013)

\bibitem{DBLP:conf/sysose/IdiriN12}
Idiri, B., Napoli, A.: The automatic identification system of maritime accident
  risk using rule-based reasoning.
\newblock In: SoSE, pp. 125--130 (2012)

\bibitem{[KBC13]}
Katsilieris, F., Braca, P., Coraluppi, S.: Detection of malicious {AIS}
  position spoofing by exploiting radar information.
\newblock In: FUSION, pp. 1196--1203 (2013)

\bibitem{katzourisML}
Katzouris, N., Artikis, A., Paliouras, G.: Incremental learning of event
  definitions with inductive logic programming.
\newblock Machine Learning \textbf{100}(2--3), 555--585 (2015)

\bibitem{[KDA+10]}
Kazemitabar, S.J., Demiryurek, U., Ali, M.H., Akdogan, A., Shahabi, C.:
  Geospatial stream query processing using {M}icrosoft {SQL} {S}erver
  {S}treaminsight.
\newblock {PVLDB} \textbf{3}(2), 1537--1540 (2010)

\bibitem{kowalski86}
Kowalski, R., Sergot, M.: A logic-based calculus of events.
\newblock New Generation Computing \textbf{4}(1) (1986)

\bibitem{kramer09}
Kr{\"a}mer, J., Seeger, B.: Semantics and implementation of continuous sliding
  window queries over data streams.
\newblock ACM Transactions on Database Systems \textbf{34}(1) (2009)

\bibitem{Laere09}
van Laere, J., Nilsson, M.: Evaluation of a workshop to capture knowledge from
  subject matter experts in maritime surveillance.
\newblock In: FUSION, pp. 171--178 (2009)

\bibitem{[LDR11]}
Lange, R., D{\"u}rr, F., Rothermel, K.: Efficient real-time trajectory
  tracking.
\newblock VLDB Journal \textbf{20}(5), 671--694 (2011)

\bibitem{li05}
Li, G., Jacobsen, H.A.: Composite subscriptions in content-based
  publish/subscribe systems.
\newblock In: Middleware (2005)

\bibitem{[MdB04]}
Meratnia, N., de~By, R.: Spatiotemporal compression techniques for moving point
  objects.
\newblock In: EDBT, pp. 765--782 (2004)

\bibitem{[MBBW15]}
Millefiori, L.M., Braca, P., Bryan, K., Willett, P.: Adaptive filtering of
  imprecisely time-stamped measurements with application to {AIS} networks.
\newblock In: FUSION, pp. 359--365 (2015)

\bibitem{[MT11]}
Moga, A., Tatbul, N.: Up{S}tream: A storage-centric load management system for
  real-time update streams.
\newblock PVLDB \textbf{4}(12), 1442--1445 (2011)

\bibitem{orourke_computational_1998}
O'Rourke, J.: Computational {Geometry} in {C}.
\newblock Cambridge University Press (1998)

\bibitem{[PVB13]}
Pallotta, G., Vespe, M., Bryan, K.: Vessel pattern knowledge discovery from
  {AIS} data: {A} framework for anomaly detection and route prediction.
\newblock Entropy \textbf{15}(6), 2218--2245 (2013)

\bibitem{paschke05}
Paschke, A.: {ECA}-{R}ule{ML}: An approach combining {ECA} rules with temporal
  interval-based {KR} event/action logics and transactional update logics.
\newblock Tech. Rep.~11, TU M\"{u}nchen (2005)

\bibitem{paschke09ruleml}
Paschke, A., Kozlenkov, A.: Rule-based event processing and reaction rules.
\newblock In: RuleML, LNCS 5858 (2009)

\bibitem{[PAK+15]}
Patroumpas, K., Artikis, A., Katzouris, N., Vodas, M., Theodoridis, Y.,
  Pelekis, N.: Event recognition for maritime surveillance.
\newblock In: EDBT, pp. 629--640 (2015)

\bibitem{[PS11]}
Patroumpas, K., Sellis, T.: Maintaining consistent results of continuous
  queries under diverse window specifications.
\newblock Information Systems \textbf{36}(1), 42--61 (2011)

\bibitem{[PPS07]}
Potamias, M., Patroumpas, K., Sellis, T.: Online amnesic summarization of
  streaming locations.
\newblock In: SSTD, pp. 148--165 (2007)

\bibitem{local-strat}
Przymusinski, T.: On the declarative semantics of stratified deductive
  databases and logic programs.
\newblock In: Found. of Deductive Databases and Logic Programming. Morgan
  (1987)

\bibitem{shahir15}
Shahir, H.Y., Glasser, U., Shahir, A.Y., Wehn, H.: Maritime situation analysis
  framework: {Vessel} interaction classification and anomaly detection.
\newblock In: {Big} {Data}, pp. 1279--1289 (2015)

\bibitem{skarlatidisTOCL}
Skarlatidis, A., Paliouras, G., Artikis, A., Vouros, G.: Probabilistic event
  calculus for event recognition.
\newblock ACM Transactions on Computational Logic \textbf{16}(2) (2015)

\bibitem{DBLP:journals/inffus/SnidaroVB15}
Snidaro, L., Visentini, I., Bryan, K.: Fusing uncertain knowledge and evidence
  for maritime situational awareness via markov logic networks.
\newblock Information Fusion \textbf{21}, 159--172 (2015)

\bibitem{terroso-saenz_complex_2015}
Terroso-Saenz, F., Valdes-Vela, M., Skarmeta-Gomez, A.F.: A complex event
  processing approach to detect abnormal behaviours in the marine environment.
\newblock Information Systems Frontiers pp. 1--16 (2015)

\bibitem{[WSCY99]}
Wolfson, O., Sistla, A., Chamberlain, S., Yesha, Y.: Updating and querying
  databases that track mobile units.
\newblock Distributed \& Parallel Databases \textbf{7}(3), 257--287 (1999)

\bibitem{DBLP:conf/sigmod/ZhangDI14}
Zhang, H., Diao, Y., Immerman, N.: On complexity and optimization of expensive
  queries in complex event processing.
\newblock In: {SIGMOD}, pp. 217--228 (2014)

\end{thebibliography}

\end{document}